\documentclass[runningheads]{llncs}

\usepackage{eccv}

\usepackage{eccvabbrv}

\usepackage{graphicx}
\usepackage{booktabs}

\usepackage[accsupp]{axessibility}  
\definecolor{cvprblue}{rgb}{0.21,0.49,0.74}
\usepackage{indentfirst}
\usepackage{bm}
\usepackage{lipsum}
\usepackage{tikz}
\usepackage{pgfplots}
\usepackage{mathtools}
\usepackage{multirow}
\usepackage{overpic}
\usepackage{makecell}
\usepackage{xcolor}
\usepackage{colortbl}
\usepackage{algorithm}
\usepackage{algpseudocode}

\definecolor{gray}{gray}{0.95}
\usepackage{listings}
\usepackage{color}
\usepackage{diagbox}
\definecolor{dkgreen}{rgb}{0,0.6,0}
\definecolor{mauve}{rgb}{0.58,0,0.82}

\usepackage{hyperref}

\usepackage{orcidlink}

\begin{document}

\title{RodinHD: High-Fidelity 3D Avatar Generation with Diffusion Models}

\titlerunning{RodinHD}

\author{Bowen Zhang\inst{1}$^*$\orcidlink{0000-0003-3520-1091} \and
Yiji Cheng\inst{2}$^*$\orcidlink{0009-0002-1457-2328} \and
Chunyu Wang\inst{3}$^{\dagger}$\orcidlink{0000-0002-9400-9107}
\and
Ting Zhang\inst{3}\orcidlink{0000-0002-3952-2522}
\and
Jiaolong Yang\inst{3}\orcidlink{0000-0002-7314-6567}
\and
Yansong Tang\inst{2}\orcidlink{0000-0002-1534-4549}
\and
Feng Zhao\inst{1}\orcidlink{0000-0001-6767-8105}
\and
Dong Chen\inst{3}\orcidlink{0000-0003-0588-9331}
\and
Baining Guo\inst{3}\orcidlink{0000-0001-8349-8868}}

\authorrunning{B.~Zhang and Y.~Cheng et al.}

{
	\renewcommand{\thefootnote}%
	{\fnsymbol{footnote}}
	\footnotetext[1]{Interns at Microsoft Research Asia. Equal contribution. $^{\dagger}$Corresponding author.}
}


\institute{University of Science and Technology of China \and Tsinghua University \and Microsoft Research Asia}

\maketitle
\begin{center}
    \scriptsize
    \centering
    \begin{tabular}{cc}
         \includegraphics[width=0.45\columnwidth]{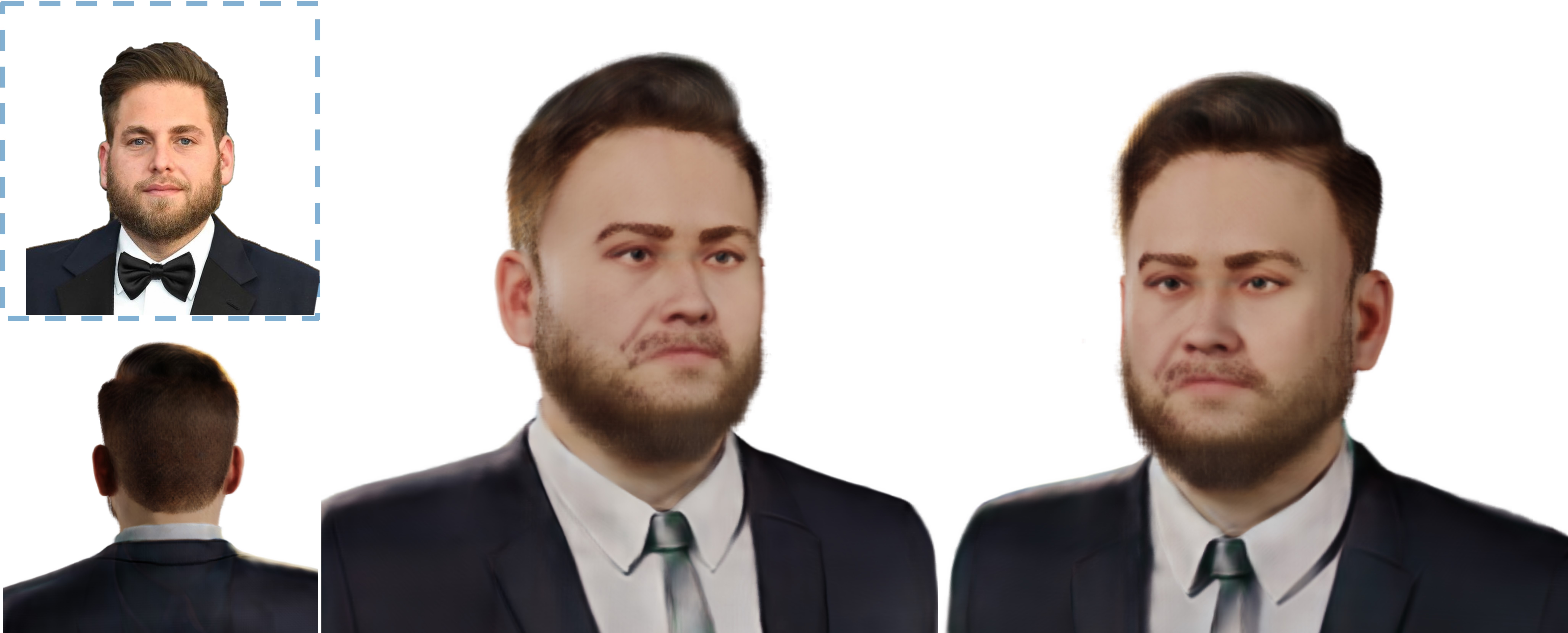} & \includegraphics[width=0.45\columnwidth]{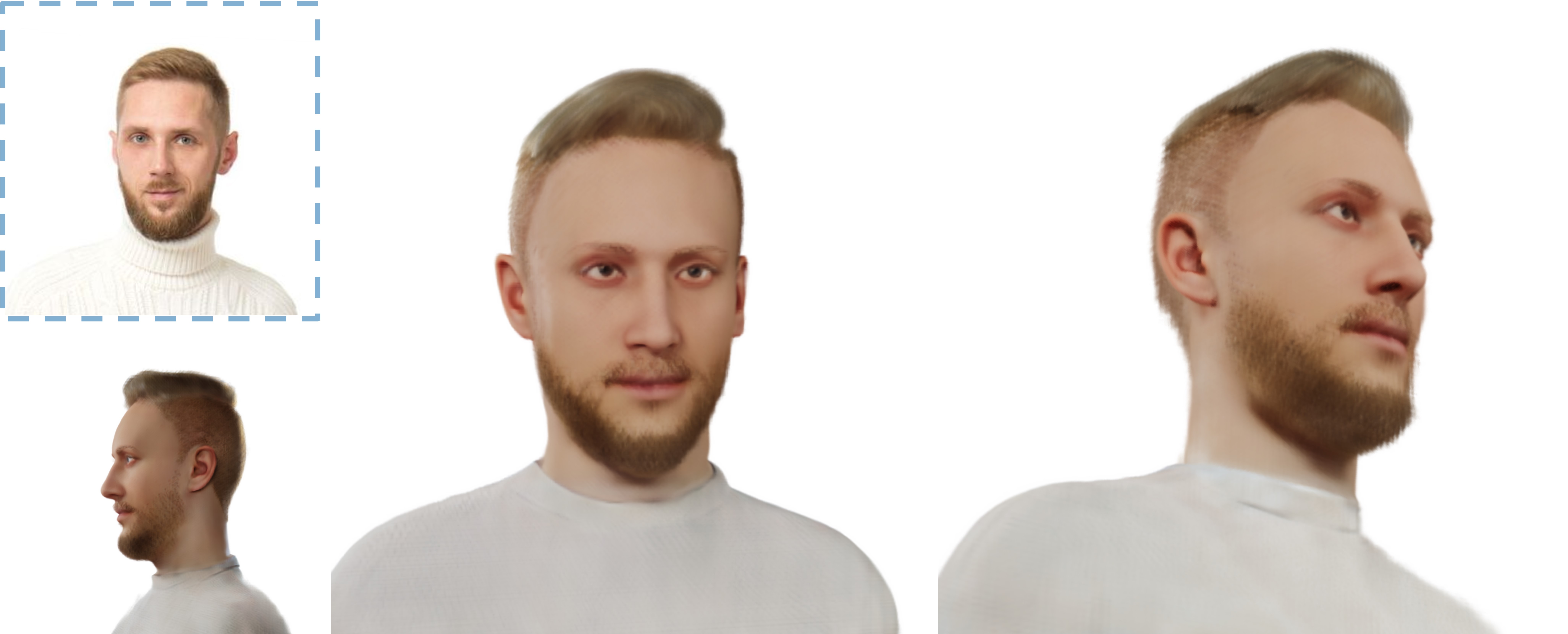}\\
         \multicolumn{2}{c}{3D avatars created from in-the-wild portrait images}\\
         \includegraphics[width=0.45\columnwidth]{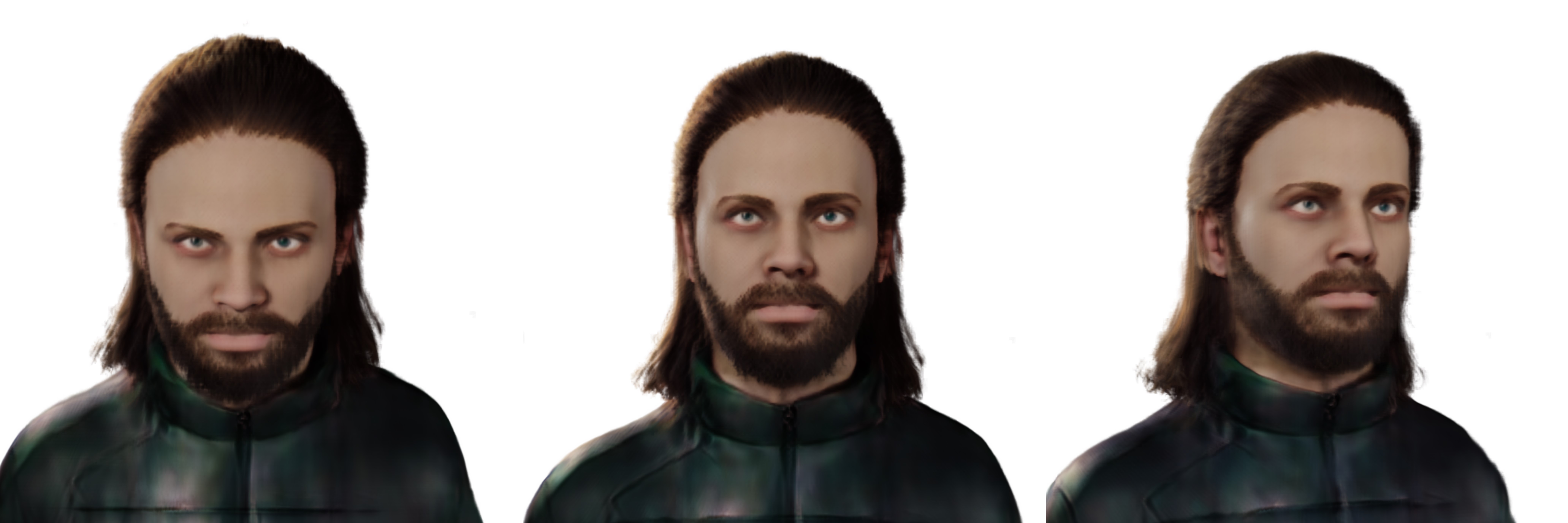} & \includegraphics[width=0.45\columnwidth]{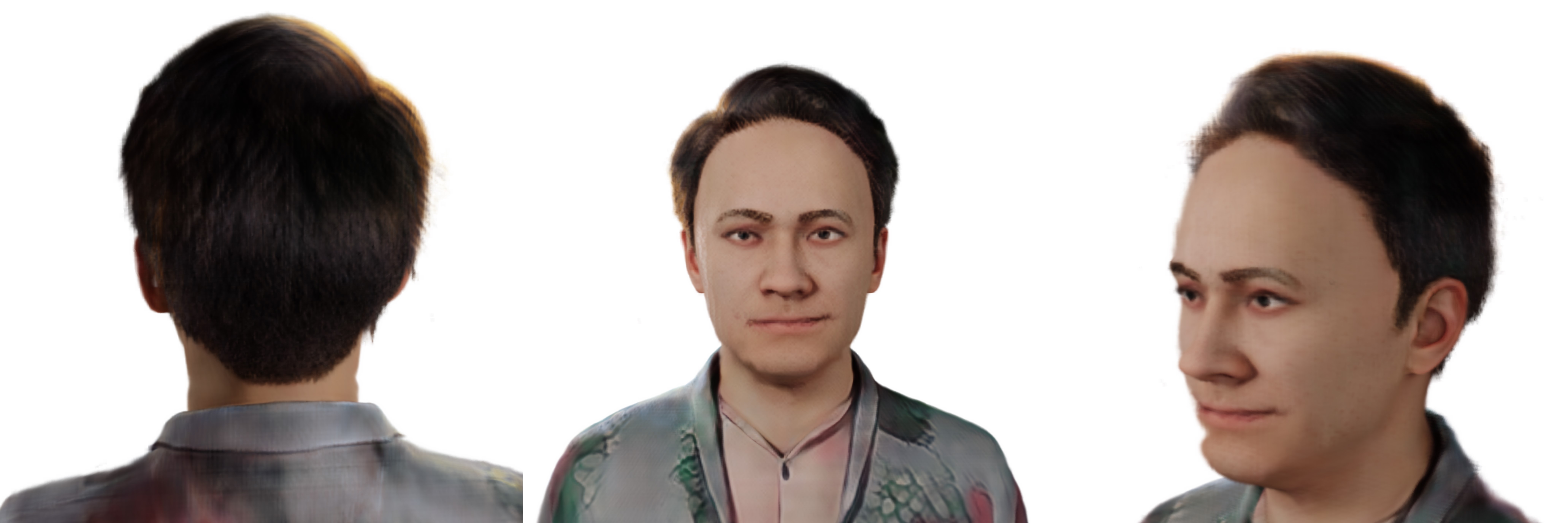}\\
         \makecell[c]{``\textit{Blender Synthetic Avatar, brown hair, boy, green} \\ \textit{eyes, black and green jacket, stubble, leather}" } & Unconditional sampled avatar
    \end{tabular}
    \captionof{figure}{RodinHD generates detailed 3D avatars from single portrait images (dashed boxes) without compromising cross-view consistency (first row). It also supports text-conditioned (second row left) or unconditional (second row right) generation. }
\label{fig:teaser}
\end{center}

\begin{abstract}
  We present RodinHD, which can generate high-fidelity 3D avatars from a portrait image. Existing methods fail to capture intricate details such as hairstyles which we tackle in this paper. We first identify an overlooked problem of catastrophic forgetting that arises when fitting triplanes sequentially on many avatars, caused by the MLP decoder sharing scheme. To overcome this issue, we raise a novel data scheduling strategy and a weight consolidation regularization term, which improves the decoder's capability of rendering sharper details. Additionally, we optimize the guiding effect of the portrait image by computing a finer-grained hierarchical representation that captures rich 2D texture cues, and injecting them to the 3D diffusion model at multiple layers via cross-attention. When trained on $46K$ avatars with a noise schedule optimized for triplanes, the resulting model can generate 3D avatars with notably better details than previous methods and can generalize to in-the-wild portrait input. See~\cref{fig:teaser} for some examples. Project page: \href{https://rodinhd.github.io/}{https://rodinhd.github.io/}.
\end{abstract}
   
\section{Introduction}
High-fidelity 3D avatar generation has many applications in fields such as gaming and metaverse. Recent development of generative diffusion models~\cite{rombach2022high,gu2022vector,podell2023sdxl,ho2022cascaded,hang2023efficient} and implicit neural radiance fields~\cite{mildenhall2021nerf,wang2023rodin} has opened up new opportunities for automatic generation of 3D avatar ~\cite{wang2023rodin,muller2023diffrf,gupta20233dgen,shue20233d} at scale. However, current methods struggle to generate fine details, which is a core challenge that has to be addressed. Otherwise, a ``toy-like'' avatar is less effective in delivering practical values.

The work of 3D generative diffusion models~\cite{wang2023rodin,shue20233d,muller2023diffrf,jun2023shap,zeng2022lion,ntavelis2023autodecoding,gupta20233dgen,bautista2022gaudi,zhang2024gaussiancube} usually follow a two-stage framework. First, they compute a proxy 3D representation of fixed length such as triplanes or volumes from the original unstructured meshes or point clouds so that they can be handled by diffusion models. A paired decoder is jointly learned to render $360^\circ$ images from the representation. Different from the dominant NeRF fitting methods, the decoder here is shared among all avatars to decode novel generated triplanes. Then, they train diffusion models on the proxy representation to generate diverse avatars. However, they struggle to generate fine details such as sharp cloth textures and hair strands. To alleviate the problem, Rodin~\cite{wang2023rodin} applies a convolution refiner~\cite{wang2021towards} to complement the missing details for each rendered image. Although the 2D refiner improves the visual quality in one view, it significantly compromises 3D consistency, which is not tolerable in many applications.

\begin{figure*}[t]
  \centering
  \includegraphics[width=0.6\textwidth]{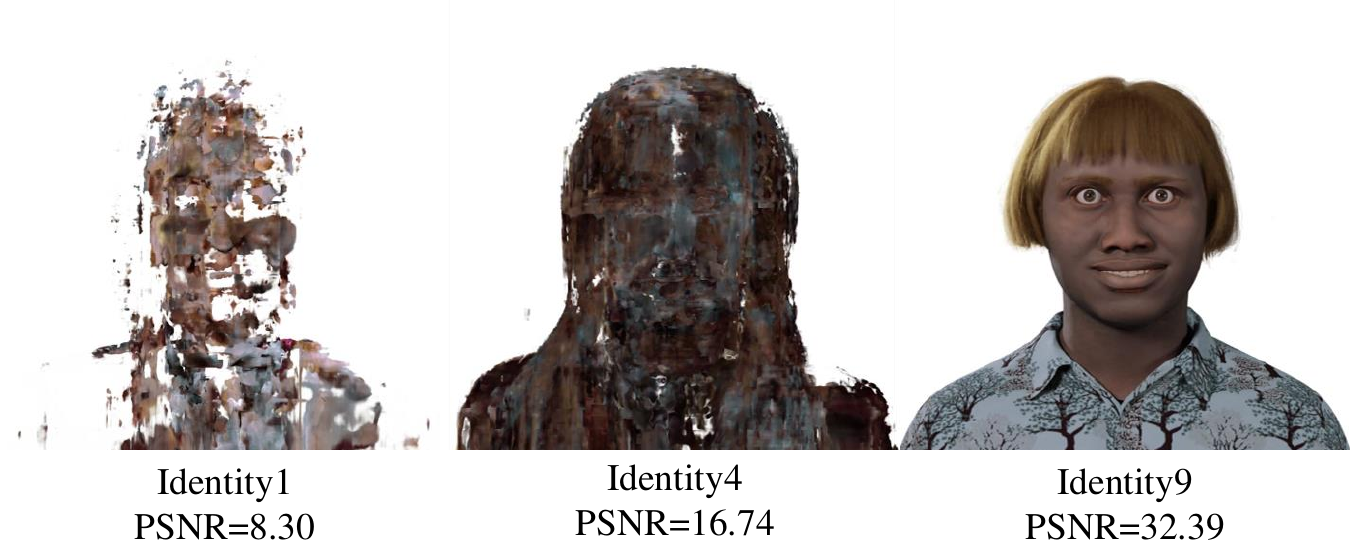}
  \caption{\textbf{Catastrophic forgetting}. As training proceeds, decoder gradually forgets the knowledge learned on the previous avatars of $1 \& 4$ and is overly adapted to avatar $9$.}
  \label{fig:catastrophic_forgetting}
\end{figure*}

In this work, we introduce RodinHD, which aims to improving the fidelity of avatars without any refiners. We begin by empirically showing that fitting triplanes sequentially on a large number of avatars suffers from catastrophic forgetting, which can result in under-fitted decoders incapable of generating intricate details on novel triplanes. This occurs because the triplane of an avatar is typically trained for many iterations before switching to the next one, in order to reduce the data transfer costs between CPUs and GPUs. However, this process gradually causes the shared decoder to forget knowledge learned from previous avatars, leading to a lack of generalizability. ~\cref{fig:catastrophic_forgetting} illustrates this issue with typical renderings of the resulting decoder. This problem has been largely overlooked in the literature, hindering the development of high-fidelity generation based on neural radiance fields.

To address this issue, we propose a novel data scheduling strategy called task replay, and a weight consolidation regularization term that effectively preserves the decoder's capability of rendering sharp details. The idea of task-replay is to switch avatars more frequently so that each can be seen periodically for multiple times, preventing the decoder from over-fitting to a single avatar. The weight consolidation regularization term prevents the critical weights from deviating far from its consolidated values. As a result, knowledge learned from previous data can be retained during training. The method effectively alleviates the forgetting problem and improve the model's capability to encode intricate details, paving the way for the subsequent generation step. 

We train a cascaded diffusion model on the triplanes for conditional generation. It consists of a \emph{base model}, which generates a low-resolution triplane conditioned on a portrait image, and an \emph{upsample model}, which subsequently generates a high-resolution triplane. Differing from the previous work~\cite{wang2023rodin} which uses CLIP~\cite{radford2021learning} to compute a global semantic token for the portrait image as conditions, we maximize its guiding effect by computing a finer-grained hierarchical representation to provide more detailed cues for the 3D diffusion model using a VAE-based image encoder. The multi-scale features are injected into different layers of U-Net via cross-attention, which significantly improves the coherence between the generated avatars and the portrait images. ~\cref{fig:pipeline} shows the process. Besides, inspired by~\cite{chen2023importance}, we also optimize the noise schedule for the triplane considering its high redundancy in both spatial and channel dimensions.

We train the model on $46$K digital avatars~\cite{wood2021fake} of diverse identities, expressions, hairstyles, and clothing. We render high-quality images at the resolution of $1024 \times 1024$. The resulting model is capable of generating highly detailed avatars with clear clothing textures and hairstyles using a simple diffusion model without extra refinement models. While only validated on avatars, the proposed techniques are general and can be applied to other 3D generation tasks.
\section{Related Work}

Early works in 3D generation have primarily focused on generating coarse 3D shapes, which are typically represented as meshes~\cite{liao2020towards, szabo2019unsupervised}, point clouds~\cite{achlioptas2018learning, li2021sp}, voxel grids~\cite{brock2016generative, wu2016learning}, and implicit neural representations~\cite{park2019deepsdf, sitzmann2020implicit}, using either GANs~\cite{goodfellow2020generative, karras2019style, zhang2022styleswin} or VAEs~\cite{kingma2019introduction}. However, it remains unclear whether these methods can be effectively applied to generate complex 3D avatars with rich details.

3D-aware GANs~\cite{chan2021pi,chan2022efficient,deng2022gram,xiang2022gram,tang2022explicitly,cheng2023efficient,schwarz2020graf,niemeyer2021giraffe,gao2022get3d,yin20233d} are able to generate high-resolution images with the aid of 2D upsampler and patch-based image discriminator. Nevertheless, they suffer from cross-view consistency~\cite{xia2023survey}. Gram~\cite{xiang2022gram} performs upsampling on the surface manifold to promote multiview consistency and efficiency but it cannot handle large viewpoint changes and complex geometry.
Moreover, they are prone to mode collapse due to training instabilities of GANs. 
Although score-distillation-based methods~\cite{poole2022dreamfusion,lin2022magic3d,tang2023make,sun2023dreamcraft3d,chen2024comboverse,tang2024make} are proposed to distill the 2D diffusion prior to a 3D representation with score function, they suffer from the Janus problem which prevents them from generating accurate geometry because the problem is under-determined.

Recently, diffusion models~\cite{dhariwal2021diffusion,ho2020denoising,song2020score} have achieved notable success in text-to-image synthesis, with a few on high-resolution image generation~\cite{rombach2022high,podell2023sdxl,saharia2022photorealistic,teng2023relay, gu2023matryoshka,ding2023patched,ma2022accelerating}. Inspired by this, many recent works apply diffusion models for 3D generation~\cite{wang2023rodin,shue20233d,muller2023diffrf,jun2023shap,zeng2022lion,ntavelis2023autodecoding,gupta20233dgen,chen2023single,bautista2022gaudi,tang2023flag3d,liu2023plan,dai2024motionlcm}. They first compute a proxy 3D representation from the raw data, and then train a diffusion model on the proxy data with either text or image as conditions.
However, most of them struggle to generate fine details such as cloth textures and hair strands, due to the lack of high-quality proxy data and a well-designed upsample network.
Rodin~\cite{wang2023rodin} exploits a convolution refiner~\cite{wang2021towards} to complement the missing details in each individually rendered image, but this compromises cross-view consistency.

Some recent work~\cite{rombach2022high,podell2023sdxl,saharia2022photorealistic,teng2023relay, gu2023matryoshka,ding2023patched,ma2022accelerating} have investigated high-resolution image generation, and obtained some interesting findings about optimal noise schedules, model architectures and capacity scaling principles. 
Despite this, we find the triplanes are essentially different from images, and directly transferring their findings barely works in our scenario.
For instance, Stable Diffusion~\cite{rombach2022high} uses a VAE to compress an image into a lower-resolution latent for diffusion. However, we observe that compressing triplanes by training a VAE will lose many high-frequency details in the renderings. Moreover, the considerations for designing the noise schedules are also different from those in the images.

\section{Method}

\begin{figure*}[t]
  \centering
  \includegraphics[width=\textwidth]{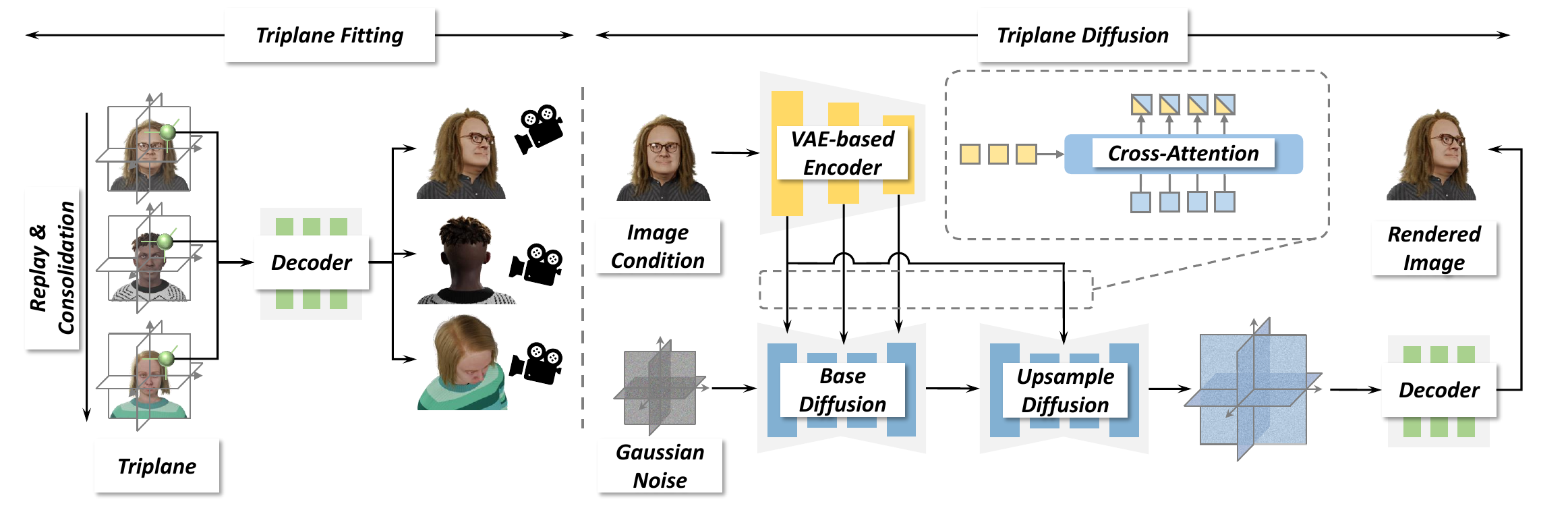}
  \caption{\textbf{Overview of our method.}  }
  \label{fig:pipeline}
\end{figure*}

Our framework comprises two primary steps: fitting and modeling. In the first step, we fit a high-resolution triplane $\bm{x}_{0} \in \mathbb{R}^{3 \times H_{\bm{x}}\times W_{\bm{x}} \times C}$ for each avatar, and learn a decoder $\mathcal{F}$, which is shared by all avatars, to render high-fidelity images from the triplanes. The parameters of $H_{\bm{x}}$ and $ W_{\bm{x}} $ denote the height and width of triplanes, taking value on 512, and $C$ is the number of channels, taking value on 32. In the second step, we train a 3D diffusion model $ \mathcal{G}$ that can generate triplanes from random noises $\bm\epsilon \in \mathbb{R}^{3 \times H_{\bm{x}}\times W_{\bm{x}} \times C}$ conditioned on a portrait image $\mathbf{I}_{\text{front}} \in \mathbb{R}^{H_\mathbf{I} \times W_\mathbf{I}\times 3}$. 
 ~\cref{fig:pipeline} shows the overall framework.
 During inference, we can render 360$^{\circ}$ high-resolution images $\left\{\hat{\mathbf{I}} \in \mathbb{R}^{H_\mathbf{I}\times W_\mathbf{I}\times 3}\right\}$ from the following process: 

\begin{align}
    &\mathcal{G}: (\bm\epsilon, \mathbf{I}_{\text{front}}) \xrightarrow{\text{Denosing}} \bm{x}_{0}, \\
    &\mathcal{F}:  \bm{x}_{0} \xrightarrow{\text{Rendering}} \left\{\hat{\mathbf{I}}\right\}.
\end{align}
The primary challenge arises from the high-resolution nature of the triplane, presenting difficulties in both fitting and modeling. 
We will elaborate on several key design considerations in the following.

\subsection{Triplane Fitting}
\label{sec:3d-data}
The increased resolution of the triplanes introduces a higher computational load, leading to potential bottlenecks in terms of both time and memory.
Therefore we split this task into
two stages to reduce computation cost. In the first stage, we jointly train the MLP decoder and the triplanes on a smaller subset of avatars. In the second stage, we fix the decoder's weights and fine-tune the triplane of each avatar independently, which can be performed in parallel.
The first stage of getting an accurate yet generalizable MLP decoder is critical in order to generate details for all avatars including those generated ones. 
 
The decoder $\mathcal{F}$ parameterized by $\bm{\omega}$ learns identity-agnostic priors from a set of $S$ avatars. Each avatar is represented as $N$ multi-view renderings $\mathcal{D}^s=\left\{\left(\mathbf{I}^{s, n}, c^{s, n}\right)\right\}_{n=1}^{N}$, where $\mathbf{I}^{s, n}$ is the RGBA image and $c^{s, n}$ is the corresponding camera configuration. 
There are two critical issues overlooked in previous work~\cite{wang2023rodin}.
1) During training, batches typically comprise rays sampled from a single avatar due to limited GPU memory. Consequently, each avatar is treated as an independent task. To minimize data transfer between CPU and GPU, each avatar undergoes multiple training iterations until convergence before being replaced by the next one. As a result, the MLP may gradually forget the previously learned knowledge and become overly adapted to the current avatar. This is known as catastrophic forgetting in continual learning. \cref{fig:catastrophic_forgetting} visualizes the phenomenon. 
2) Additionally, the approach encounters training instabilities when switching between avatars due to the substantial gaps between them. This usually results in an under-fitted MLP that is incapable of decoding high-frequency details, even with the second triplane finetuning stage.
\cref{fig:fitting_ablation} compares the differences. In the following, we formally describe the task-replay strategy and weight consolidation regularizer to address the above challenges.

\noindent\textbf{Task replay.}
The core idea is to switch avatars more frequently, allowing each avatar to be seen periodically for multiple times and preventing the decoder from over-fitting to a single avatar. As a result, the decoder is exposed to triplanes of a variety of avatars, ensuring its generalization ability. To implement this strategy, each avatar is trained multiple times, with each time fitting for a shorter time without requiring convergence by tuning ``outer\_loop\_iteration" and 
 ``inner\_loop\_iteration'' in~\cref{alg:triplane_fitting},
while the naive method trains each avatar only once, and fits it to convergence before switching to the next avatar.
The task replay strategy proves effective in mitigating the risk of over-fitting to a single avatar, which may otherwise result in poor generalization performance. 

\begin{algorithm}[tb]
    \caption{The First Stage of Triplane Fitting}\label{alg:triplane_fitting}
    \begin{algorithmic}[1]
    \Require dataset $\left\{\mathcal{D}^s\right\}_{s=1}^{S}$, triplane parameters $\left\{\bm{x}^s\right\}_{s=1}^{S}$, shared MLP $\mathcal{F}$ parameters $\bm{\omega}$, IWC states $\left\{\bm{\Omega}^s=\mathbf{0}, \bm{\omega}^{s,*} \right\}_{s=1}^{S}$, learning rate $\alpha, \beta$.
    \Repeat
        \State Sample from $\mathcal{D}^s$ and load triplane $\bm{x}^s$ from disk
        \Repeat
            \State Sample rays $r \sim \mathcal{D}^s$
            \State $\mathcal{L}(\bm{x}^s, \bm{\omega}; \bm{\Omega}^s, \bm{\omega}^{s,*}) \leftarrow \mathcal{L}_{\mathcal{D}^s}  + \mathcal{L}_{\text{IWC}}$
            \State $\bm{x}^s \leftarrow  \bm{x}^s -\alpha \nabla_{\bm{x}^s}\mathcal{L}(\bm{x}^s, \bm{\omega}; \bm{\Omega}^s, \bm{\omega}^{s,*})$
            \State $\bm{\omega} \leftarrow  \bm{\omega} -\beta \nabla_{\bm{\omega}}\mathcal{L}(\bm{x}^s, \bm{\omega}; \bm{\Omega}^s, \bm{\omega}^{s,*})$
        \Until{inner\_loop\_iteration} 
        \State $\left\{\bm{\Omega}^s, \bm{\omega}^{s,*} \right\} \leftarrow \left\{(\nabla_{\bm{\omega}}\mathcal{L}_{\mathcal{D}^s})^2, \bm{\omega} \right\}$ \Comment{update state}
    \Until{outer\_loop\_iteration}
 \end{algorithmic}
 \end{algorithm}

\noindent\textbf{Weight consolidation.} Learning the triplanes and the decoder also suffers from instabilities caused by the occasionally occurring large gradients when we switch between the avatars. To mitigate this issue, we introduce an Identity-aware Weight Consolidation (IWC) regularizer, a technique that stabilizes learning by consolidating knowledge and reducing drastic shifts in the learning landscape. Specifically, it uses the elastic weight consolidation regularizer~\cite{kirkpatrick2017overcoming} to prevent the most important weights of this avatar from deviating far from its consolidated values during training:
\begin{equation}
\begin{aligned}
\mathcal{L}_{\text{Fitting}} &= \mathcal{L}_{\mathcal{D}^s}+\mathcal{D}_{\text{IWC}} \\
&=\mathcal{L}_{\mathcal{D}^s}+\frac{\lambda}{2} \sum_i \bm{\Omega}^s_i\left(\bm{\omega}_i-\bm{\omega}^{s, *}_i\right)^2,
\end{aligned}
\end{equation}
where $\mathcal{L}_{\mathcal{D}_s}$ is the conventional rendering loss between rendered images and ground-truth images, $\bm{\omega}^{s, *}_i$ is the $i^{\text{th}}$ MLP weight obtained when training the avatar in the previous epoch, and $\bm{\Omega}^s_i$ is the importance of $i^{\text{th}}$ weight, calculated by the $i^{\text{th}}$ diagonal element of the Fisher Information Matrix $\bm{\Omega}^{s}$. Since the Fisher Information Matrix is equivalent to the second derivatives of the loss near a minimum and it can be computed from first-order derivatives, computing the IWC loss is efficient.

\begin{figure}[t]
  \centering
  \includegraphics[width=0.65\textwidth]{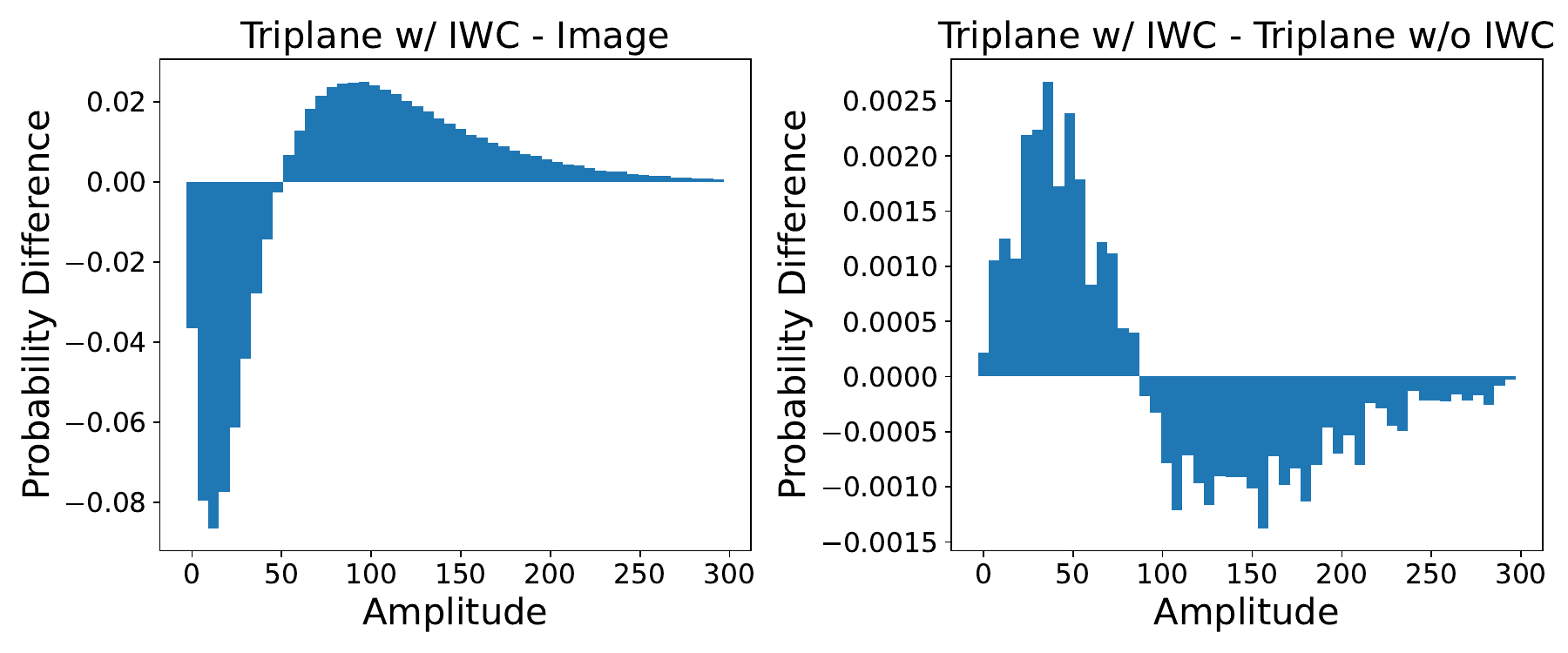}
  \caption{\textbf{Frequency difference between two sources}. \textbf{Left:} Triplanes have more high-frequency components than images. \textbf{Right:} Triplanes learned with our proposed IWC have fewer high-frequency components. }
  \label{fig:freq_analysis}
\end{figure}

We find that our approach not only improves the details but also reduces the high-frequency components in the triplanes, making them easier to learn by diffusion models~\cite{li2023generative,ma2022accelerating}. See  \cref{fig:freq_analysis} for comparison. We also use TV loss and $L_2$ regularization on the triplanes to promote smoothness as in the prior work~\cite{wang2023rodin}. 

\subsection{Triplane Diffusion}
\label{sec:3d-diffusion}
We train a cascaded diffusion model to generate high-resolution triplanes. It consists of a \emph{base model}, which generates a low-resolution triplane conditioned on a portrait image $\mathbf{I}_{\text{front}}$, and an \emph{upsample model}, which subsequently generates a high-resolution triplane. During training, we first obtain the destructed triplane $\bm{x}_t$ according to $\bm{x}_t\coloneqq\alpha_t\bm{x}_0+\sigma_t\bm{\epsilon}$, where $\alpha_t, \sigma_t$ define the noise schedule which determines the destruction strength of the signal, and $t$ is a continuous number ranged from 0 to 1. We train the base and upsample diffusion models in two separate steps. For base diffusion training, we parameterize the diffusion model $\hat{\bm{\epsilon}}_\theta$ to predict the noise~\cite{ho2020denoising} added to the low-resolution triplane $\bm{x}_t^{\text{LR}}$:
\begin{equation}
    \mathcal{L}_{\text {simple}}^{\text{LR}}=\mathbb{E}_{t, \bm{x}_0^{\text{LR}}, \bm{\epsilon}}\left[\left\|\hat{\bm{\epsilon}}_\theta\left(\alpha_t \bm{x}_0^{\text{LR}}+\sigma_t \bm{\epsilon}, t, \mathbf{I}_{\text{front}}\right)-\bm{\epsilon}\right\|_2^2\right] .
\end{equation}

Despite the $\mathcal{L}_{\text{simple}}^{\text{LR}}$, we also use the variational lower bound to optimize the negative log-likelihood of estimated distribution by incorporating $\mathcal{L}_{\text{vlb}}^{\text{LR}}$ following~\cite{nichol2021improved} for higher generation quality. Our upsample diffusion model is learned to enhance the high-fidelity details of the low-resolution triplane, which is conditioned on both $\bm{x}_0^{\text{LR}}$ and $\mathbf{I}_{\text{front}}$. We directly parameterize the model to predict the noiseless input $\bm{x}_0^{\text{HR}}$:
\begin{equation}
    \mathcal{L}_{\text {simple}}^{\text{HR}}=\mathbb{E}_{t, \bm{x}_0^{\text{HR}}, \bm{\epsilon}}\left[\left\|\hat{\bm{x}_\theta}\left(\alpha_t \bm{x}_0^{\text{HR}}+\sigma_t \bm{\epsilon}, t, \mathbf{I}_{\text{front}}, \bm{x}_0^{\text{LR}} \right)-\bm{x}_0^{\text{HR}}\right\|_2^2\right] .
\end{equation}

To ensure the rendered images of our generated triplanes have compelling visual quality, we also adopt image-level supervision when training the upsample model inspired by previous works~\cite{wang2023rodin}. Specifically, we penalize the discrepancy between the rendered image patch $\hat{\mathbf{I}}_{\text{patch}}$ from predicted triplane $\hat{\bm{x}_\theta}$ and the ground truth image patch $\mathbf{I}_{\text{patch}}$:
\begin{equation}
\begin{split}
    \mathcal{L}_{\text {Image}}^{\text{HR}} &= \mathcal{L}_{\text {Pixel}} +\mathcal{L}_{\text {Perc}} \\
    &= \mathbb{E}_{t, \hat{\mathbf{I}}_{\text{patch}}} (\sum_l\left\|\Psi^l(\hat{\mathbf{I}}_{\text{patch}})-\Psi^l(\mathbf{I}_{\text{patch}})\right\|_2^2) \\
    &+\mathbb{E}_{t, \hat{\mathbf{I}}_{\text{patch}}}(\left\|\hat{\mathbf{I}}_{\text{patch}}-\mathbf{I}_{\text{patch}}\right\|_2^2),
\end{split}
\end{equation}
where $\Psi^l$ denotes the multi-scale feature extracted using a pre-trained VGG~\cite{simonyan2014very}. Ensuring the diffusion model scales well with the increased resolution is crucial. As high-resolution triplanes involve a large number of parameters, processing such vast amounts of information efficiently can be a substantial hurdle. Next, we introduce improvements to our diffusion model.

\begin{figure}[t]
    \centering
    \footnotesize
    \setlength\tabcolsep{1pt}
    {
    \renewcommand{\arraystretch}{0.6}
    \begin{tabular}{@{}cc@{}}
         \includegraphics[width=0.25\columnwidth]{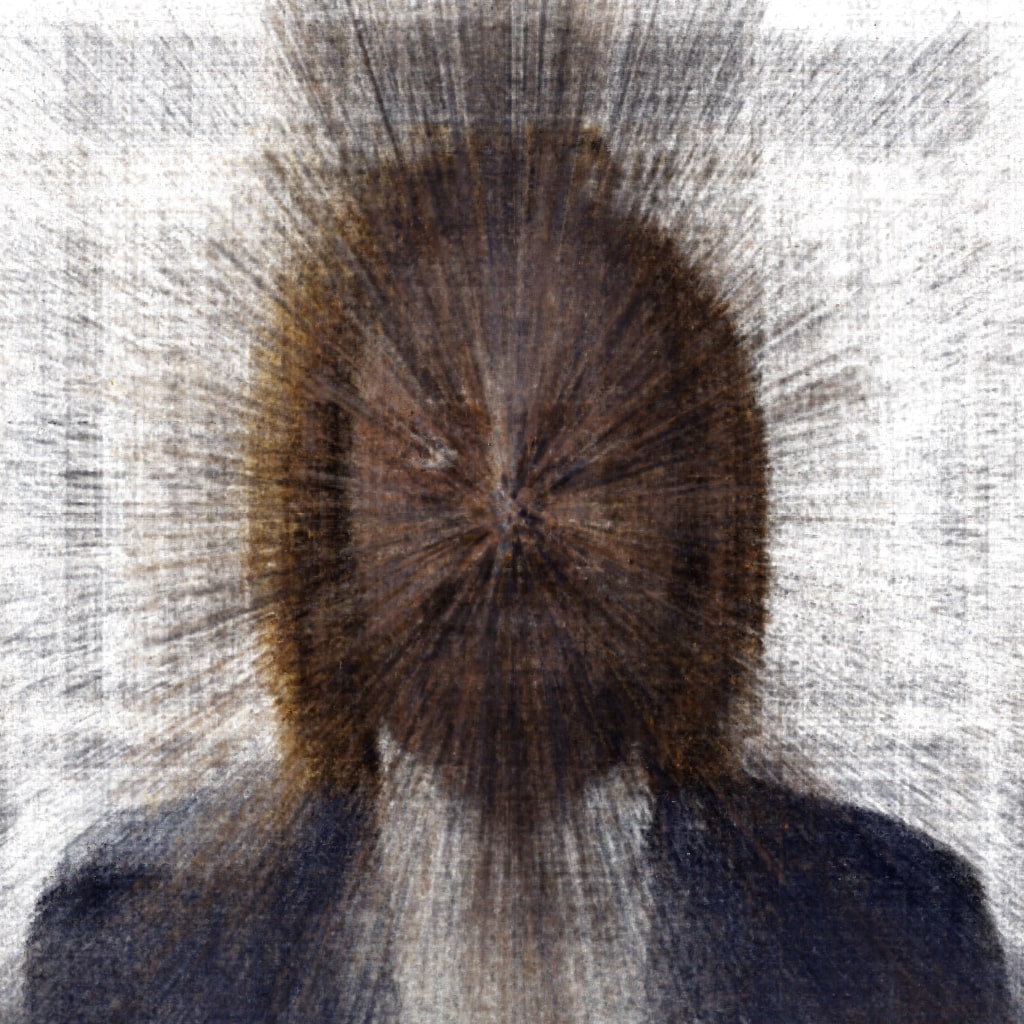} & \includegraphics[width=0.25\columnwidth]{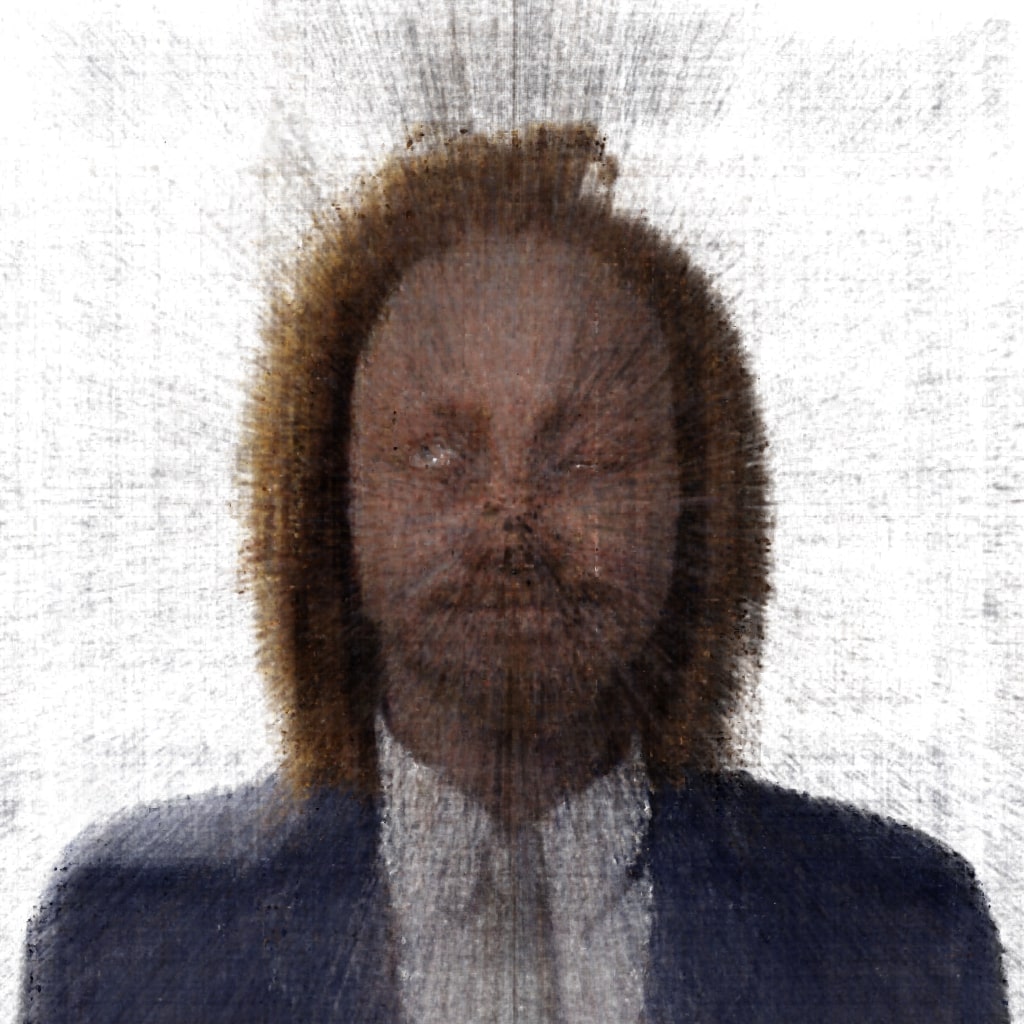}\\
         & \\
         8 channels & 32 channels
    \end{tabular}
    }
    \caption{Rendered images from triplanes with 8 and 32 channels, respectively. The triplanes are destructed with the same noise level ($\text{logSNR}(t) = 0.57$). The 32-channel triplane has larger redundancy so it is less destructed.}
    \label{fig:vis_noise_channel_relation}
\end{figure}

\noindent\textbf{Multi-scale image feature conditioning.} It is difficult to hallucinate detailed 3D avatars from scratch. So, we propose to supply ample details from the portrait image to alleviate the difficulty. Previous works such as~\cite{wang2023rodin} compute a global semantic token using the pre-trained CLIP image encoder~\cite{radford2021learning}, which results in a substantial loss of detailed information.

To fully utilize the information of the portrait, we compute a multi-scale feature representation using a pre-trained Variational Autoencoder~\cite{podell2023sdxl}. Since the VAE is trained to accurately reconstruct the input images, the low-level visual details are well preserved in the latent features. Formally, we denote the VAE encoder as $E$, and the frontal portrait image $\mathbf{I}_{\text{front}} \in \mathbb{R}^{H\times W\times3}$. We compute the conditional signals for our diffusion model by:
\begin{equation}
\{\bm{y}_1, \bm{y}_2, \bm{y}_3\} = E(\mathbf{I}_{\text{front}}),
\end{equation}
where $\bm{y}_1\in \mathbb{R}^{\frac{H}{2}\times\frac{W}{2}\times C}, \bm{y}_2\in \mathbb{R}^{\frac{H}{4}\times\frac{W}{4}\times2C},$  and $\bm{y}_3\in \mathbb{R}^{\frac{H}{8}\times\frac{W}{8}\times4C}$ denote the multi-scale spatial features, and $C$ is the based channel dimension.

Since the 2D portrait is not aligned with the 3D triplane, directly harmonizing the conditional signals with diffusion U-Net features via  concatenation~\cite{nichol2021glide,ho2022cascaded} or addition~\cite{zhang2023adding} is problematic. Therefore, we elect to perform cross-attention~\cite{vaswani2017attention}. In this way, the network is learned to automatically discover the spatial correspondence between triplanes and 2D images.

Formally, let  $\bm{x}_i$ be the feature maps of the $i$-th attention resolution in U-Net, we inject $\bm{y}_i$ by

\begin{equation}
\begin{aligned}
    \{\bm{y}_i^{k}\}_{k=1}^K &= \text{PatchPartition}(\bm{y}_i), \\
    \bm{x}_{i}^{\prime} &= \text{CrossAttn}(\bm{x}_i; \{\bm{y}_i^{k}\}_{k=1}^K),
\end{aligned}
\end{equation}
where $K$ and $\bm{x}_{i}^{\prime}$ are the number of patches and the output features, respectively. For the base model, the conditions are injected into both encoder and decoder layers that have the same resolutions. For the upsample model, the conditions are only injected into the middle latent features to reduce computation costs.

\definecolor{myblue}{RGB}{0, 28, 127}
\definecolor{mygreen}{RGB}{18, 113, 28}
\definecolor{myorange}{RGB}{177, 63, 13}
\begin{figure}[t]
    \centering  
    \begin{subfigure}[h]{0.47\linewidth}  
        \centering  
        \begin{tikzpicture}[scale=0.55] 
        \begin{axis}[  
            title=Base Diffusion,
            label style={font=\small},  
            /pgfplots/xtick = {0.0, 0.2, 0.4, 0.6, 0.8, 1.0}, 
            restrict y to domain=-20:10,
            legend entries = {Default Linear, Ours},  
            legend style = {draw=none, font=\small},  
            ]  
            \addplot [no markers, line width=3pt, dashed, myblue] table {data/log_snr_base_linear.dat};    
            \addplot [no markers, line width=3pt, solid, orange] table {data/log_snr_base_cosine_light.dat};
        \end{axis}  
        \end{tikzpicture}
    \end{subfigure}  
    \hfill  
    \begin{subfigure}[h]{0.47\linewidth}  
        \centering  
        \begin{tikzpicture}[scale=0.55]  
         \begin{axis}[  
            title=Upsample Diffusion,
            label style={font=\small},  
            /pgfplots/xtick = {0.0, 0.2, 0.4, 0.6, 0.8, 1.0}, 
            restrict y to domain=-40:5,
            legend entries = {Default Linear, Ours},  
            legend style = {draw=none, font=\small},  
            ]  
            \addplot [no markers, line width=3pt, dashed, myblue] table {data/log_snr_sr_linear.dat};    
            \addplot [no markers, line width=3pt, solid, orange] table {data/log_snr_sr_sigmoid.dat};
        \end{axis} 
        \end{tikzpicture}    
    \end{subfigure}  
    \caption{\textbf{Optimized noise schedule.} LogSNR comparison between the default and our optimized noise schedules for the base (left) and upsample (right) networks.}  
    \label{fig:adjusted_noise_schedule}  
\end{figure}
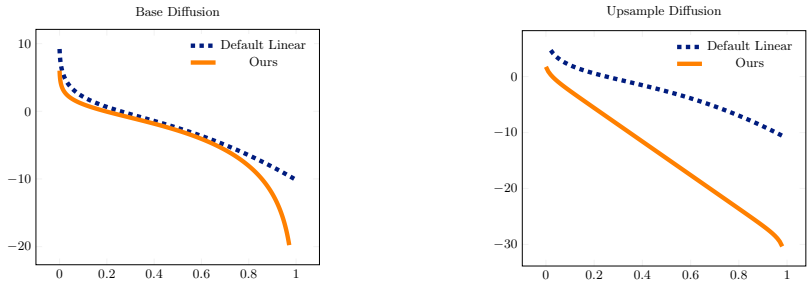  

\noindent\textbf{Optimized noise schedule.} In high-resolution image generation, some work~\cite{chen2023importance,hoogeboom2023simple,lin2023common,gu2022f} find that using a stronger noise is critical to retain the learning difficulty for images with higher resolutions. We similarly study the optimal noise schedule for triplanes in 3D generation. Compared to images, triplanes have large spatial resolutions and channel numbers, and even long-range dependencies introduced by 3D correspondence. All these factors increase the redundancy in the representation. 
~\cref{fig:vis_noise_channel_relation} shows an example. We fit two triplanes with the same spatial resolution but different channel dimensions (8 vs. 32), and add the same level of noises ($\text{SNR}(t) = \alpha_t^2/\sigma_t^2$) to them, respectively. However, we can see that the rendered images are destructed differently. The triplane with fewer channels suffers from more disruption compared with the one with larger channels. 

Considering the larger redundancy in triplanes, we propose to apply a stronger noise to fully destruct triplanes to prevent the model from under-training. To be more specific, we utilize the adjusted cosine noise schedule~\cite{chen2023importance,nichol2021improved} in our base diffusion and a much stronger sigmoid noise schedule~\cite{chen2023importance,jabri2022scalable} in the upsample diffusion stage. The adjusted noise schedules are shown in~\cref{fig:adjusted_noise_schedule}. They are much stronger than the default linear noise scheduling~\cite{ho2020denoising} designed for low-resolution images \eg $32\times32, 64\times64$.


\section{Experiments}

\subsection{Dataset and Metrics}
We conduct experiments on 46K  avatars created from Blender~\cite{wood2021fake}. For each avatar, we uniformly render 300 multi-view images at the resolution of $1024 \times 1024$. We evaluate our model's conditional and unconditional generation capability, respectively. For image-conditioned generation, we compute numerical results using the common metrics of FID~\cite{heusel2017gans}, LPIPS~\cite{zhang2018unreasonable}, and Structural Similarity Index Measure (SSIM) between 5K rendered multi-view images from generated avatars with ground-truth images. For unconditional generation, we report FID of 5K rendered images from randomly sampled avatars. We also measure cross-view consistency by fitting a NeuS model~\cite{wang2021neus} from the multi-view renderings of the generated avatars.

\begin{figure*}[t]
  \centering
  \begin{overpic}
    [width=0.95\linewidth]{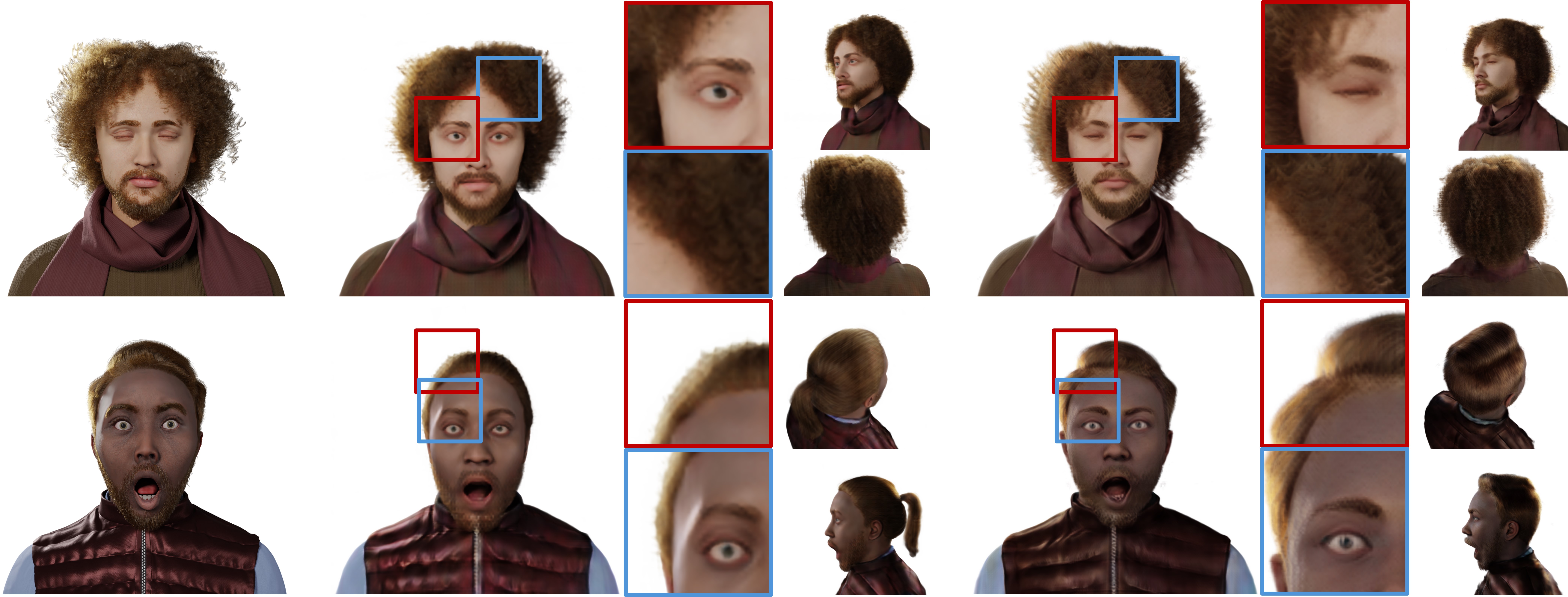}
    \put(5,-3){\small Reference}
    \put(40,-3){\small Rodin~\cite{wang2023rodin}}
    \put(73,-3){\small \textbf{Our RodinHD}}
  \end{overpic}
  \caption{Avatars generated by different methods.}
  \label{fig:cond_gen}
\end{figure*}

\begin{table}[t]
    \caption{Quantitative results of conditional avatar generation.}
  \centering
  \small
  \setlength\tabcolsep{5pt}
  \begin{tabular}{lccc}
  \toprule  
  Models & FID$\downarrow$ & LPIPS$\downarrow$ & SSIM$\uparrow$ \\
  \midrule
  Rodin~\cite{wang2023rodin} & 33.20 & 0.323 & 0.758 \\
  \cellcolor{gray}\textbf{Ours} & \cellcolor{gray}\textbf{26.49}& \cellcolor{gray}\textbf{0.299} & \cellcolor{gray}\textbf{0.765} \\
  \bottomrule
  \end{tabular}
  
  \label{table:cond_quantitative_comparison}
  \end{table} 
\subsection{Implementation Details}
The triplane resolution is $512 \times 512$ and its channel number is 32. We randomly select 64 avatars for jointly training the triplanes and the NeRF decoder for two days on 8 Tesla V100 GPUs, with task replay and IWC regularization applied. Then, we fit the triplanes for all avatars independently, with each taking about $15$ minutes. We adopt the U-Net architecture~\cite{dhariwal2021diffusion} as diffusion backbone.
For the configuration of conditional feature injection, please refer to supplementary material. 
We utilize conditional augmentation~\cite{ho2022cascaded,wang2023rodin} when training the upsample diffusion to reduce the domain gap between training and inference. Both our base and upsample models are trained on 32 Tesla V100 (32G) GPUs, with batch sizes 96 and 32 respectively. We randomly drop the conditional features with $20\%$ probability, which enables us to perform classifier-free guidance and unconditional generation.

\subsection{Main Results}
\begin{figure*}[t]
  \centering
  \small
  \begin{tabular}{@{}c@{\hspace{15mm}}c@{\hspace{15mm}}c@{}} 
    \includegraphics[width=0.25\linewidth]{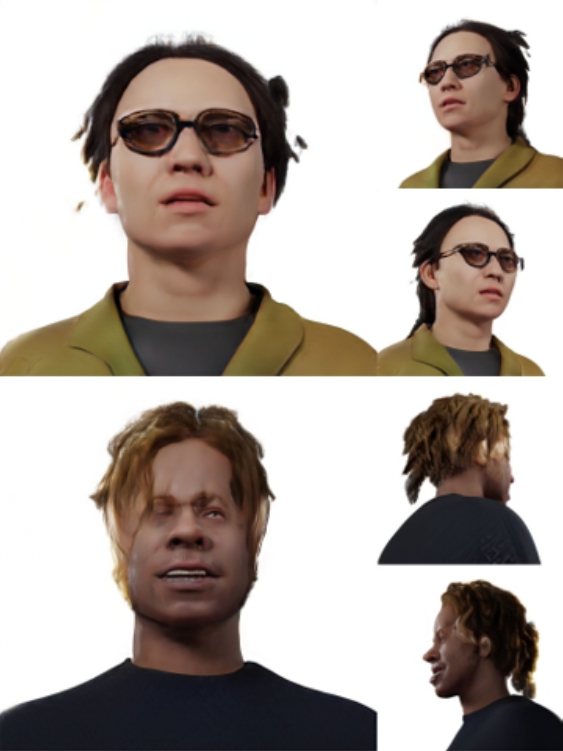}  &    
    \includegraphics[width=0.25\linewidth]{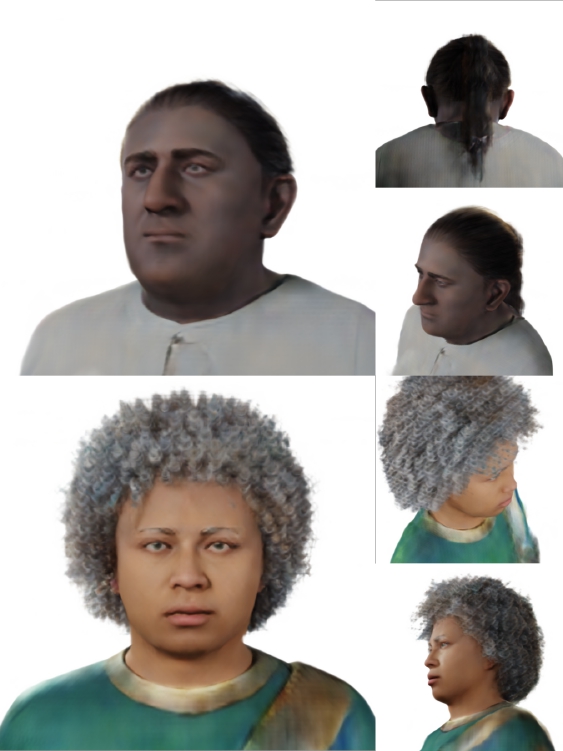} &
    \includegraphics[width=0.25\linewidth]{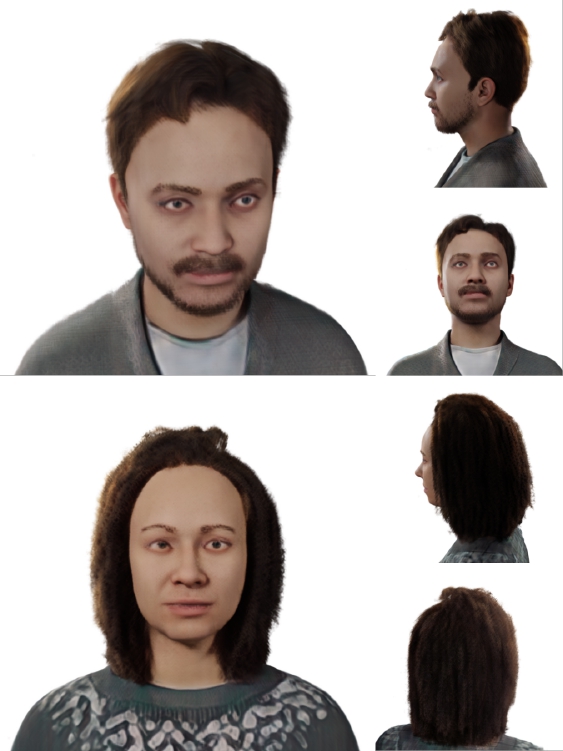} \\ 
    EG3D~\cite{chan2022efficient} & Rodin~\cite{wang2023rodin} & \textbf{Our RodinHD}\\
  \end{tabular}
  \caption{Qualitative results of unconditional avatar generation.}
  \label{fig:uncond_comparison}
\end{figure*}

\begin{table}[t]
  \caption{Quantitative results of unconditional avatar generation. \textbf{The subscript $^*$ indicates that 2D refinement is applied to the rendered images}. }
  \centering
  \footnotesize
  \begin{tabular}{l@{\hspace{3mm}}c@{\hspace{3mm}}c@{\hspace{3mm}}c@{\hspace{3mm}}c@{\hspace{3mm}}c@{\hspace{3mm}}c}
  \toprule  
   & Pi-GAN & GIRAFFE & EG3D$^{*}$ & Rodin$^{*}$ & Rodin & \textbf{Ours}\\
  \midrule
  FID $\downarrow$ & 78.3 & 64.6 & 40.5 & 30.29 & 45.70 & 32.62 \\
  \bottomrule
  \end{tabular}
  \label{table:uncond_gen}
  \end{table} 

\noindent\textbf{Conditional generation.} We compare our approach with Rodin~\cite{wang2023rodin}. We report the results of Rodin without 2D refinement for fair comparison. As shown in~\cref{fig:cond_gen}, our model captures the detailed appearance and vivid expression of the given portrait, benefiting from high-quality triplane fitting and strong guidance of the proposed multi-scale image feature conditioning. On the contrary, Rodin fails to synthesize details of avatars, \eg, hairstyles and closed eyes. Moreover, our model enables us to directly render compelling images in high resolution without any 2D refinement, demonstrating the strong capacity of the proposed model. Furthermore, we achieve the best results across all metrics as shown in~\cref{table:cond_quantitative_comparison}, demonstrating the superiority of our model.

\noindent\textbf{Unconditional generation.} We compare our method with the state-of-the-art methods, including both 3D-aware GANs~\cite{chan2021pi,niemeyer2021giraffe,chan2022efficient} and 3D-diffusion-based Rodin~\cite{wang2023rodin}. We present the results of Rodin before and after the 2D refiner (denoted as Rodin$^*$), respectively. 
~\cref{table:uncond_gen} shows the results. Our approach achieves significantly better results than the methods except Rodin$^*$ which uses a 2D refiner. However, as we will discuss in the following, our method achieves notably better 3D consistency. We also provide visual comparison in~\cref{fig:uncond_comparison}. While prior arts tend to generate blur rendering, our model is able to provide complex and diverse avatars with rich details, \eg, hair and clothing.

\begin{table}[t]
  \caption{3D consistency measured by the fitting quality of NeuS~\cite{wang2021neus}.}
  \centering
  \setlength\tabcolsep{5pt}
  \small
  \begin{tabular}{lccc}
  \toprule  
  Models &  PSNR$\uparrow$ & LPIPS$\downarrow$ & SSIM$\uparrow$\\
  \midrule
  Rodin$^{*}$~\cite{wang2023rodin} & 31.73 & 0.051 & 0.973 \\
  \cellcolor{gray}\textbf{Ours} & \cellcolor{gray}\textbf{35.46} & \cellcolor{gray}\textbf{0.041} & \cellcolor{gray}\textbf{0.975} \\
  \bottomrule
  \end{tabular}
  \label{table:3d_consistency_comparison}
  \end{table}

\begin{figure}[t]
  \centering
  \small
  \setlength\tabcolsep{15pt}
  \begin{tabular}{ccc} 
    \includegraphics[width=0.2\linewidth]{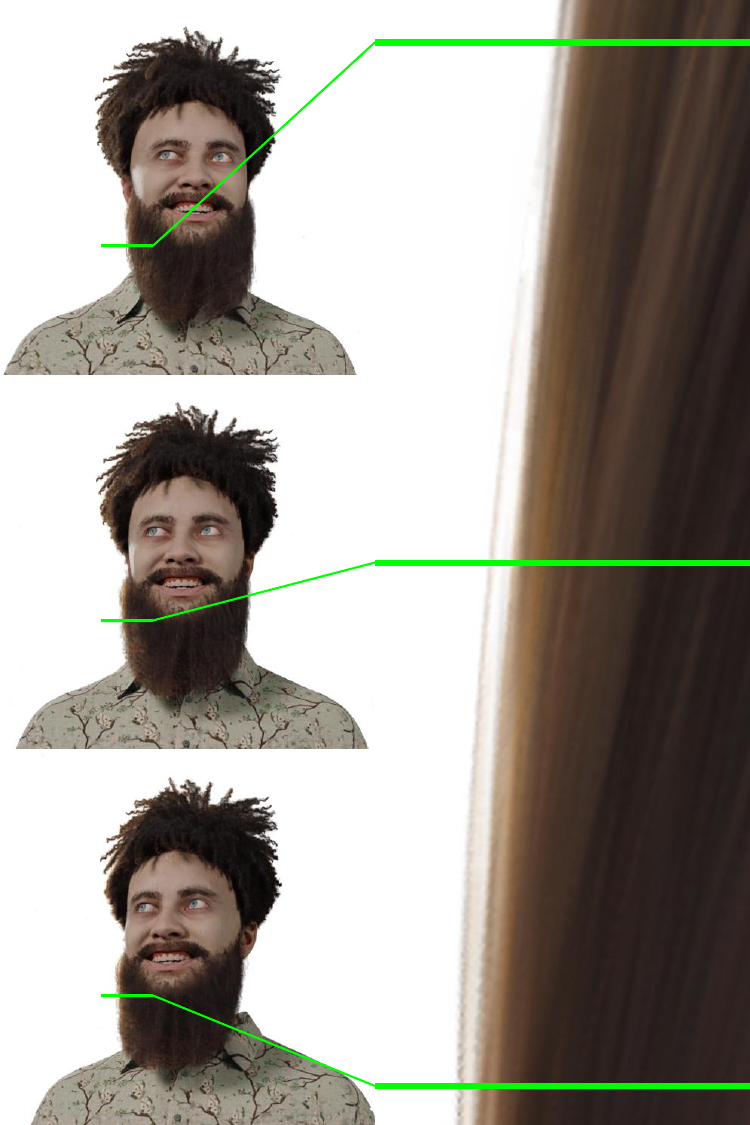}  &
    \includegraphics[width=0.2\linewidth]{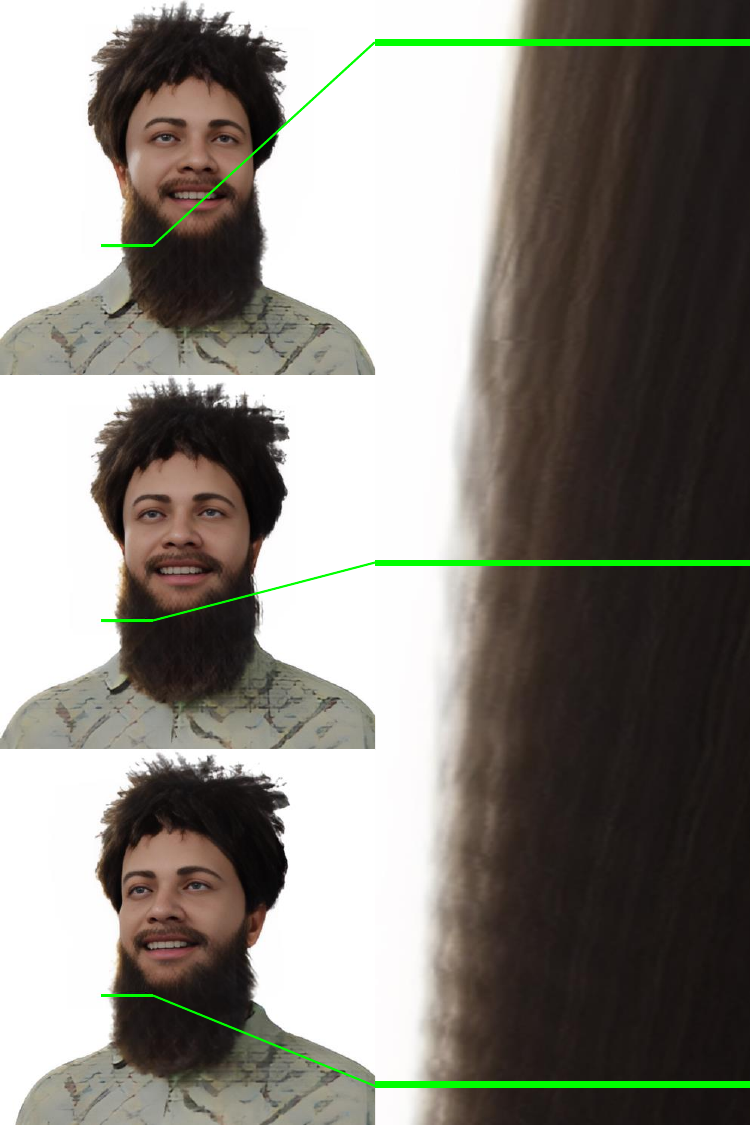} &
    \includegraphics[width=0.2\linewidth]{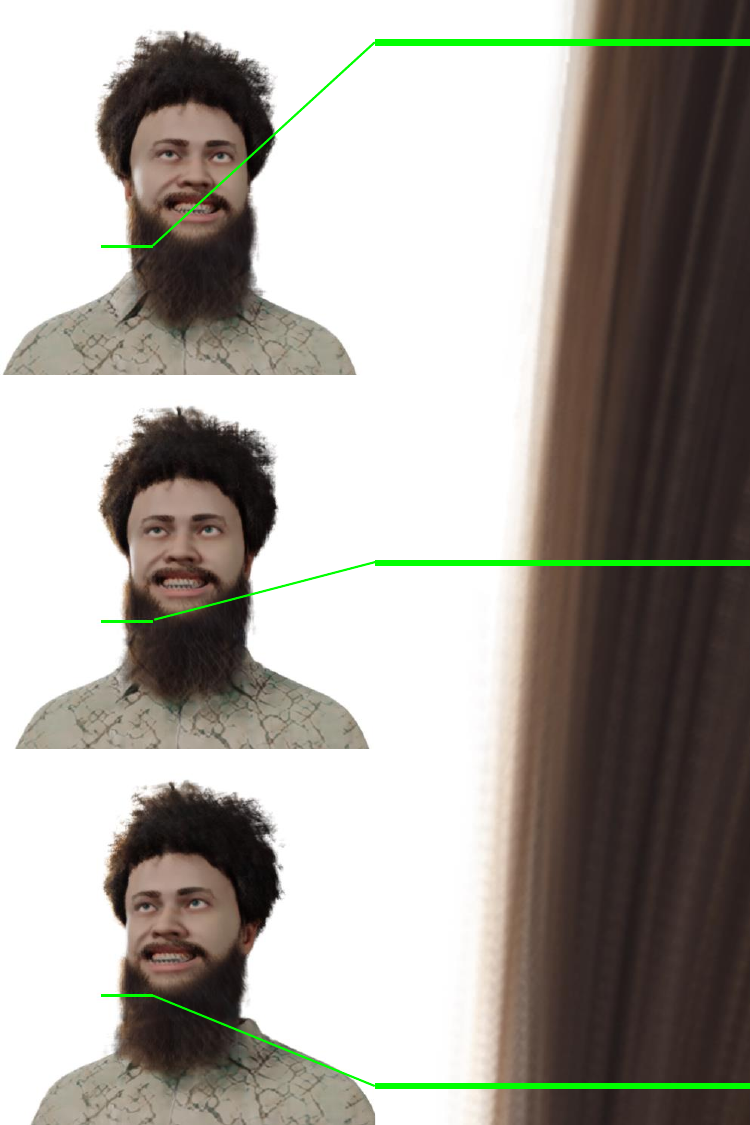}  \\
    GT & Rodin$^{*}$~\cite{wang2023rodin} & \textbf{Our RodinHD}
    \end{tabular}
  \caption{Visual comparison of 3D consistency akin to the Epipolar Line Images~\cite{bolles1987epipolar}. We visualize the stacked texture of a fixed line as the camera smoothly rotates horizontally. Our model yields smooth and natural texture as GT, whereas a clear flickering pattern is shown in Rodin's results, indicating the 3D inconsistency using 2D refinement.}
  \label{fig:3d_consistency}
\end{figure}

\noindent\textbf{3D consistency.} We evaluate the 3D consistency of the generated results by visualizing the spatiotemporal textures following~\cite{xiang2022gram}. We include Rodin$^{*}$, ours and ground-truth triplane renderings for comparison in~\cref{fig:3d_consistency}. As the camera moving smoothly, the texture of the fixed horizontal line is expected to be smooth and nature as GT triplane renderings. However, the spatiotemporal textures of Rodin$^{*}$ suffer from obvious flickering, which suggests inconsistency across views (see video in supplementary for more intuitive comparison). In contrast, our model produces smooth and natural results as GT, demonstrating the strong consistency of the proposed model. Moreover, the renderings of Rodin$^{*}$ fail to maintain the skin tone of the input while our model provides more faithful results. We also quantitatively measure the 3D consistency by training a multi-view reconstruction method NeuS~\cite{wang2021neus} on 300 rendering images of Rodin$^{*}$ and ours respectively. The results of our model achieve significantly better metrics thanks to the 3D consistency, whereas Rodin$^{*}$ obtains worse results since the 2D refiner breaks the 3D consistency.

\begin{table}[t]
    \caption{Ablation study of the proposed components.} 
    \scriptsize
    \begin{subtable}{.45\linewidth}
      \centering
        \begin{tabular}{lccc}
            \toprule
            Training strategy & PSNR$\uparrow$ & LPIPS$\downarrow$ & SSIM$\uparrow$ \\
            \midrule
            A. Rodin (256) & 30.31 & 0.131 & 0.862 \\
            B. Rodin (512) & 30.56 & 0.129 & 0.863 \\
            C. + Task replay & 30.93 & 0.106 & 0.890 \\
            D. + Weight decay & 31.00 & 0.094 & 0.909 \\
            E. + Consolidation & \textbf{31.45} & \textbf{0.086} & \textbf{0.911} \\
            \bottomrule
        \end{tabular}
        \caption{}
    \end{subtable}%
    \begin{subtable}{.5\linewidth}
      \centering
        \begin{tabular}{lccc}
            \toprule
            Model configuration & FID$\downarrow$ & LPIPS$\downarrow$ & SSIM$\uparrow$ \\
            \midrule
            A. Rodin (256) & 33.20 & 0.323 & 0.758 \\
            B. + Naively scale to 512 & 35.84 & 0.324 & 0.740 \\
            C. + Strong noise schedule & 27.74 & 0.318 & 0.745 \\
            D. + Single-scale  cond. & 27.13 & 0.301 & 0.763 \\
            E. + Multi-scale cond. & \textbf{26.49} & \textbf{0.299} & \textbf{0.765} \\
            \bottomrule
        \end{tabular}
        \caption{}
    \end{subtable} 
    \label{table:ablation-fitting-diffusion}
\end{table}

\begin{figure}[t]
  \centering
  \small
  \setlength\tabcolsep{3pt}
  \begin{tabular}{cccc} 
    \includegraphics[width=0.2\linewidth]{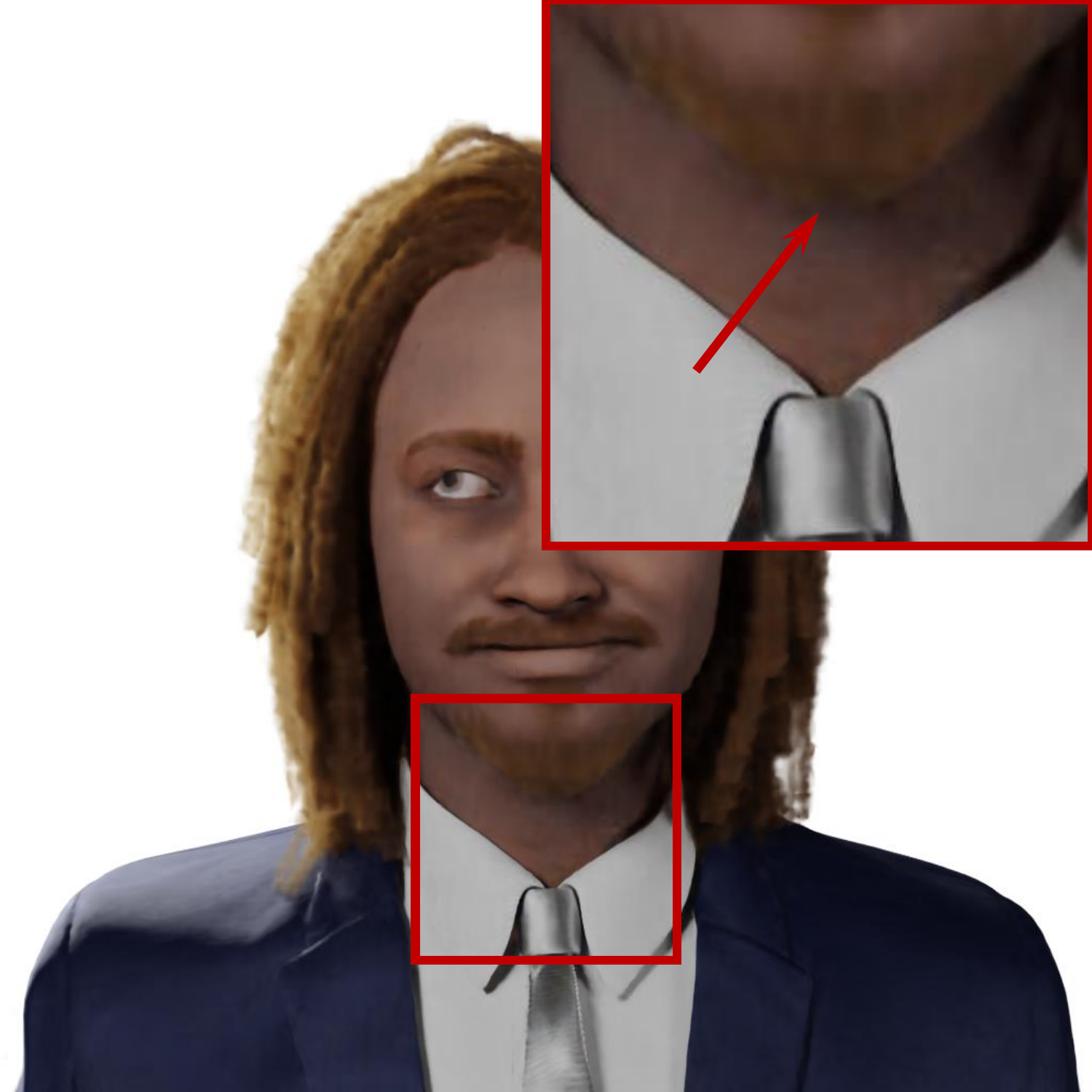}  &
    \includegraphics[width=0.2\linewidth]{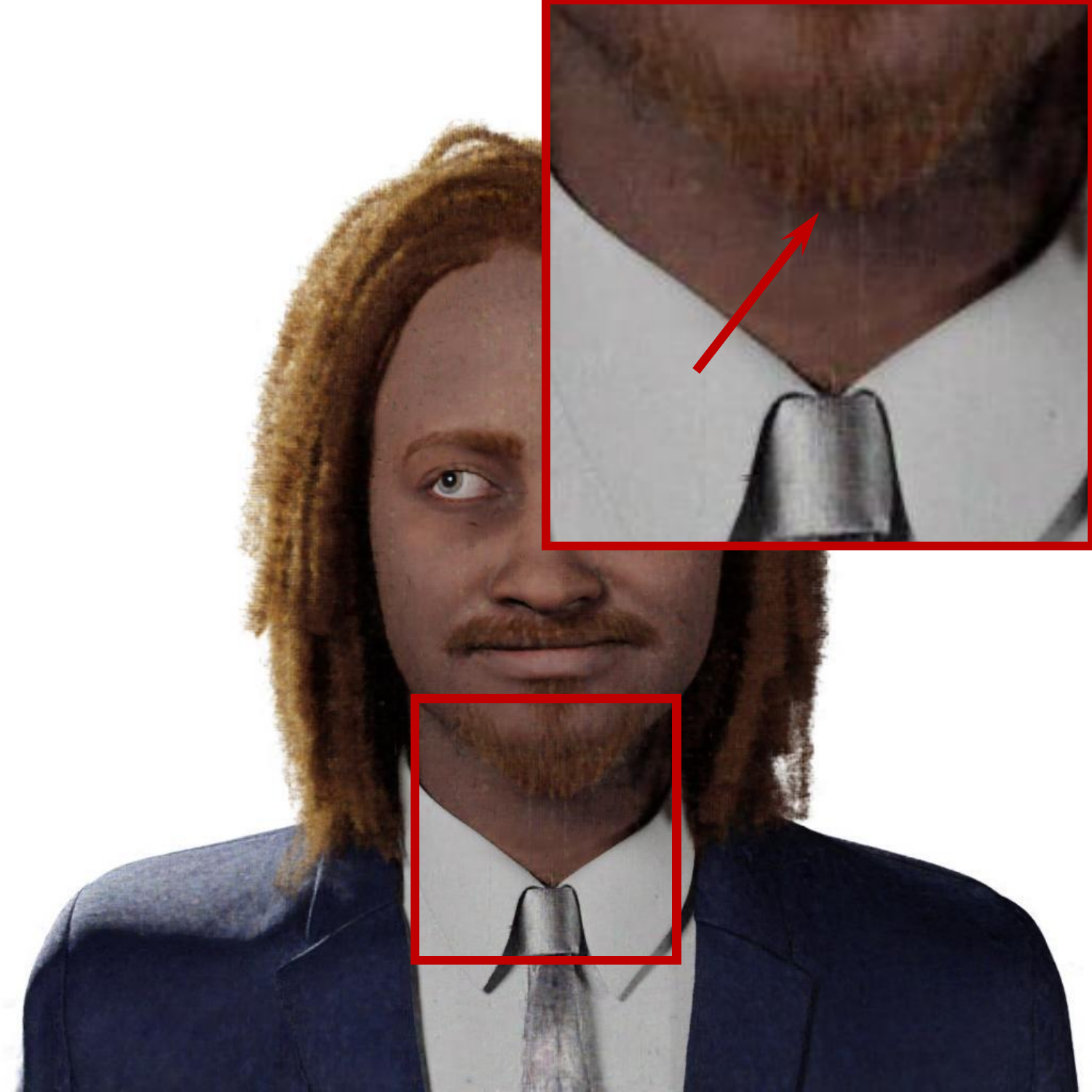} &
    \includegraphics[width=0.2\linewidth]{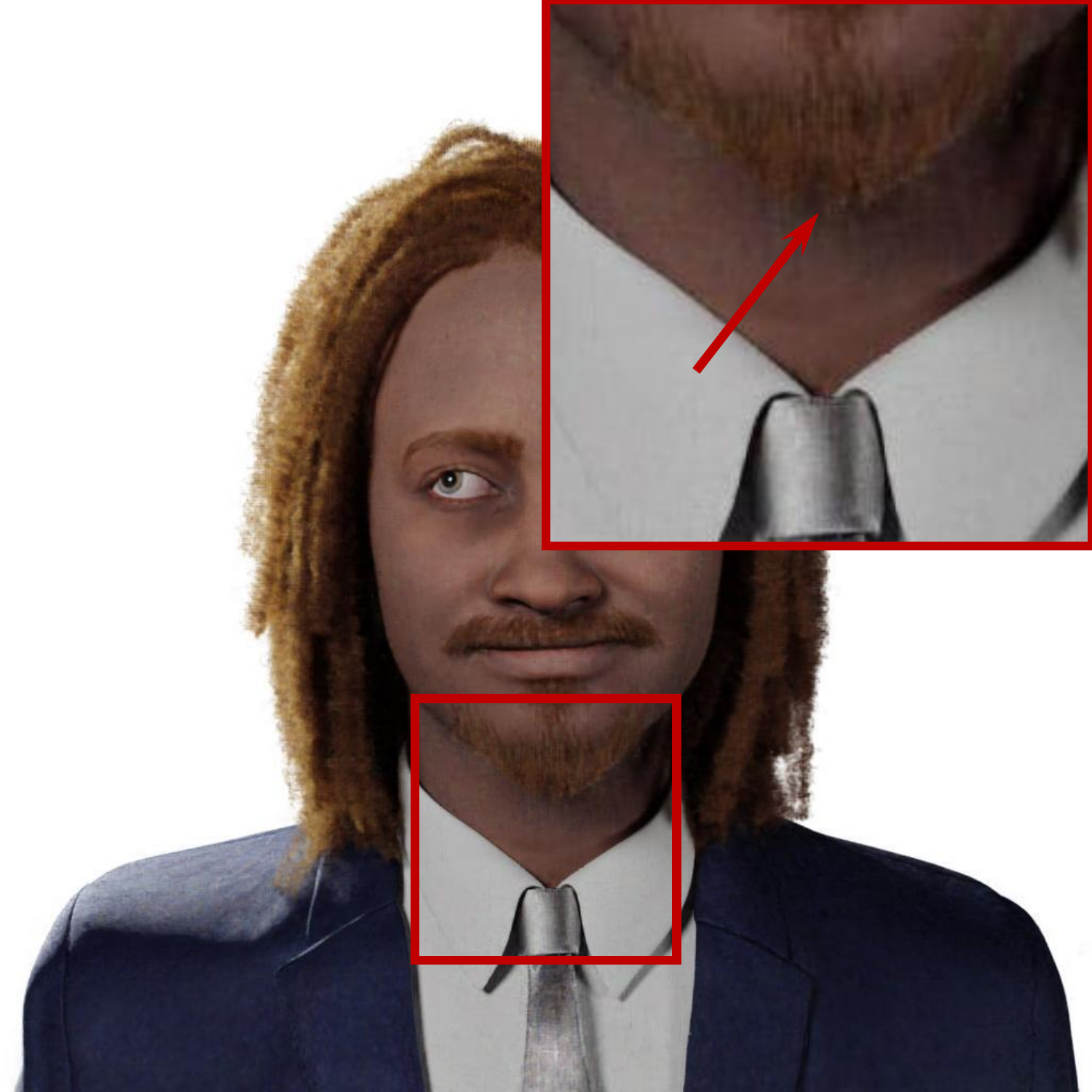} &
    \includegraphics[width=0.2\linewidth]{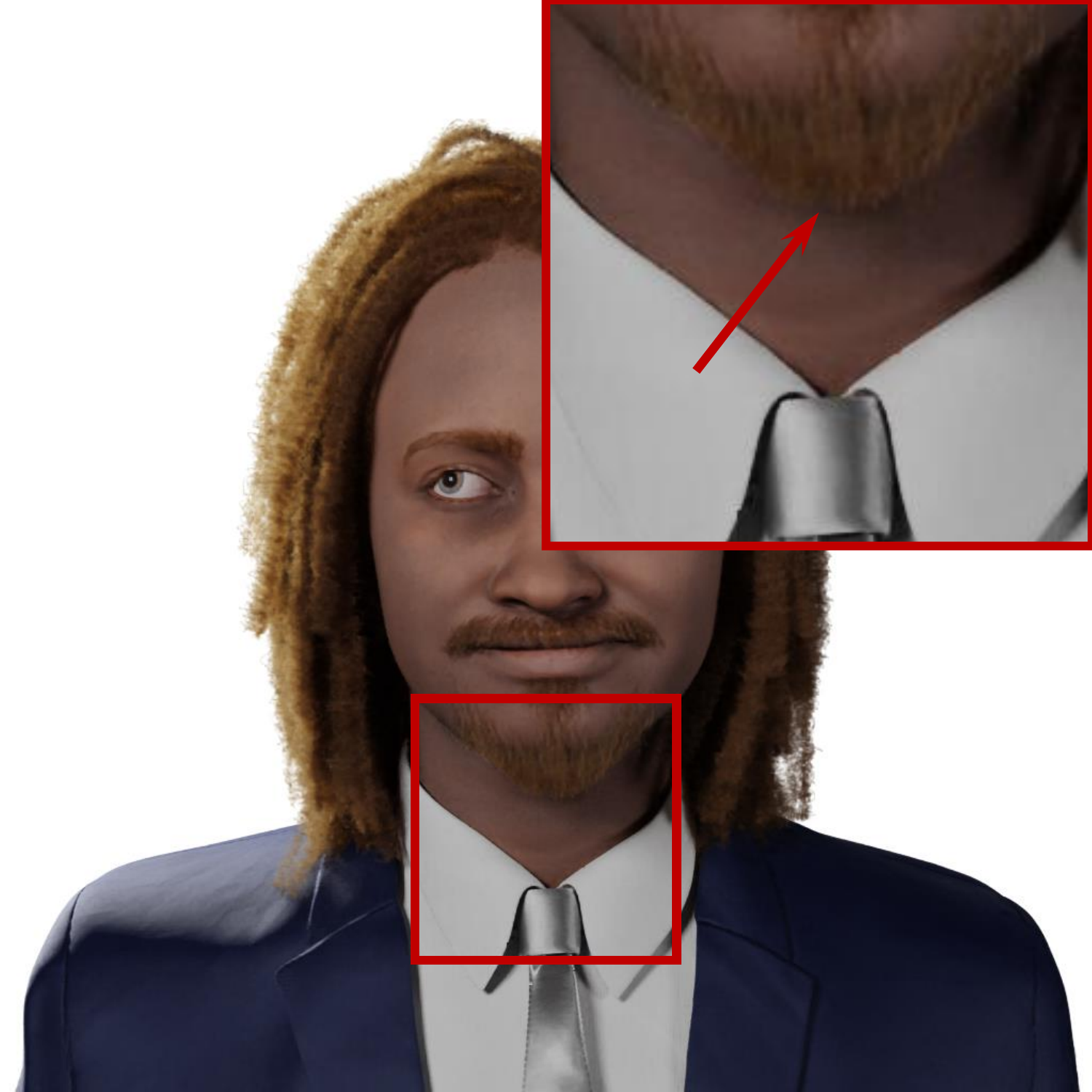} \\
    ~\cref{table:ablation-fitting-diffusion}(a) A. & ~\cref{table:ablation-fitting-diffusion}(a) B. & ~\cref{table:ablation-fitting-diffusion}(a) C. & ~\cref{table:ablation-fitting-diffusion}(a) E.
    \end{tabular}
  \caption{Qualitative ablation for the representation fitting.}
  \label{fig:fitting_ablation}
\end{figure}

\begin{figure}[th!]
  \centering
  \small
  \setlength\tabcolsep{3pt}
  \begin{tabular}{cccc} 
    \includegraphics[width=0.2\linewidth]{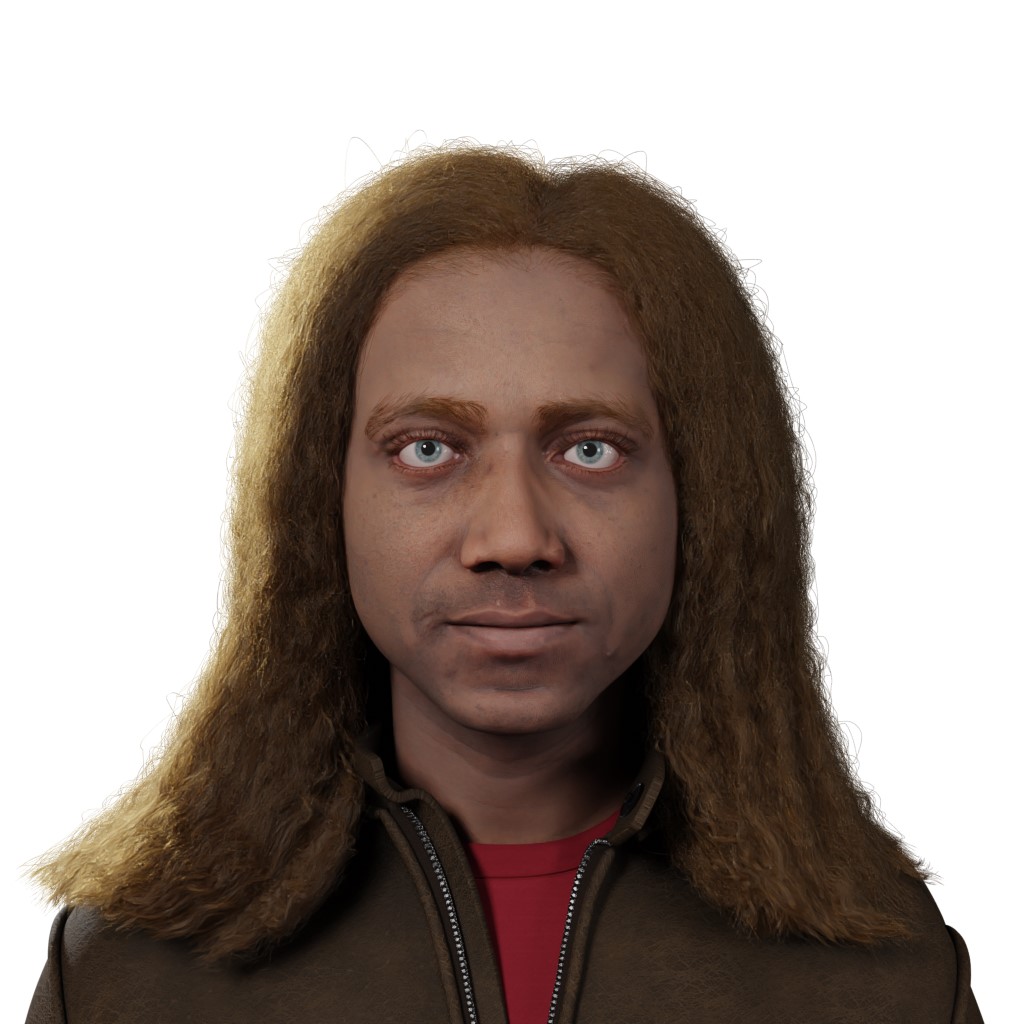}  &
    \includegraphics[width=0.2\linewidth]{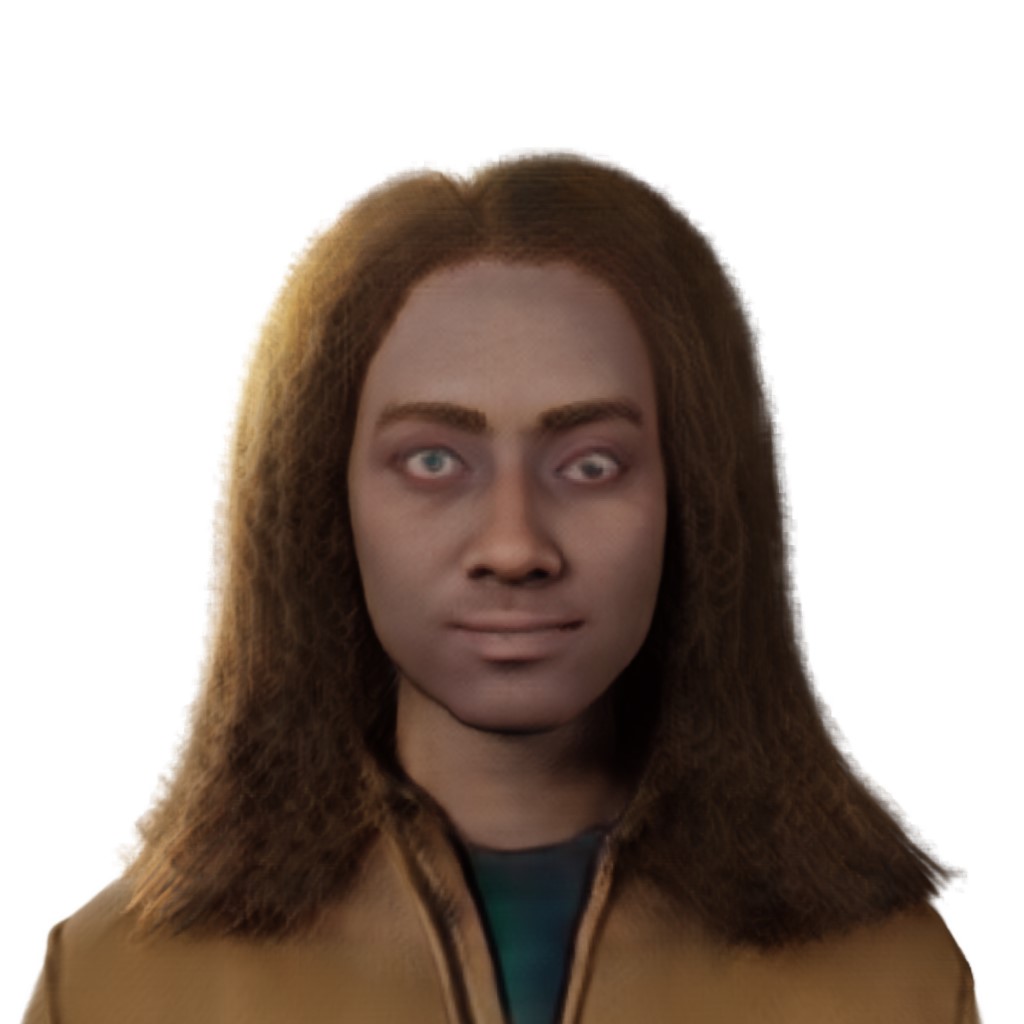} &
    \includegraphics[width=0.2\linewidth]{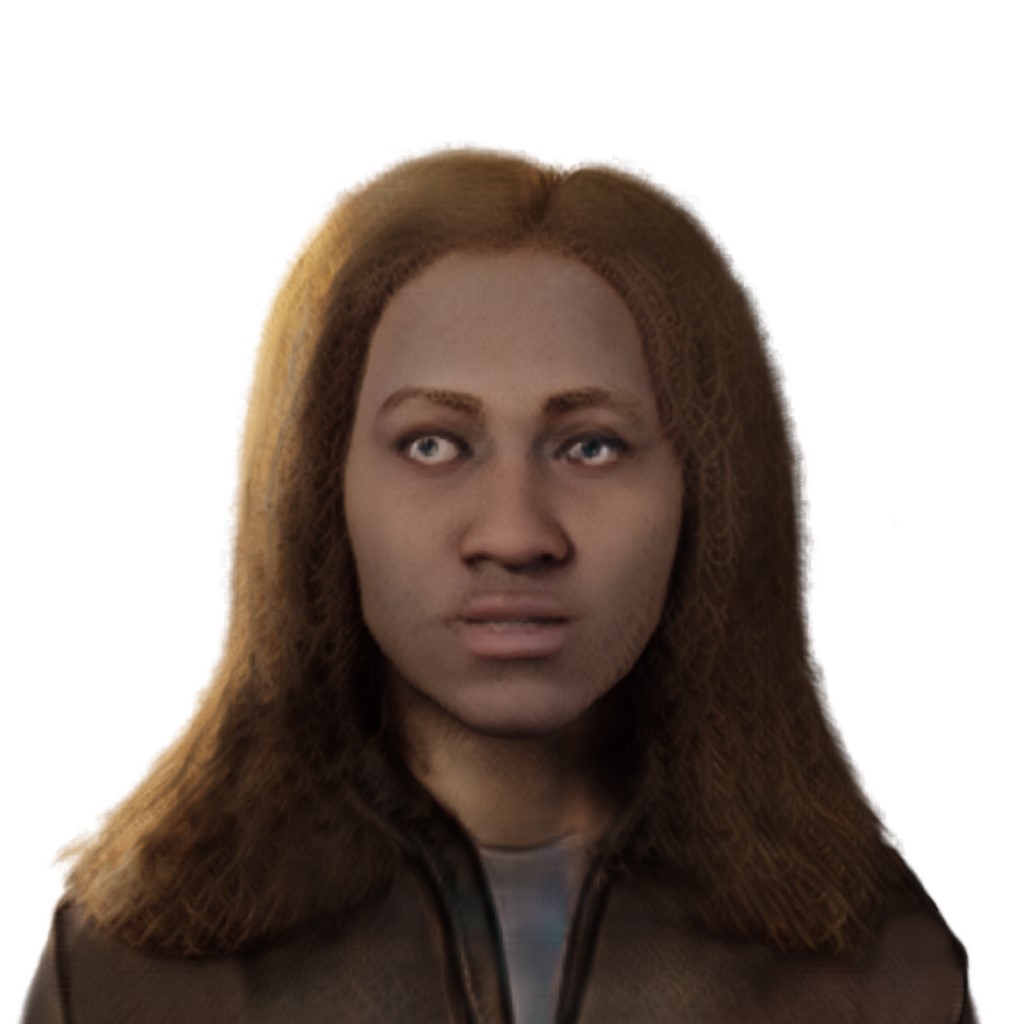} &
    \includegraphics[width=0.2\linewidth]{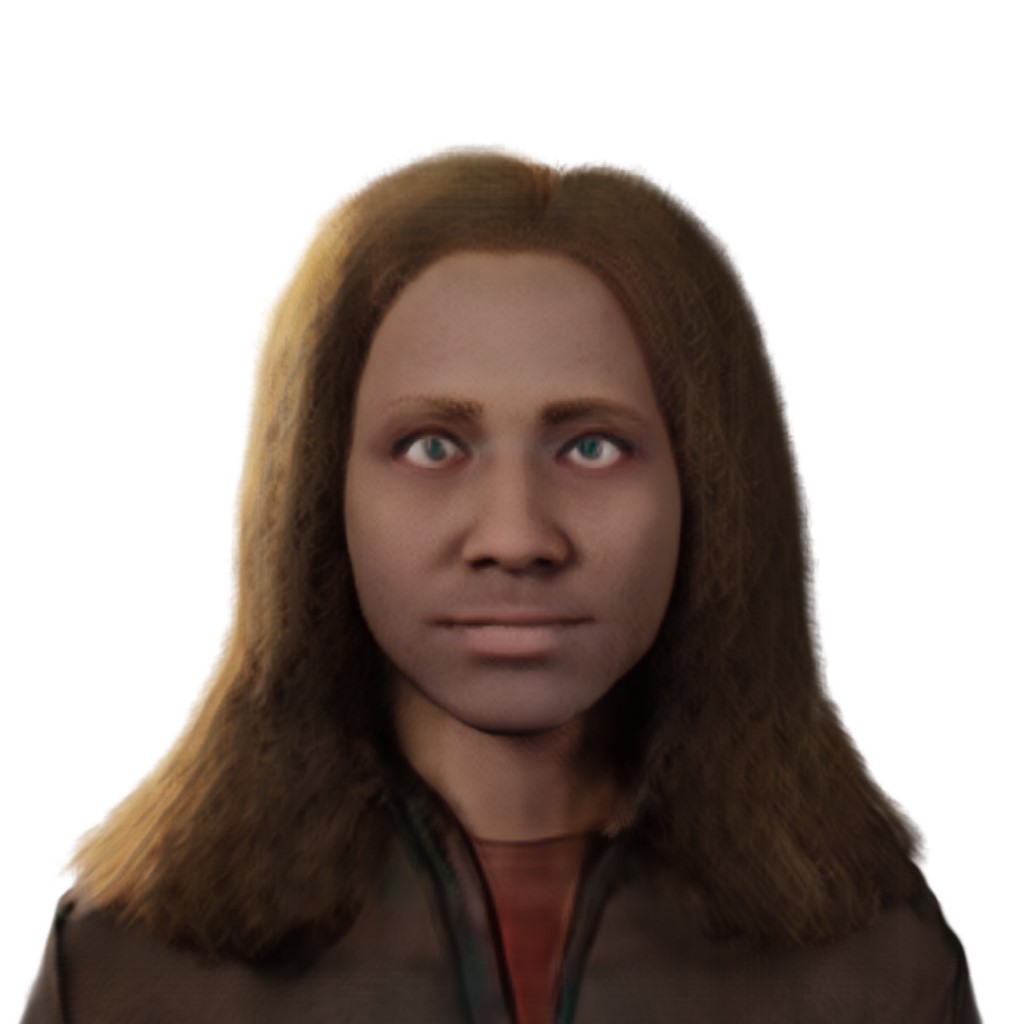} \\ 
    Reference & \cref{table:ablation-fitting-diffusion}(b) C. & \cref{table:ablation-fitting-diffusion}(b) D. & \cref{table:ablation-fitting-diffusion}(b) E. \\
  \end{tabular}
  \caption{Qualitative ablation for 3D diffusion model. Multi-scale feature conditioning results in significantly faithful and detailed generation results.}
  \label{fig:vis_ablation_model}
\end{figure}

\subsection{Ablation Study}

We first study the factors that affect triplane fitting. As shown in~\cref{table:ablation-fitting-diffusion}(a), we present results for five baselines. Rodin (256) is the method proposed in~\cite{wang2023rodin}. In Rodin (512), we naively increase the triplane resolution to 512. We can see that directly increasing the triplane resolution provides very marginal gains due to the forgetting problem. With our proposed bunch of techniques, scaling the resolution can notably increase PSNR to $31.54$. 

~\cref{fig:fitting_ablation} visualizes the differences of the baselines. Rodin (256) cannot obtain a sharp beard due to the low-resolution triplanes. Rodin (512) obtains sharp details but there are many noises caused by the under-fitted decoder. In contrast, our method with task replay and consolidation weight regularization is able to get clean and sharp details. Also, see the areas around the silver tie for differences.

We further evaluate the factors that affect the generation results of our diffusion model in~\cref{table:ablation-fitting-diffusion}(b). Starting from Rodin~\cite{wang2023rodin} which trains diffusion on $256\times256\times32$ triplanes, we only scale the triplane to $512\times512\times32$ with other factors unchanged. We can see that the results even become worse. We think this is because the original noises are not suitable for larger triplanes. Using a stronger noise schedule obtains more reasonable results. Replacing the original CLIP image encoder with the VAE encoder (single-scale) also improves the results. Our method with multi-scale features achieves the best results. 
We also provide visual comparison as shown in~\cref{fig:vis_ablation_model}, the introduced multi-scale features of the input portrait provide significantly more detailed texture cues, enabling high-fidelity avatar creation.

\begin{figure*}[t]
    \small
    \centering
    \begin{tabular}{c@{\hspace{3mm}}c@{\hspace{1mm}}c@{\hspace{3mm}}c@{\hspace{1mm}}c@{\hspace{3mm}}} 
    \includegraphics[width=0.15\linewidth]{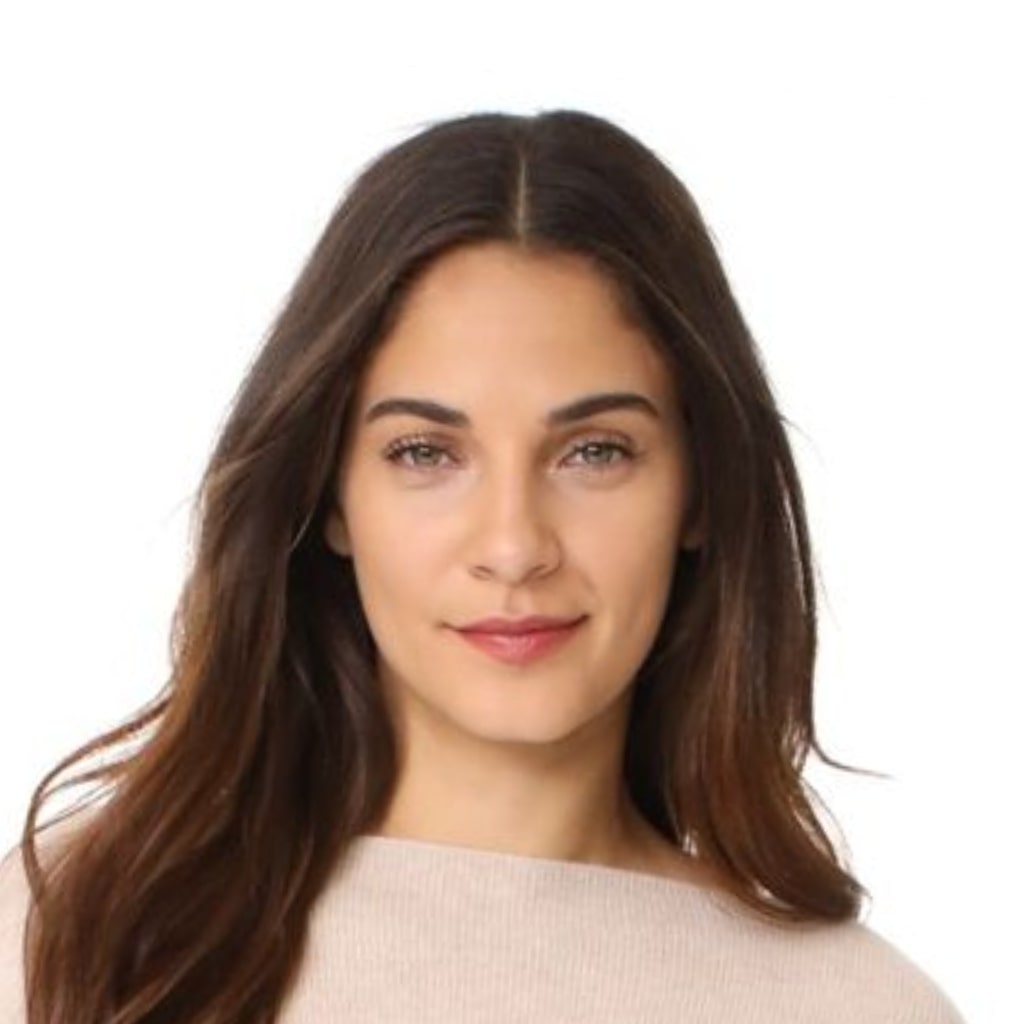}  &
    \includegraphics[width=0.15\linewidth]{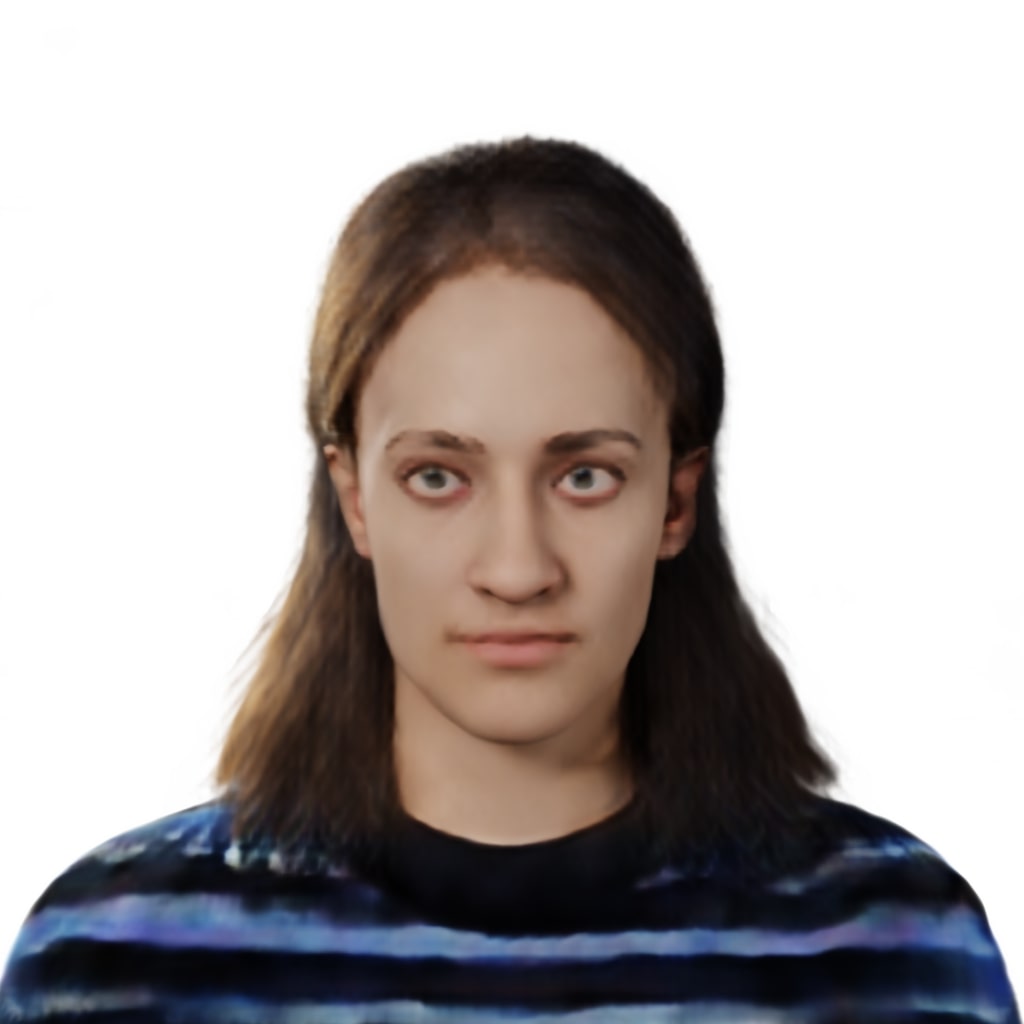} &
    \includegraphics[width=0.15\linewidth]{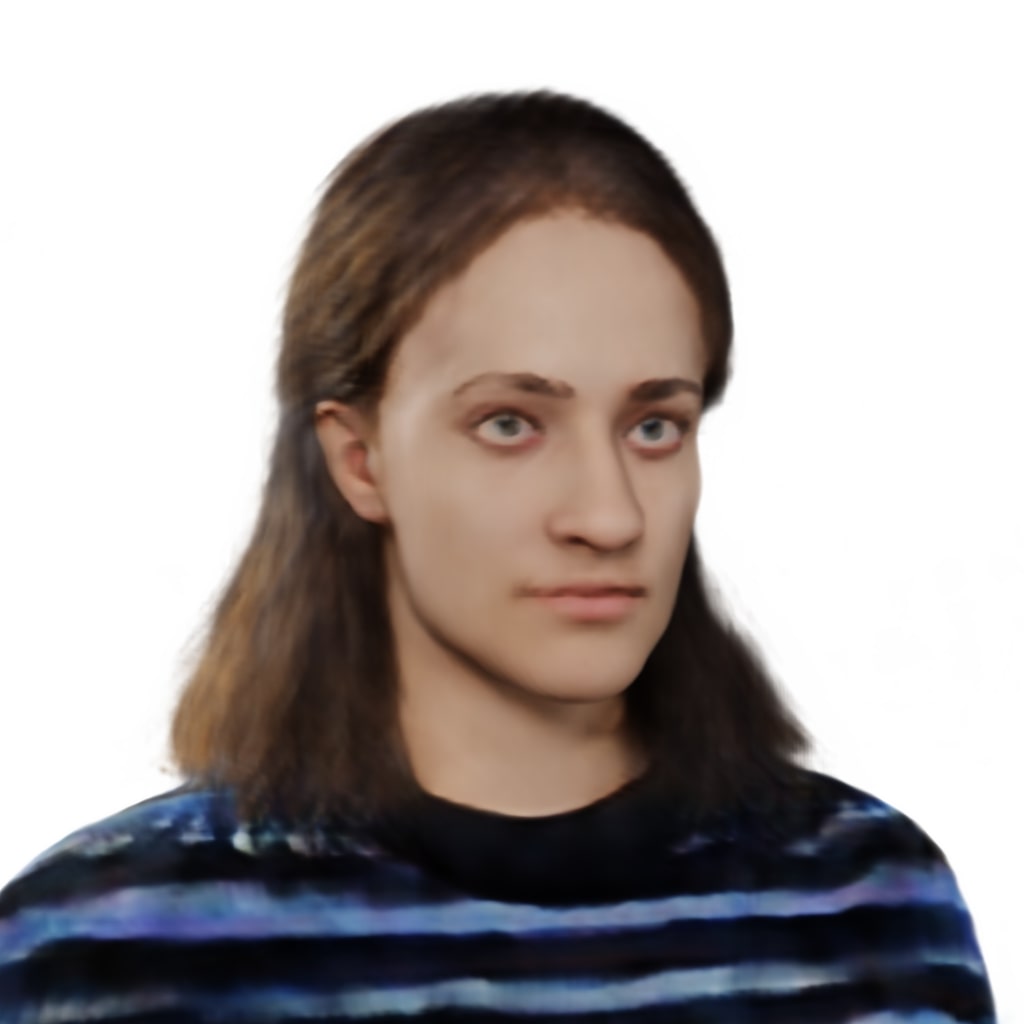} &
    \includegraphics[width=0.15\linewidth]{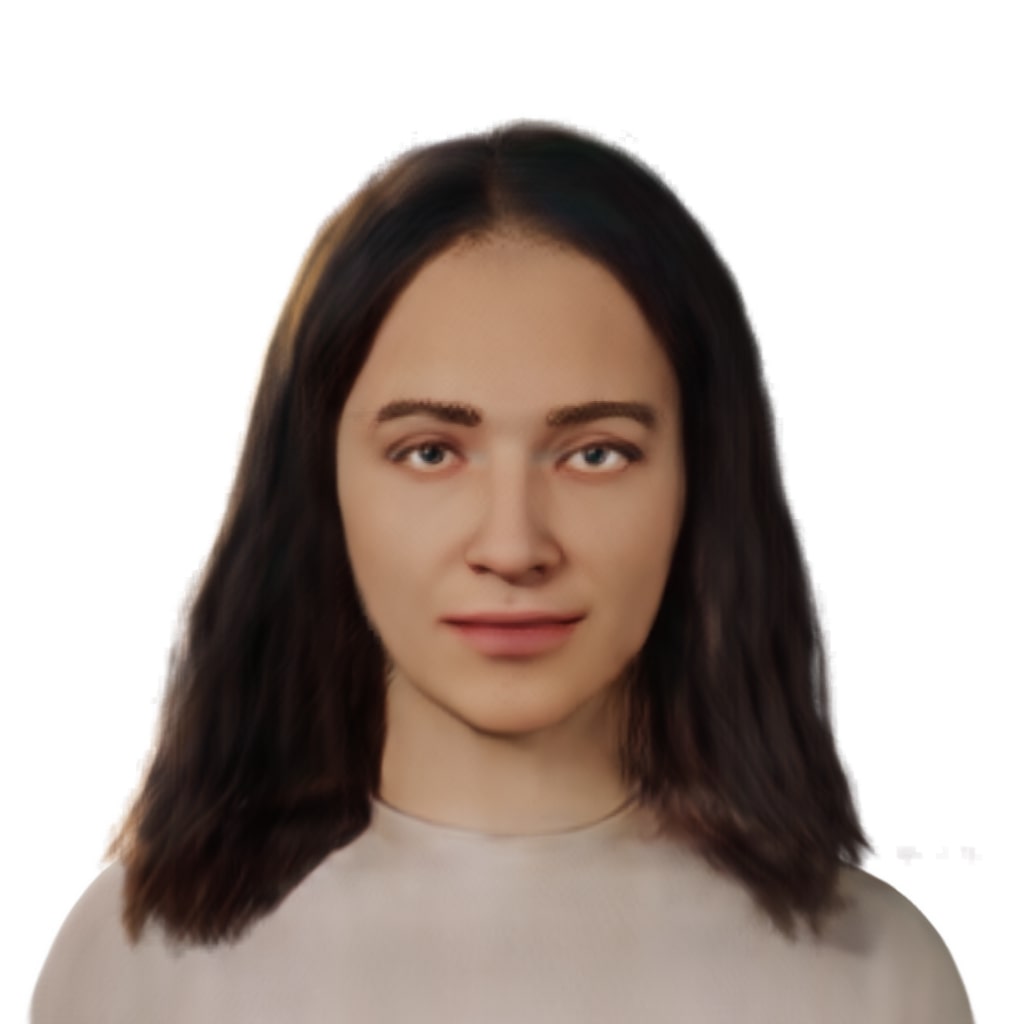} &
    \includegraphics[width=0.15\linewidth]{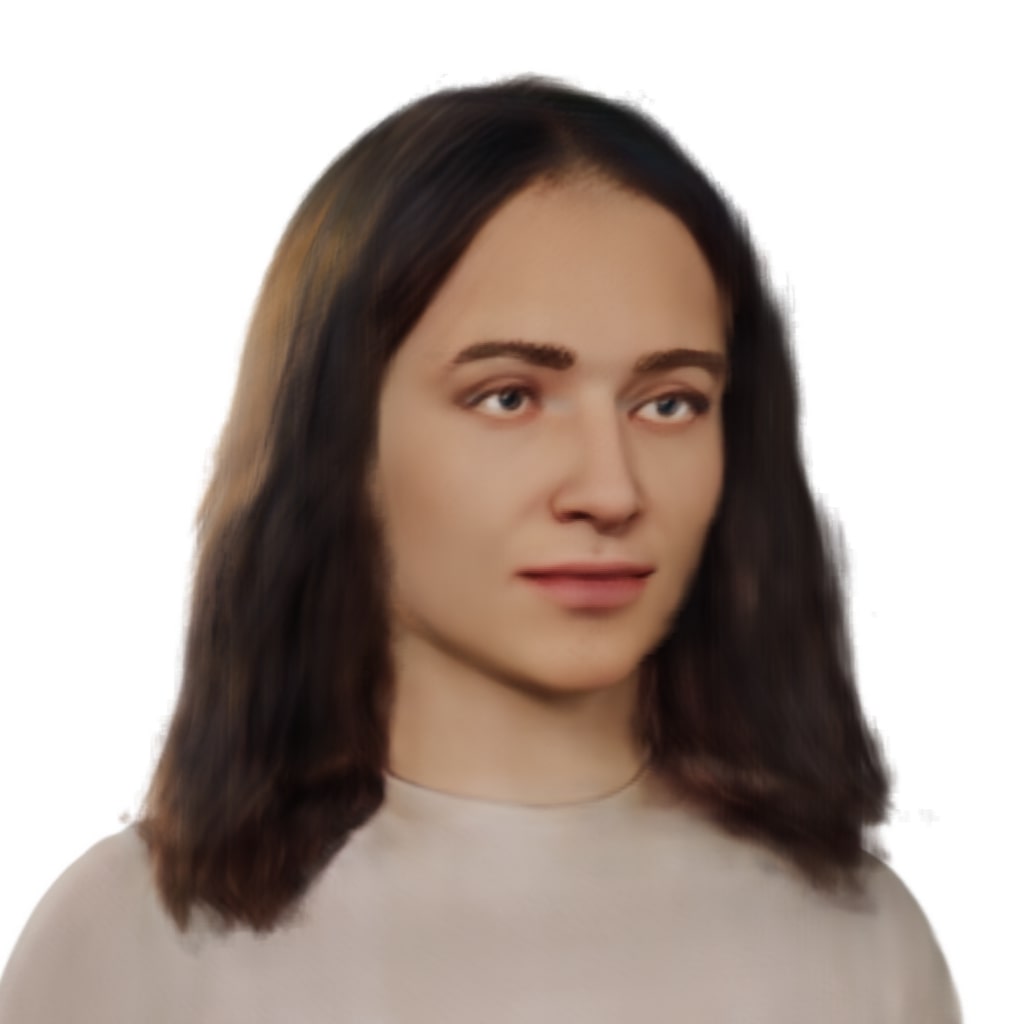} \\
    \includegraphics[width=0.15\linewidth]{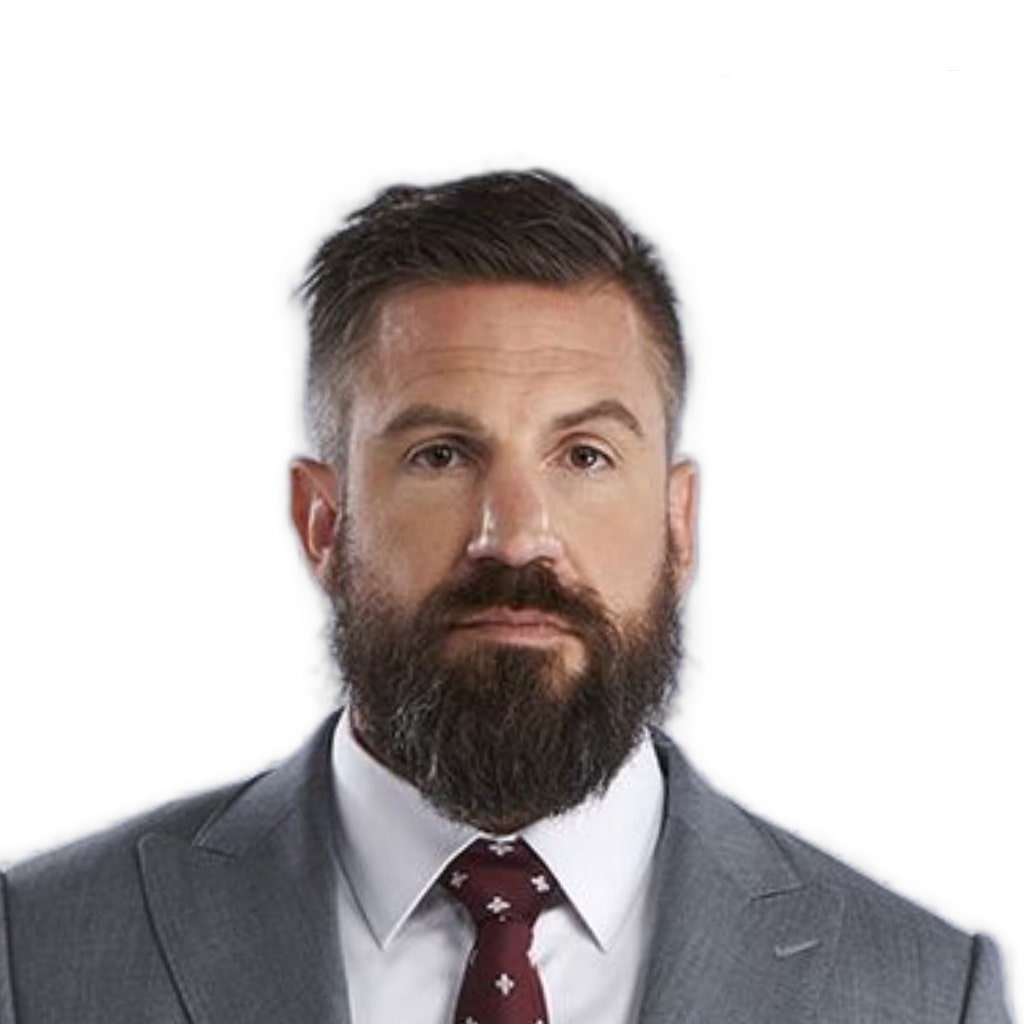}  &
    \includegraphics[width=0.15\linewidth]{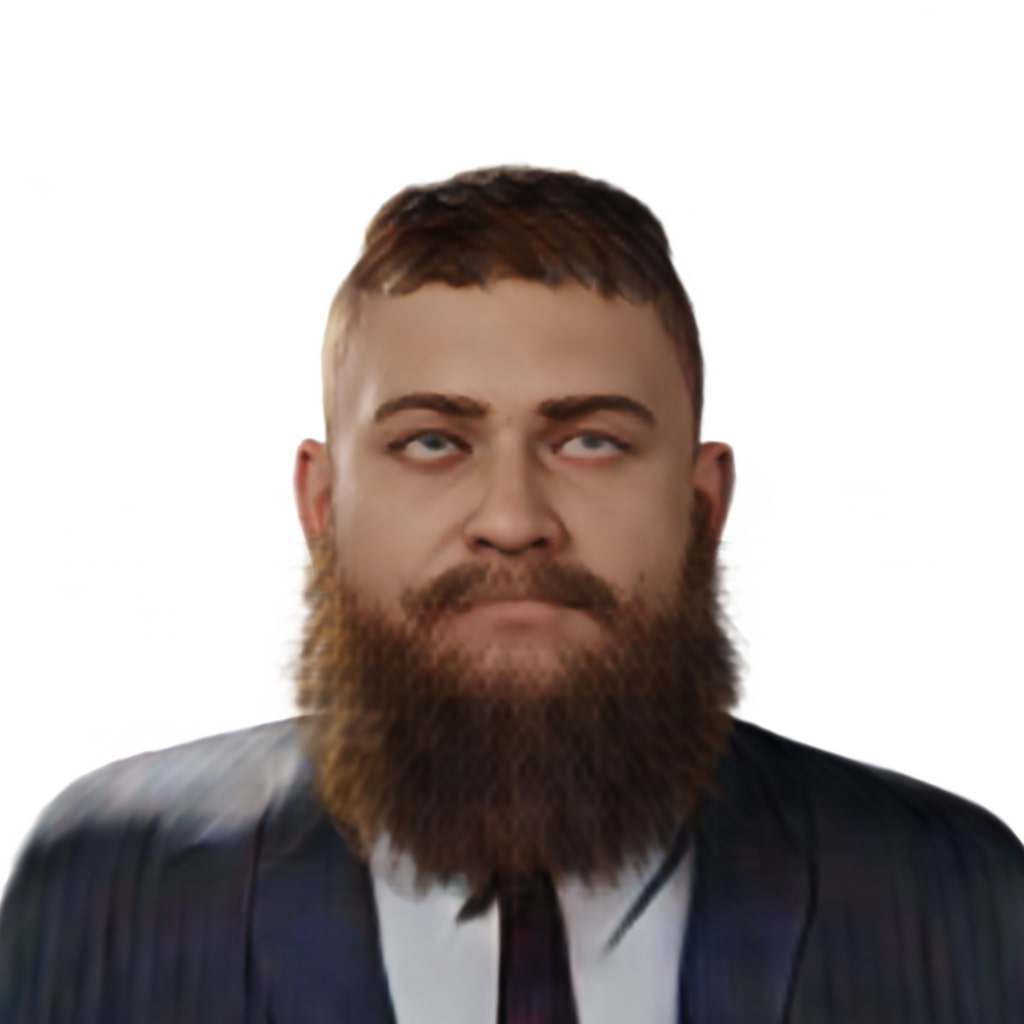} &
    \includegraphics[width=0.15\linewidth]{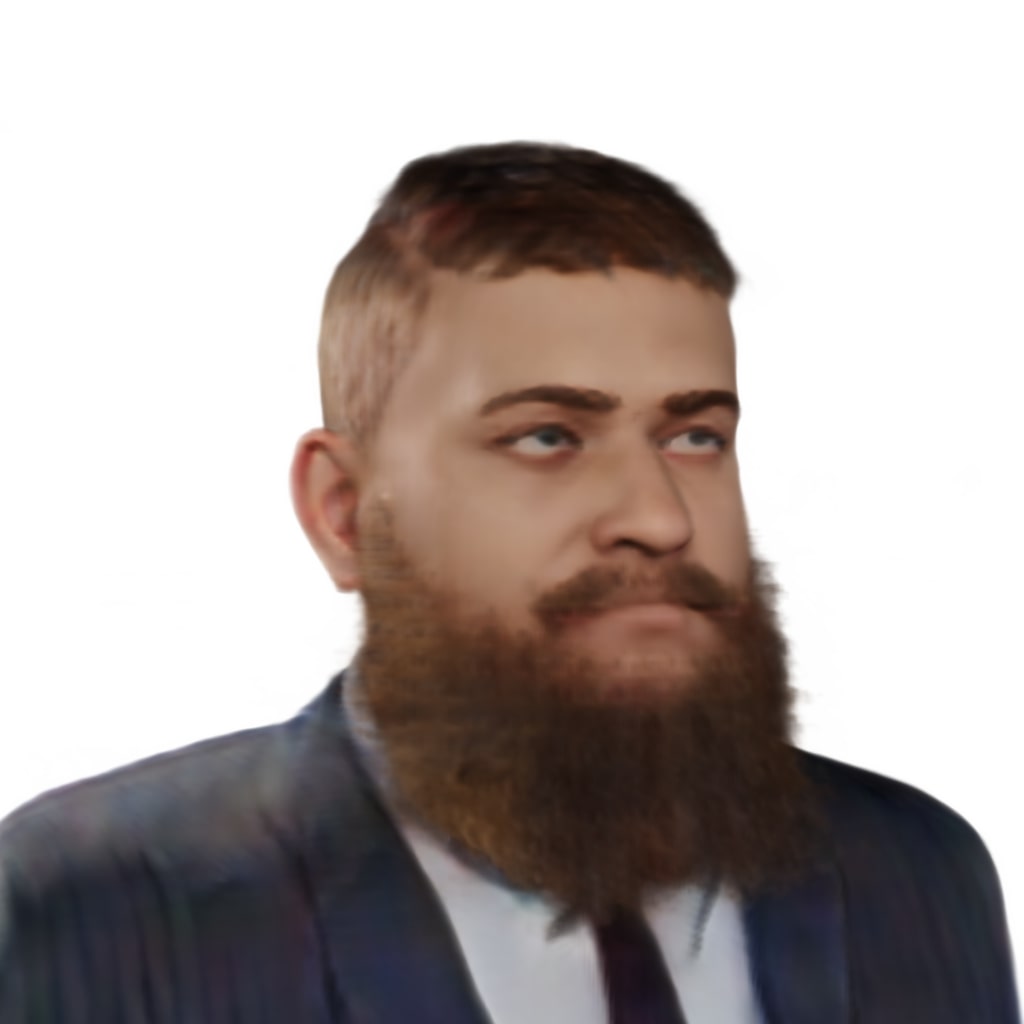} &
    \includegraphics[width=0.15\linewidth]{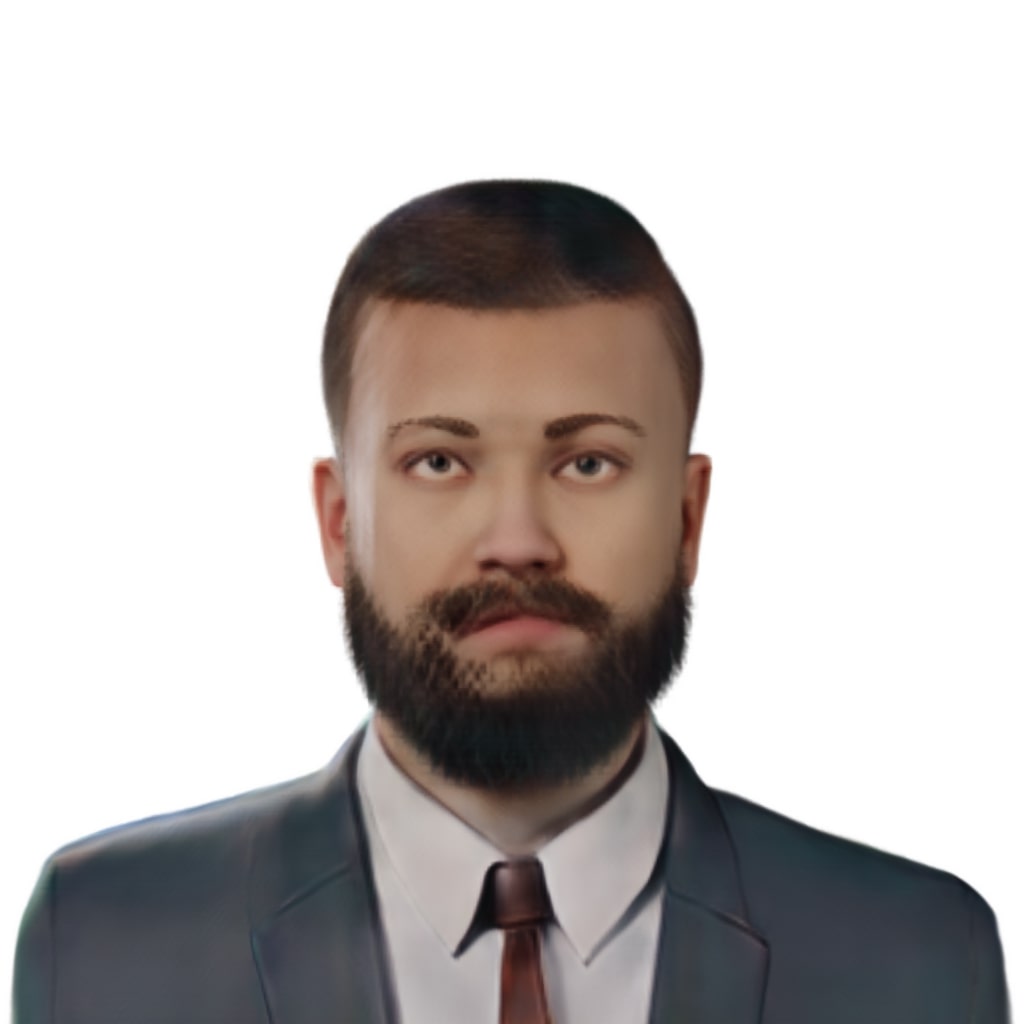} &
    \includegraphics[width=0.15\linewidth]{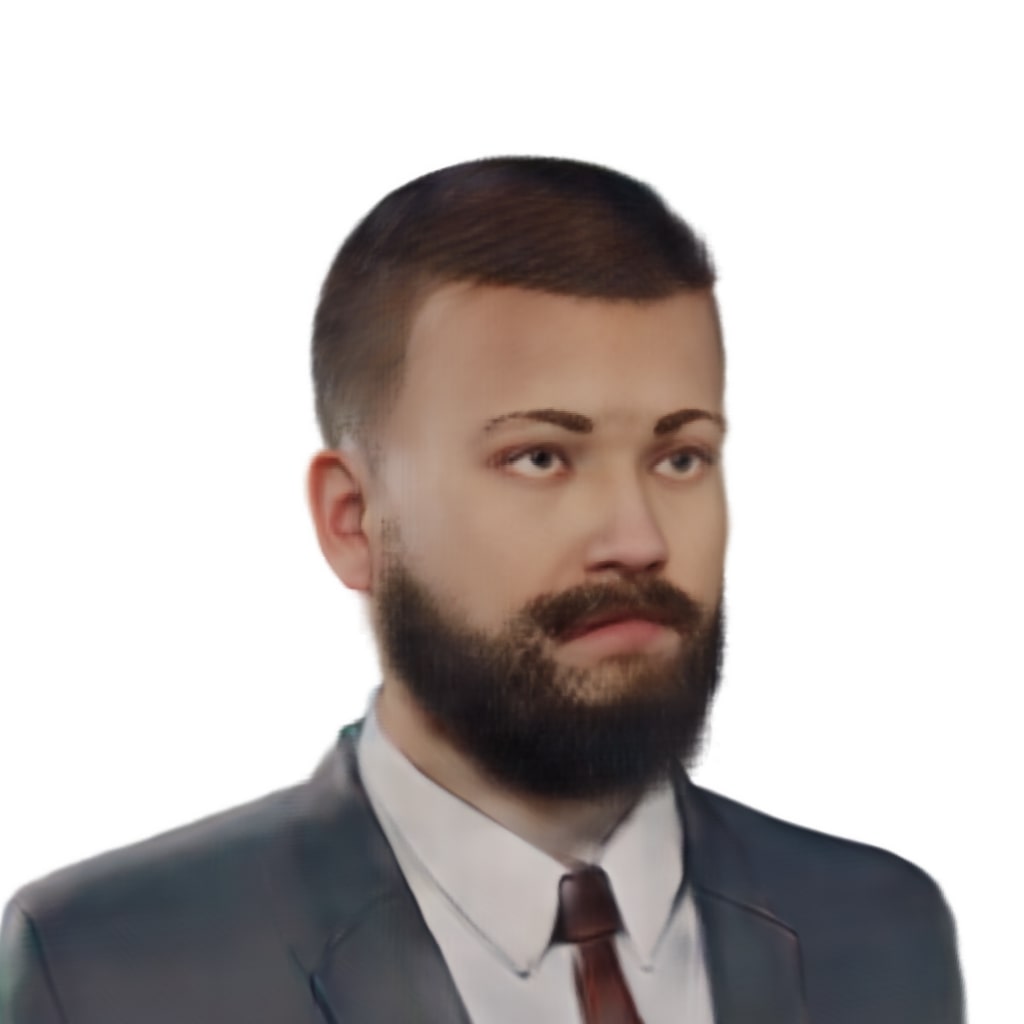} \\
    Reference & \multicolumn{2}{c}{Rodin} & \multicolumn{2}{c}{\textbf{Our RodinHD}}\\
  \end{tabular}
  \captionof{figure}{Avatars generated conditioned on single in-the-wild portraits.}
  \label{fig:real_world_dataset}
\end{figure*}

\begin{figure*}[t]
  \centering
  \scriptsize
  \begin{tabular}{cc} 
    \includegraphics[width=0.47\linewidth]{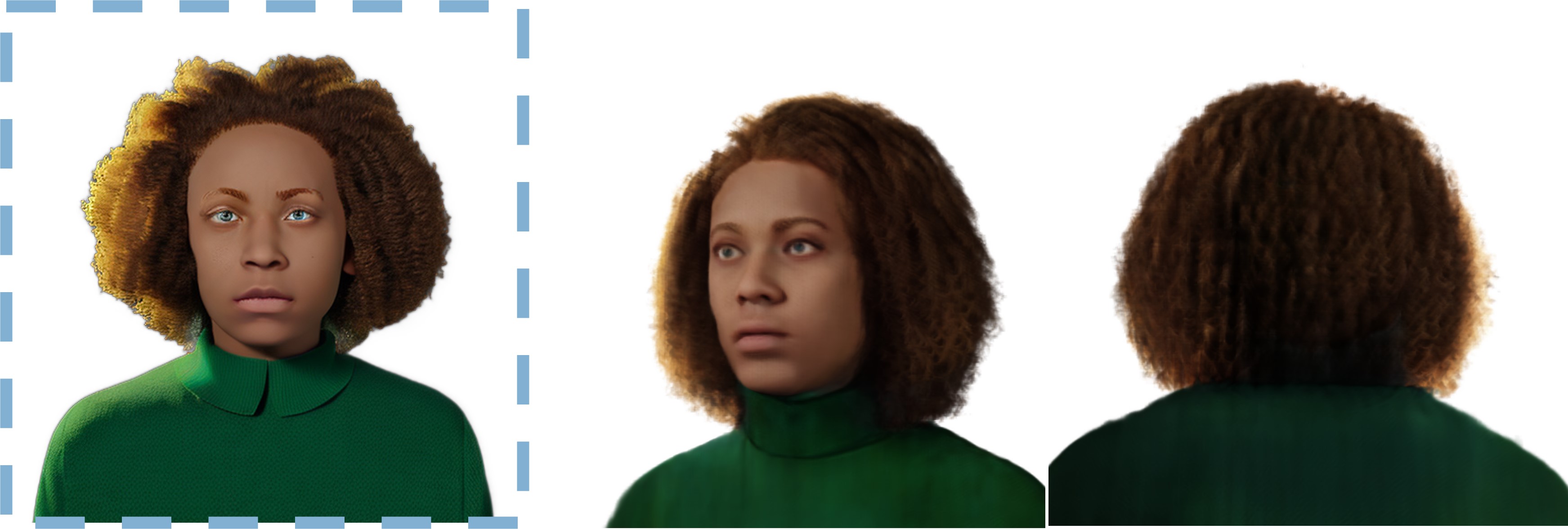} &
    \includegraphics[width=0.47\linewidth]{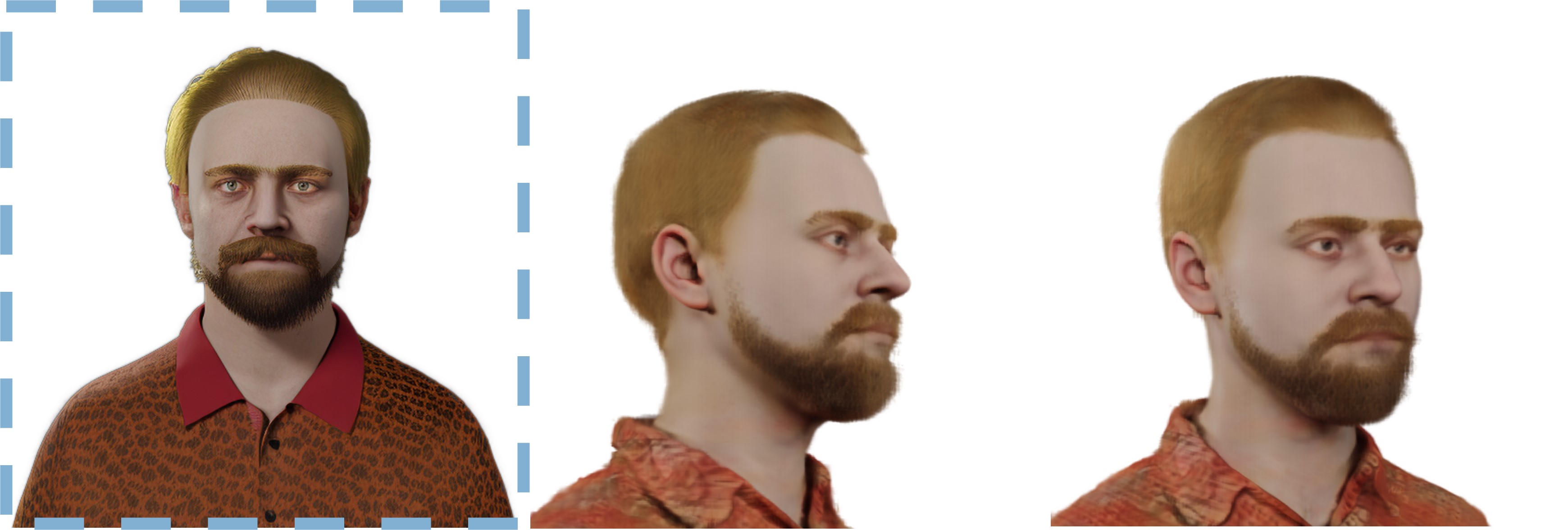} \\
    \makecell[c]{``\textit{Blender Synthetic Avatar, girl, blue eyes,} \\ \textit{brown afro hair, curly hair, green sweater}'' } &
    \makecell[c]{``\textit{Blender Synthetic Avatar, blonde hair, boy, } \\ \textit{brown eyes, medium and facial hair,} \\ \textit{red and orange shirt, mustache}'' }
  \end{tabular}
  \caption{RodinHD can create high-fidelity avatars based on the reference portraits (dashed boxes) generated by finetuned 2D text-to-image diffusion models.}
  \label{fig:text_cond_vis}
\end{figure*}

\subsection{Applications}

\noindent\textbf{Avatar creation from in-the-wild portrait.} Despite training on the synthetic dataset~\cite{wood2021fake}, our model is robust to create 3D avatars conditioned on single in-the-wild portraits. The results in~\cref{fig:real_world_dataset} validate our method's capability for generalization to real-world images, which outperforms Rodin~\cite{wang2023rodin} in retaining both identity and details.

\noindent\textbf{Text-conditioned avatar creation.} To enable high-quality avatar creation from text, we first convert the text prompt to reference portrait leveraging strong 2D diffusion network~\cite{rombach2022high}, thereafter generate a high-fidelity avatar with the reference portrait. To be more specific, we use 40 selected frontal images and corresponding text prompts in our synthetic dataset to perform LoRA-based~\cite{hu2021lora} finetuning for Stable Diffusion. We employ the term ``Blender Synthetic Avata'' as our trigger words within prompts to generate frontally aligned image inputs, which is a widely adopted technique for steering the style of images produced by fine-tuned SD.
We provide the samples of text-to-avatar creation in~\cref{fig:text_cond_vis}.

\section{Conclusion}
We present RodinHD for high-fidelity 3D avatar generation by improving both the data fitting and the diffusion modeling. To maintain the compelling visual details of avatars, we propose a task-relay strategy and identity-aware weight consolidation regularizer for high-quality and robust data fitting in large-scale. Furthermore, to model the distribution of highly detailed avatars, we introduce a multi-scale visual feature conditioning mechanism in our cascaded diffusion model, which provides fine-grind guidance to diffusion generation. In addition, we study previous 2D optimized noise scheduling in both high-resolution and high-dimensional 3D diffusion training. Our optimized noise schedules effectively enhance the details of the generated avatars. Extensive experiments show the proposed framework can generate high-fidelity 3D avatars with rich details, which is also promising to apply our model to general 3D scene modeling.

\section*{Acknowledgments} This work was supported in part by the Anhui Provincial Natural Science Foundation under Grant 2108085UD12. We acknowledge the support of GPU cluster built by MCC Lab of Information Science and Technology Institution, USTC. 

\bibliographystyle{splncs04}
\bibliography{main}
\clearpage
\appendix

\section{Additional Implementation Details}
\noindent\textbf{Triplane fitting.} 
We split the triplane fitting into two stages to reduce computation costs. 
In the first stage, we jointly train the MLP decoder and the triplanes on a subset of 64 avatars. During each inner loop iteration, $8192$ rays are randomly sampled for loss calculation. We optimize one avatar per GPU for each outer loop iteration due to large GPU memory consumption. 
The detailed hyper-parameters of the first stage are listed in~\cref{tab/supp:hypers}, including ablation studies.
In the second stage, we fix the decoder’s weights and fine-tune the triplanes of 46$K$ avatars independently.
The fitting iteration for each avatar is set to 25000.
For rendering efficiency, an occupancy grid of $128^3$ resolution is maintained~\cite{muller2022instant} to skip ray marching steps in empty space. Since we do not have the occupancy grid of the diffusion generated triplane, we update the occupancy grid 16 times from zero initialization before performing volumetric rendering.

\noindent\textbf{Diffusion training.} For triplane $\bm{x} = (\bm{x}_{uv}, \bm{x}_{wu}, \bm{x}_{vw})$ of shape $\mathbb{R}^{3 \times H\times W \times C}$, we perform triplane roll-out  $\overline{\bm{x}} =  \mathtt{hstack}(\bm{y}_{uv},\bm{y}_{wu},\bm{y}_{vw})\in \mathbb{R}^{H\times {3W} \times C}$ in order to employ the well-designed 2D UNet model in diffusion~\cite{nichol2021improved,dhariwal2021diffusion} following~\cite{wang2023rodin}. We also leverage 3D-aware convolution~\cite{wang2023rodin} for cross-plane feature communication. The portrait image is resized to $256 \times 256$ and the resulting multi-scale features have the resolution of $128 \times 128$, $64 \times 64$, and $32 \times 32$. The conditional features are injected to the base diffusion model at layers with resolutions of $32\times32$, $16\times16$, and $8\times8$, respectively, through cross attention. The upsample diffusion model only uses the conditional features of $128 \times 128$, which are injected to the middle latent features.

For our base diffusion model, we adopt the UNet model architecture from~\cite{dhariwal2021diffusion}. We train our base model using AdamW optimizer~\cite{adamw} with a learning rate $1e-5$. To condition on the multi-scale image features of input portrait as illustrated in Sec. 3.2, we perform cross attention at resolutions $(32, 16, 8)$. Our optimized noise schedule is based on the cosine schedule mentioned in~\cite{chen2023importance}, and we further adjust its hyper-parameters for 3D diffusion training. We provide the detailed configurations of the model and diffusion below.

For our upsample diffusion model, we also adopt the UNet model architecture from ~\cite{dhariwal2021diffusion}. We train our upsample model using Adam optimizer~\cite{kingma2014adam} with a learning rate $1e-5$. We remove self-attention due to unaffordable computation cost at high resolutions, and perform cross attention at resolution $128$ for conditioning on input portrait features. Our optimized noise schedule for upsample diffusion is based on the sigmoid noise schedule in ~\cite{chen2023importance}, then we carefully adjust the hyper-parameters for 3D diffusion training. The detailed configurations of the model and diffusion are shown below.

\begin{lstlisting}
# 128x128 Base diffusion
UNet configuration = {
    "channels": 192,
    "channel_mult": (1, 1, 2, 3, 4),
    "embed_dim": 768,
    "num_res_blocks": (3, 3, 3, 3, 3),
    "attn_resolutions": (32, 16, 8),
    "ms_vae_feature_cross_attn_res": (32, 16, 8),
    "3D_aware_conv_res": (128),
    "dropout": 0,
    "feature_pooling_type": "attention",
    "use_scale_shift_norm": True
}

Diffusion configuration = {
    "Training steps": 1000,
    "Noise schedule": Cosine(start=0.2, end=1, tau=3),
    "Inference steps": 10,
    "Inference sampler": "DDPM"
}

# 128x128 -> 512x512 Upsample diffusion
UNet configuration = {
    "channels": 128,
    "channel_mult": (1, 2, 4),
    "embed_dim": 512,
    "num_res_blocks": (2, 2, 6),
    "ms_vae_feature_cross_attn_res": (128),
    "3D_aware_conv_res": (512, 256, 128),
    "dropout": 0,
    "feature_pooling_type": "attention",
    "use_scale_shift_norm": False
}

Diffusion configuration = {
    "Training steps": 100,
    "Noise schedule": Sigmoid(start=0, end=3, tau=0.1),
    "Inference steps": 10,
    "Inference sampler": "DDPM"
}

\end{lstlisting}

\begin{table*}[t]
    \caption{Hyper-parameters for the first stage of fitting, including ablation studies.}
    \setlength\tabcolsep{3pt}
    \small
    \begin{center}
    \begin{tabular}{lcccc}
    \toprule
     & Rodin (512) & + Task relay & + Wight decay & Ours \\
    \midrule
    Inner loop iterations & 15000 & 5000 & 5000 & 5000 \\
    Outer loop iterations per avatar & 1 & 30 & 30 & 30 \\
    Loss Weight of TV regularization   & 1e-2 & 1e-2 & 1e-2 & 1e-2 \\
    Loss Weight of L2 regularization    & 1e-4 & 1e-4 & 1e-4 & 1e-4 \\
    Loss Weight of IWC regularization   & 0 & 0 & 0 & 0.1 \\
    Loss Weight of weight decay & 0 & 0 & 1e-4 & 0 \\
    Triplane learning rate & 2e-3 & 2e-3 & 2e-3 & 2e-3  \\
    Decoder learning rate & 2e-4 & 2e-4 & 2e-4 & 2e-4 \\
    Ray batch size & 8192 & 8192 & 8192 & 8192 \\
    Samples per ray & 1024 & 1024 & 1024 & 1024 \\
    \bottomrule
    \end{tabular}
    \end{center}
    \label{tab/supp:hypers}
\end{table*}

\begin{table*}[th]
\caption{Comparison fitting quality (PSNR) of Triplane Resolution and Channel. }
\centering
\setlength\tabcolsep{5pt}
\begin{tabular}{lcccc}
\toprule
\diagbox{\textit{Res.}}{\textit{Ch.}} & 4 & 8 & 16 & 32 \\
\midrule
128   & 30.15 & 30.71 & 31.21 & 31.59 \\
256   & 30.24 & 31.01 & 31.44 & 31.67 \\
512   & 30.38 & 31.31 & 31.60 & \textbf{31.71} \\
\bottomrule
\end{tabular}
\label{tab/supp:fitting}
\end{table*}

\section{Additional Analysis and Visualization}

\begin{figure*}[t]
  \centering
  \includegraphics[width=\textwidth]{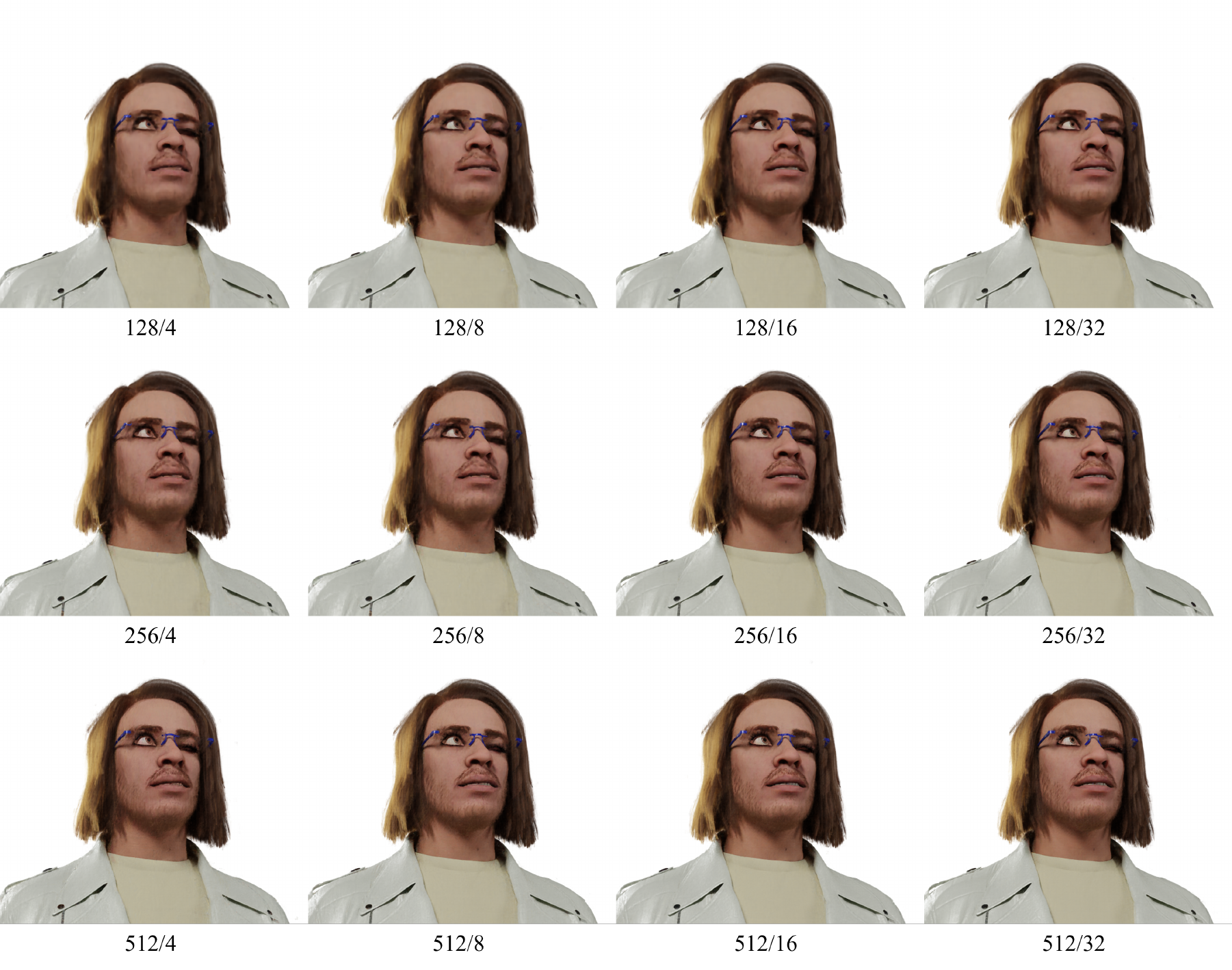}
  \caption{Comparison on Triplane Resolution and Channel. Zoom in for better visualization.}
  \label{fig/supp:fitting}
  \vspace{-5mm}
\end{figure*}

\noindent\textbf{Choices of triplane resolution and channel.}
We argue that both triplane resolution and channel affect the preservation of high-frequency information in renderings.
To validate this argument, we experiment with different triplanes from a resolution set of $\{128, 256, 512\}$ and a channel set of $\{4, 8, 16, 32\}$ to fit $1024 \times 1024$ images of one subject and show the results in~\cref{tab/supp:fitting} and~\cref{fig/supp:fitting}. 
Overall, the fitting quality increases with the triplane resolution and channel.
High-resolution triplane can render high-frequency detail, and low-resolution triplane tends to produce blurring results.
On the other hand, the triplane with more channels can keep high-fidelity appearance without introducing noisy pattern, but low-resolution triplane with more channels can not achieve better high-frequency detail preservation than high-resolution triplane.
We thus choose to utilize $512 \times 512 \times 32$ triplanes in our experiments.

\begin{figure*}[t]
  \centering
  \small
  \setlength\tabcolsep{1pt}
  \begin{tabular}{ccccccc} 
    \includegraphics[width=0.13\linewidth]{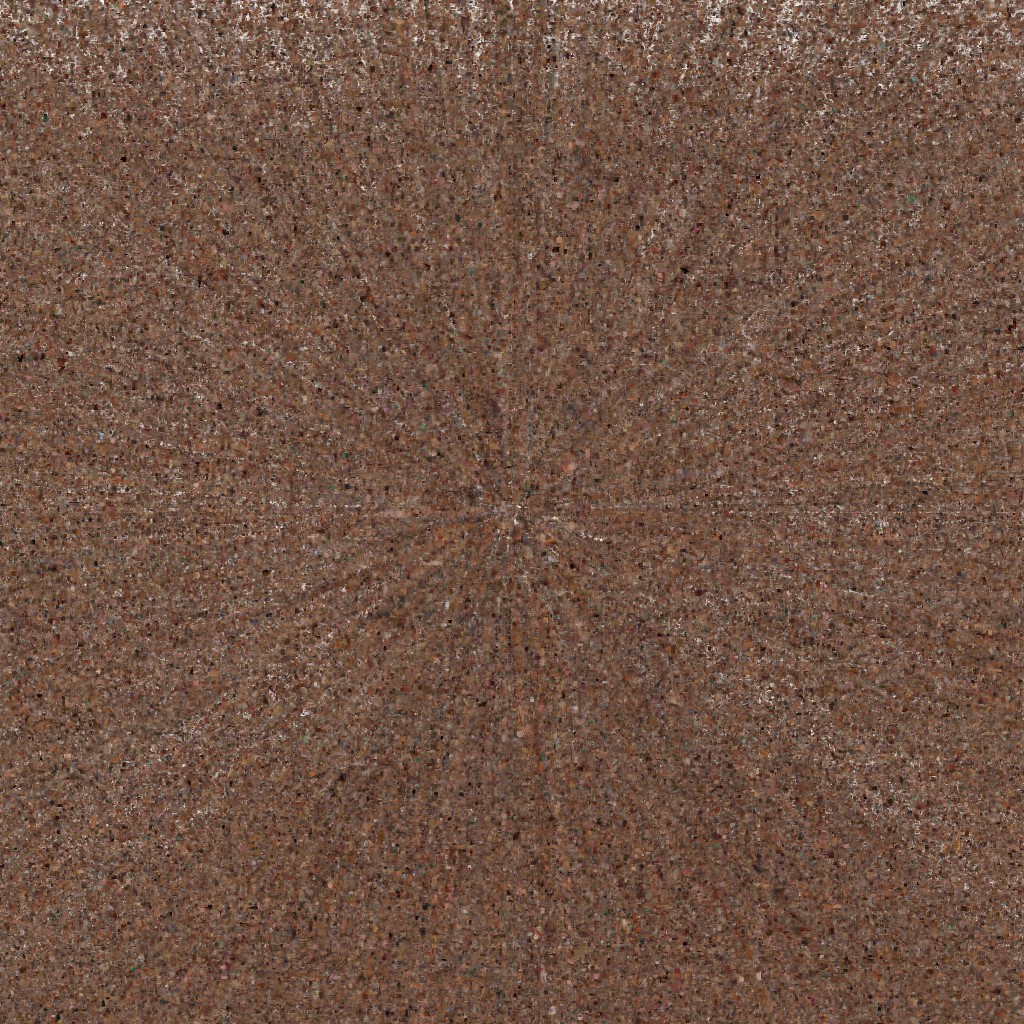}  &
    \includegraphics[width=0.13\linewidth]{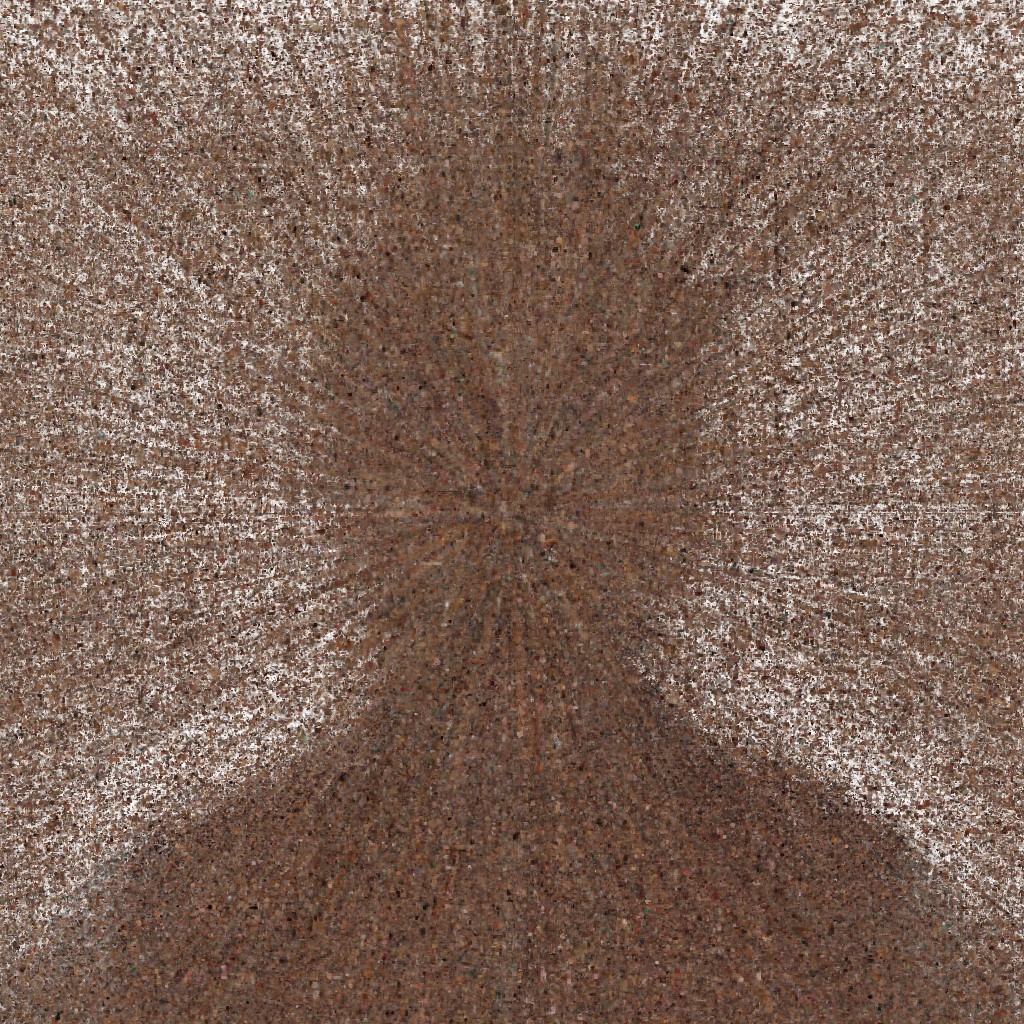} &
     \includegraphics[width=0.13\linewidth]{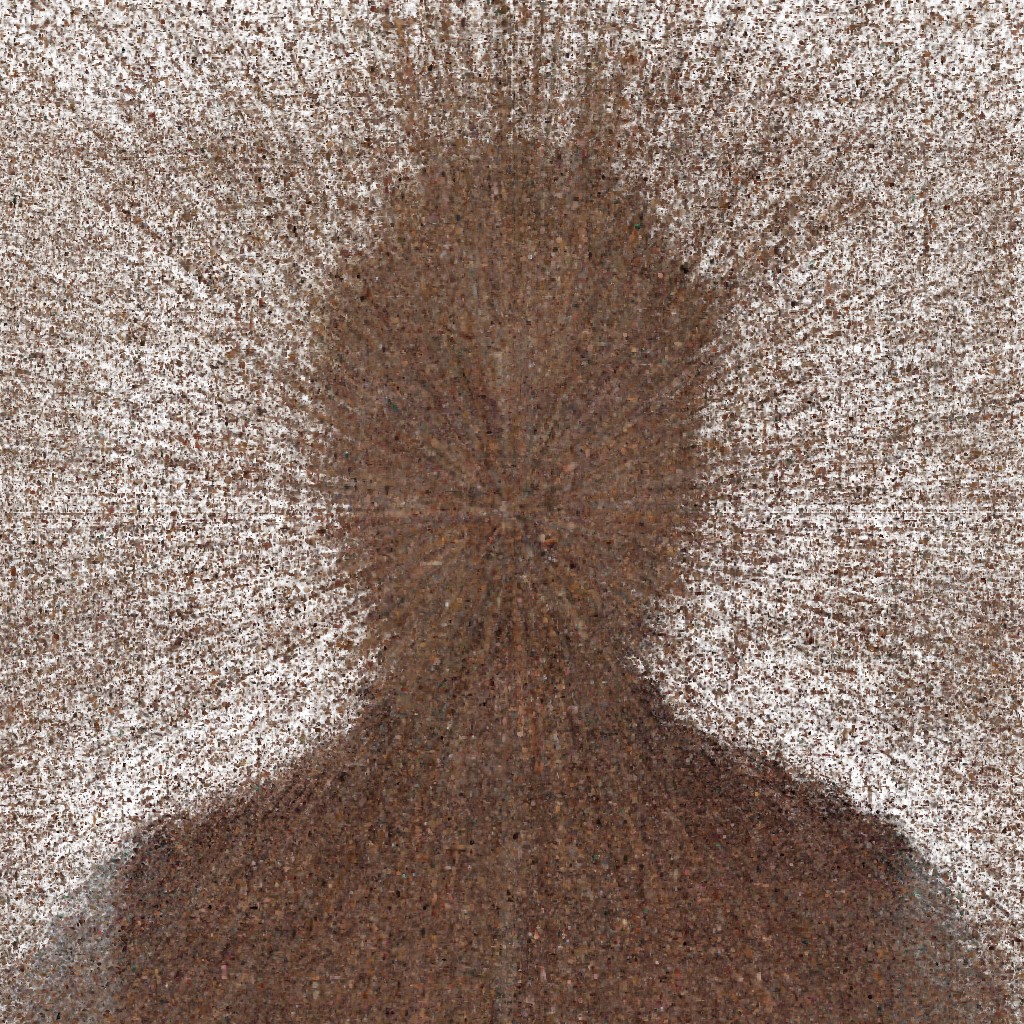}  &
    \includegraphics[width=0.13\linewidth]{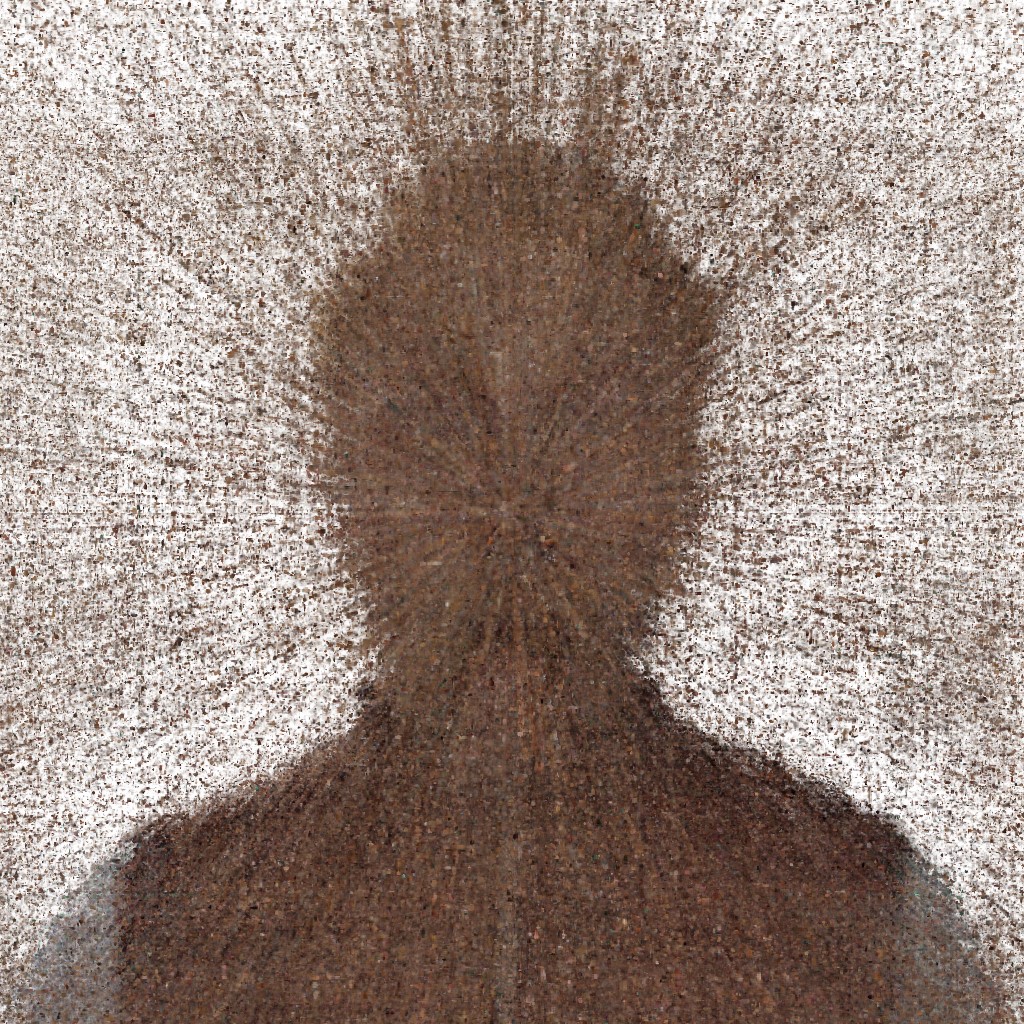} &
    \includegraphics[width=0.13\linewidth]{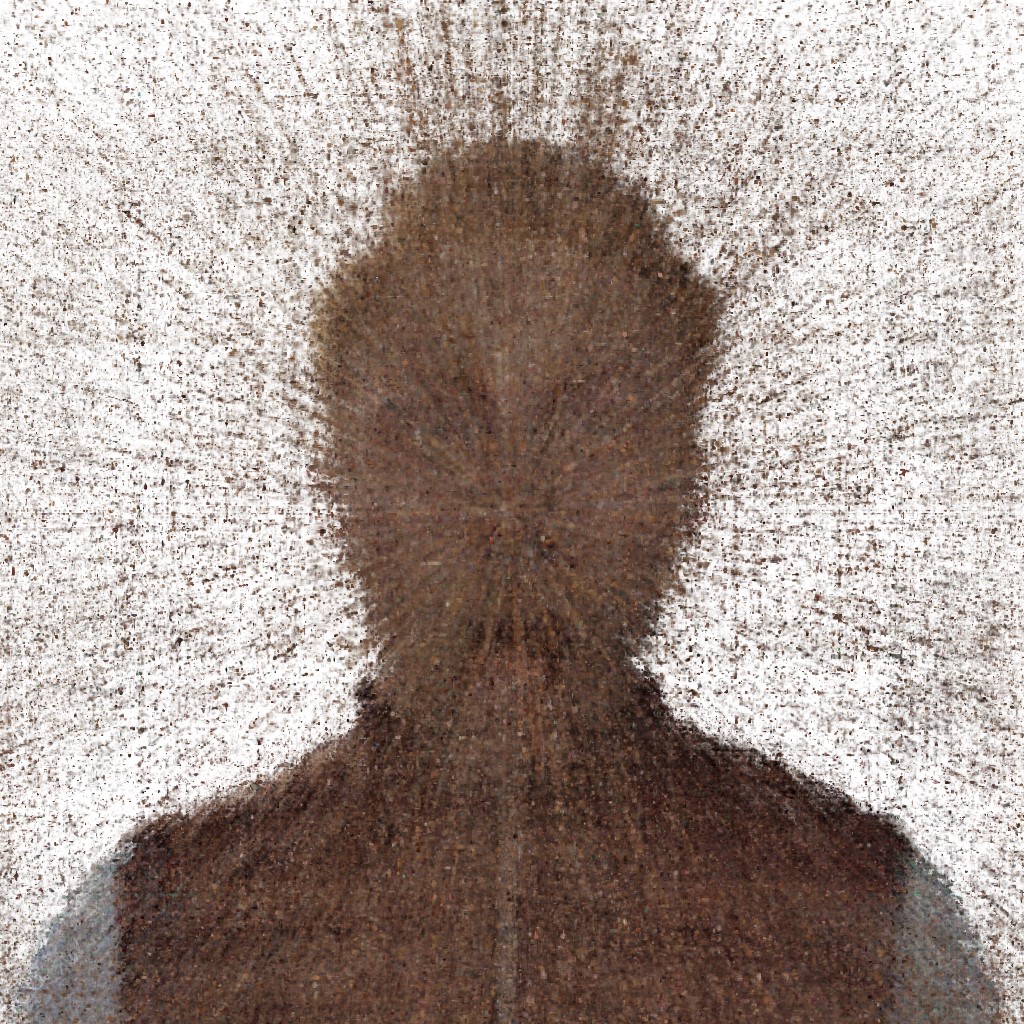}  &
    \includegraphics[width=0.13\linewidth]{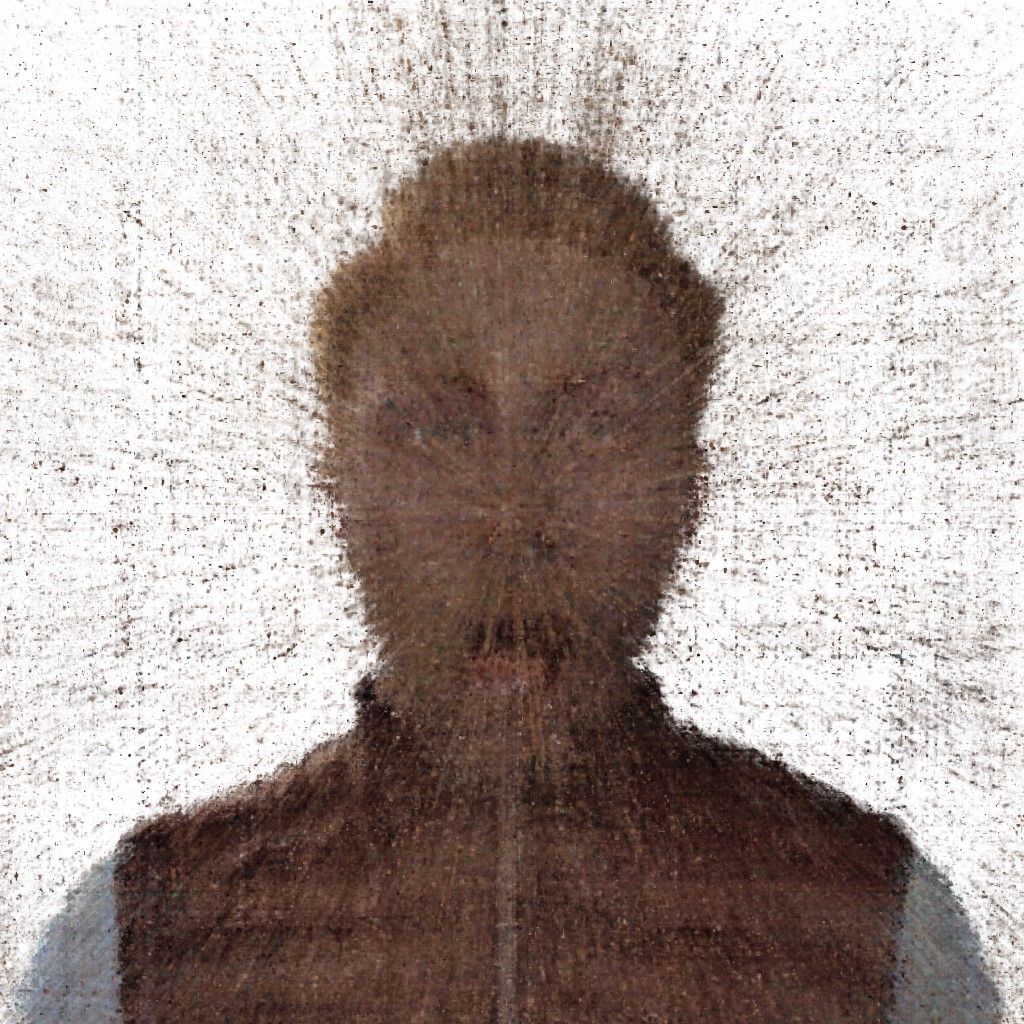} &
    \includegraphics[width=0.13\linewidth]{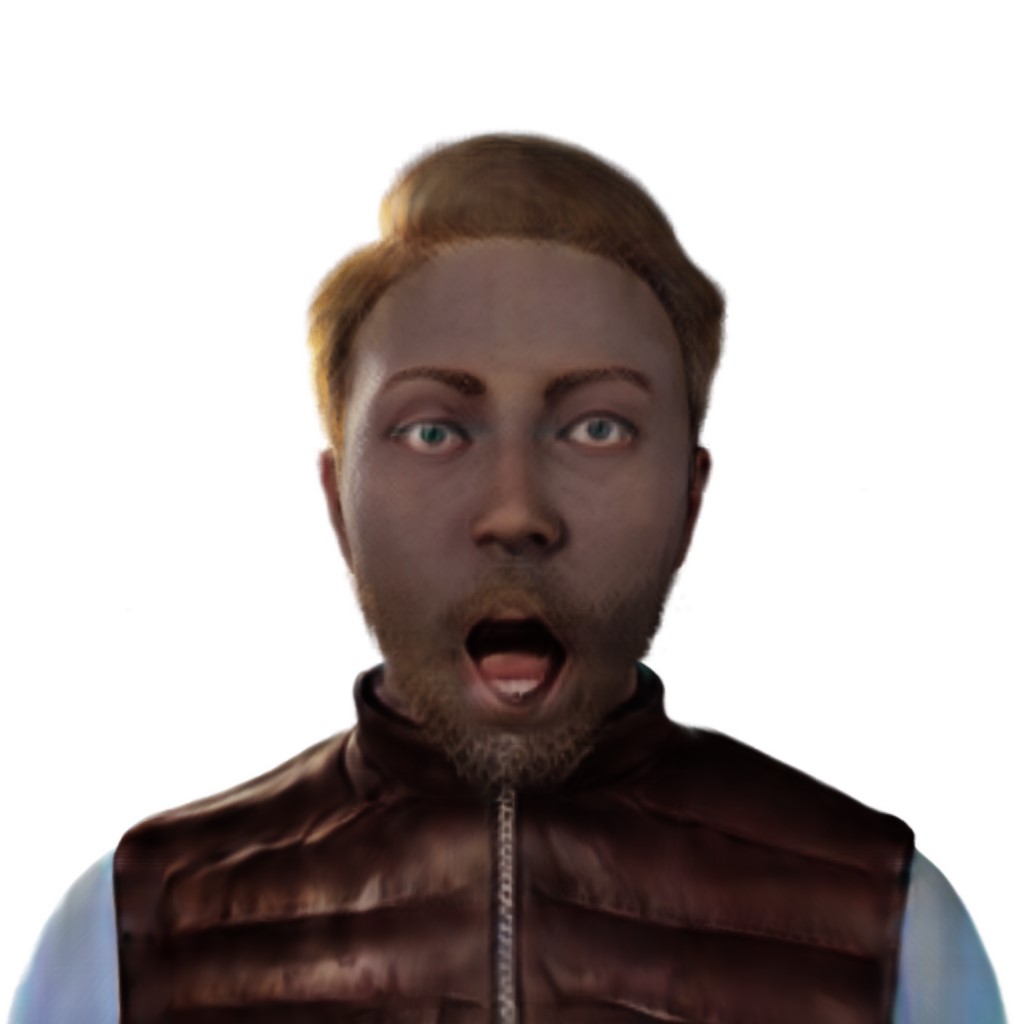}  
  \end{tabular}
  \caption{Visualization of intermediate results in the denoising process.}
  \label{fig/supp:diffusion_vis}
  \vspace{-5mm}
\end{figure*}

\noindent\textbf{Visualization of intermediate results in the denoising process.} During inference, our model starts from isotropic Gaussian noise and progressively reduces the noise to obtain the final high-quality triplanes. We visualize the renderings of generated results $\bm{x}_t$ of intermediate timesteps $t\in (0, 1)$ in the denoising process to provide a comprehensive understanding of the triplane diffusion procedure. From~\cref{fig/supp:diffusion_vis}, we observe that our model establishes the global structure of the avatar, and subsequently adds more detail, which is similar to ~\cite{wang2023rodin,shue20233d}.

\begin{table}[t]
  \caption{Additional comparison of conditional avatar generation. \textbf{The subscript $^*$ indicates that 2D refinement is applied to the rendered images}.}
  \centering
  \small
  \setlength\tabcolsep{3pt}
  \begin{tabular}{lccccccc}
  \toprule  
  Models & FID$\downarrow$ & PSNR$\uparrow$ & CSIM$\uparrow$ & CSIM-CrossView$\uparrow$ & AED$\downarrow$ & APD\%$\downarrow$ & ASD$\downarrow$ \\
  \midrule
  Rodin & 33.20 & 18.28 & 0.64 & \textbf{0.85} & 0.21 & 2.21 & \textbf{0.44} \\
  Rodin$^*$ & \textbf{20.51} & 17.31 & 0.63 & 0.83 & 0.20 & 2.21 & \textbf{0.44}\\
  \cellcolor{gray}\textbf{Ours} & \cellcolor{gray}26.49 & \cellcolor{gray}\textbf{20.33}& \cellcolor{gray}\textbf{0.68} & \cellcolor{gray}\textbf{0.85} & \cellcolor{gray}\textbf{0.18} & \cellcolor{gray}\textbf{1.78} & \cellcolor{gray}\textbf{0.44}
  \\
  \bottomrule
  \end{tabular}
 \label{supp/table:cond_quantitative_comparison}
\end{table}

\section{Additional Comparison}

\noindent\textbf{More quantitative comparison with Rodin.} We additionally compare our method with Rodin~\cite{wang2023rodin} on conditional avatar generation using more evaluation metrics. We evaluate the cosine similarity of identity embedding derived from ArcFace~\cite{deng2019arcface} between generated avatars and ground-truths (CSIM), as well as between paired renderings of generated avatars from different camera viewpoints (CSIM-CrossView). We also include Average Expression Distance (AED), Average Pose Distance (APD) and Average Shape Distance (ASD) between the reconstructed 3D faces~\cite{deng2019accurate} of generated avatars and ground-truth avatars. The results presented in~\cref{supp/table:cond_quantitative_comparison} demonstrate that our model excels in preserving identity and accurately generating expression, pose, and geometry. While Rodin's 2D refinement achieves lower FID scores, it struggles to maintain the identity and expression details of the conditioned portraits.

\begin{figure}[t]
  \centering
  \small
  \begin{tabular}{@{}c@{}c@{}} 
    \includegraphics[width=0.5\linewidth]{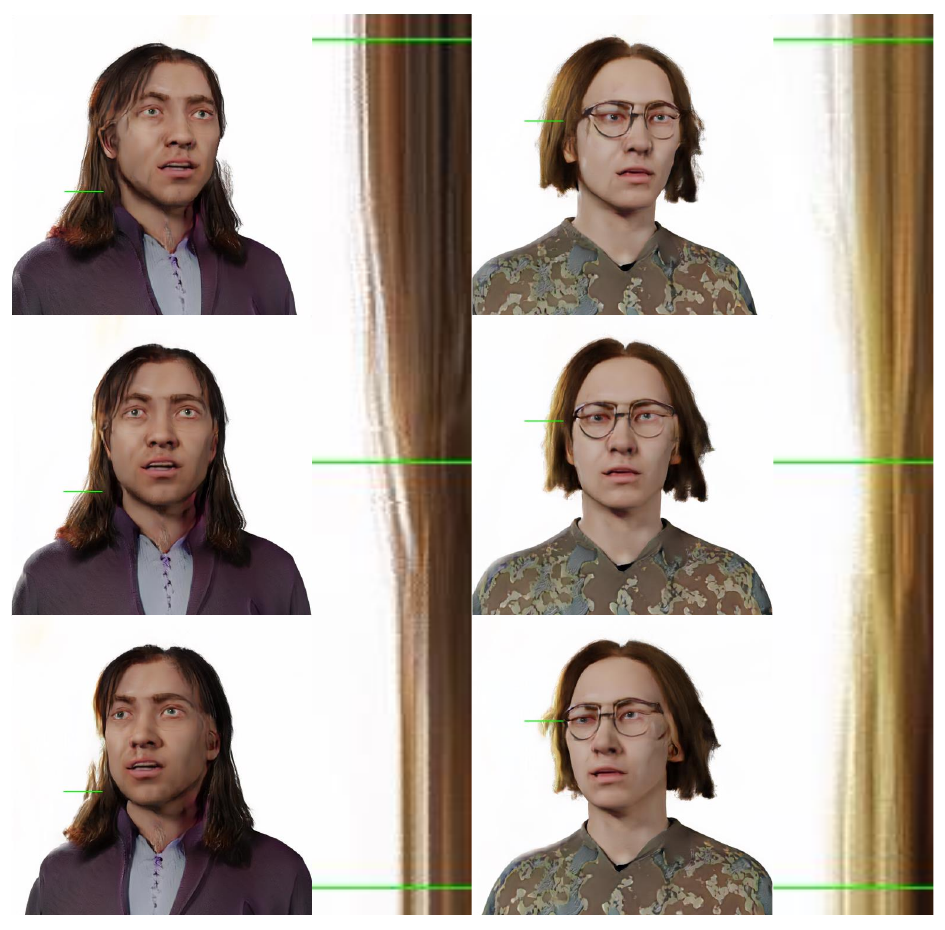} &
    \includegraphics[width=0.5\linewidth]{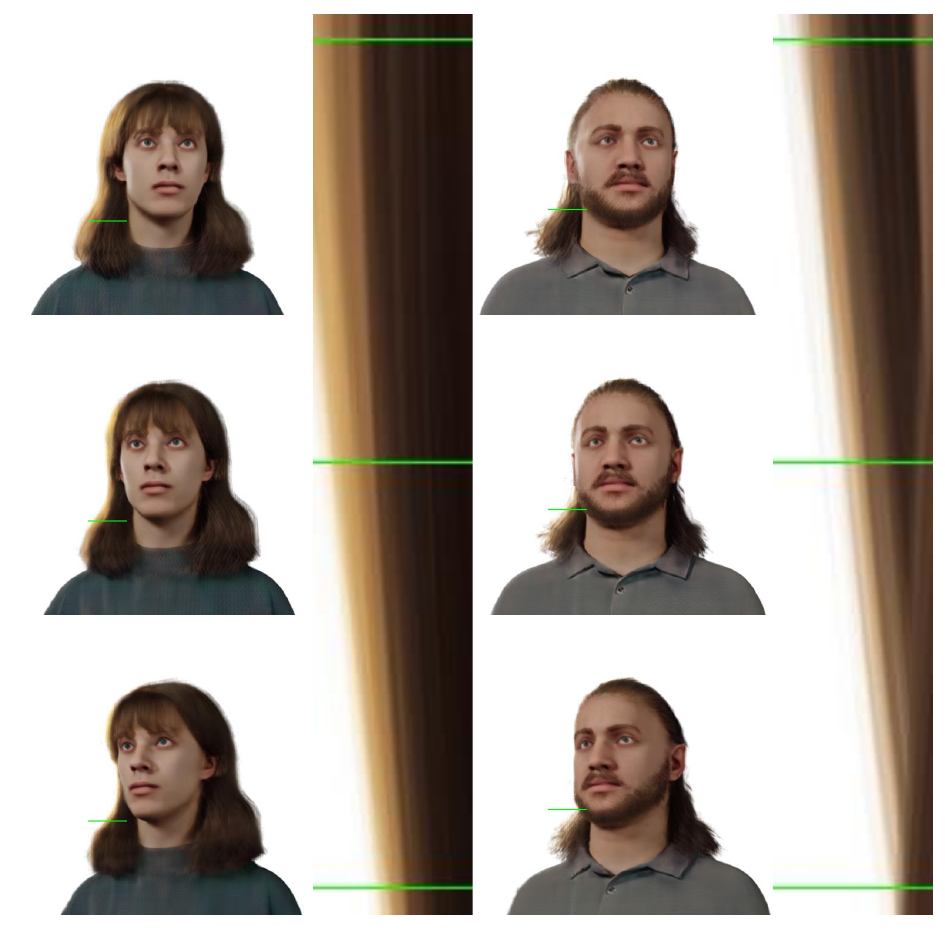} \\ 
    EG3D & \textbf{Our RodinHD}\\
  \end{tabular}
  \caption{Visual comparison of 3D consistency akin to the Epipolar Line Images~\cite{bolles1987epipolar}. Our model yields smooth and natural texture, whereas EG3D produces obvious texture flickering, indicating the 3D inconsistency with 2D refinement.}
\label{supp/fig:3d_consistency}
\vspace{-5mm}
\end{figure}

\noindent\textbf{Comparison of 3D consistency with EG3D.} We additionally compare our 3D consistency with SOTA 3D-aware GANs, EG3D. We evaluate the 3D consistency of unconditional generated results in~\cref{supp/fig:3d_consistency}, which is similar to Fig. 9 of main paper. Since EG3D also utilizes a 2D super-resolution module, the results in~\cref{supp/fig:3d_consistency} yield obvious texture flickering, whereas our method leads to a natural and smooth texture pattern. We also provide numerical comparison in~\cref{supp/table:3d_consistency} by fitting a NeuS model from generated multi-views following Tab. 3 of main paper. Our generated results achieve significantly better metrics due to multi-view consistency.

\begin{table}[t]
  \caption{Quantitative comparison of 3D consistency with EG3D.}
  \centering
  \small
  \setlength\tabcolsep{3pt}
  \begin{tabular}{lccc}
            \toprule  
             & PSNR$\uparrow$ & SSIM$\uparrow$ & LPIPS$\downarrow$ \\
            \midrule
            EG3D  & 29.51 & 0.962 & 0.052\\
            \cellcolor{gray}\textbf{Ours}  & \cellcolor{gray}\textbf{33.39} & \cellcolor{gray}\textbf{0.967}  & \cellcolor{gray}\textbf{0.043} \\
            \bottomrule
        \end{tabular}
  \label{supp/table:3d_consistency}
\end{table} 

\begin{table}[t]
  \caption{Average ranking of user study in conditional generation.}
  \centering
  \small
  \setlength\tabcolsep{3pt}
  \begin{tabular}{lccccccc}
  \toprule  
  Models & ID Similarity$\downarrow$ & 3D Consistency$\downarrow$ & Visual Fidelity$\downarrow$ \\
  \midrule
  Rodin & 2.54 & 2.12 & 2.61 \\
  Rodin$^*$ & 2.06 & 2.77 & \textbf{1.26} \\
  \cellcolor{gray}\textbf{Ours} & \cellcolor{gray}\textbf{1.39} & \cellcolor{gray}\textbf{1.11}& \cellcolor{gray}2.13\\
  \bottomrule
  \end{tabular}
 \label{supp/table:user_study}
 \vspace{-5mm}
\end{table} 

\section{Additional Results}

\begin{figure*}[t]
  \centering
  \small
  \setlength\tabcolsep{1pt}
  \begin{tabular}{c} 
    \includegraphics[width=1.0\linewidth]{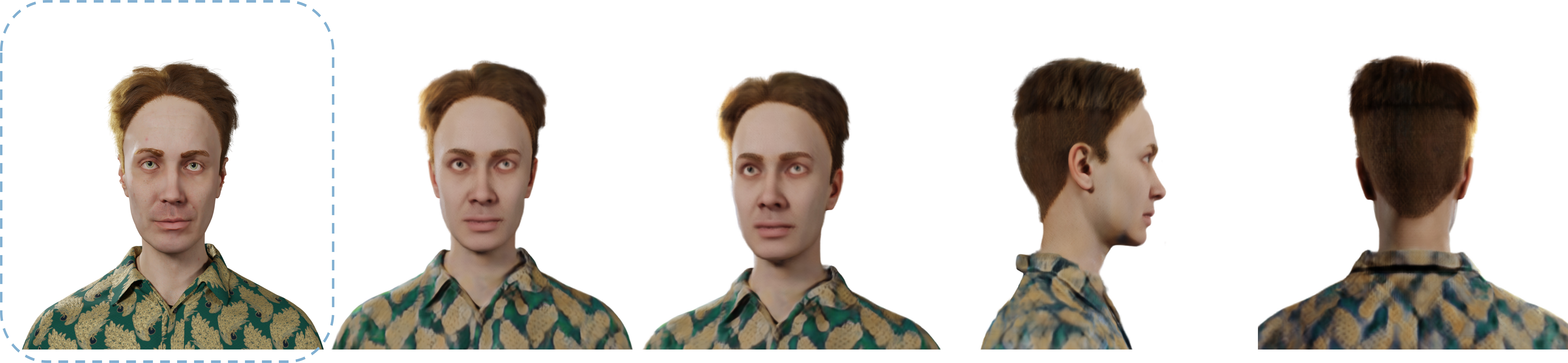} \\
    \includegraphics[width=1.0\linewidth]{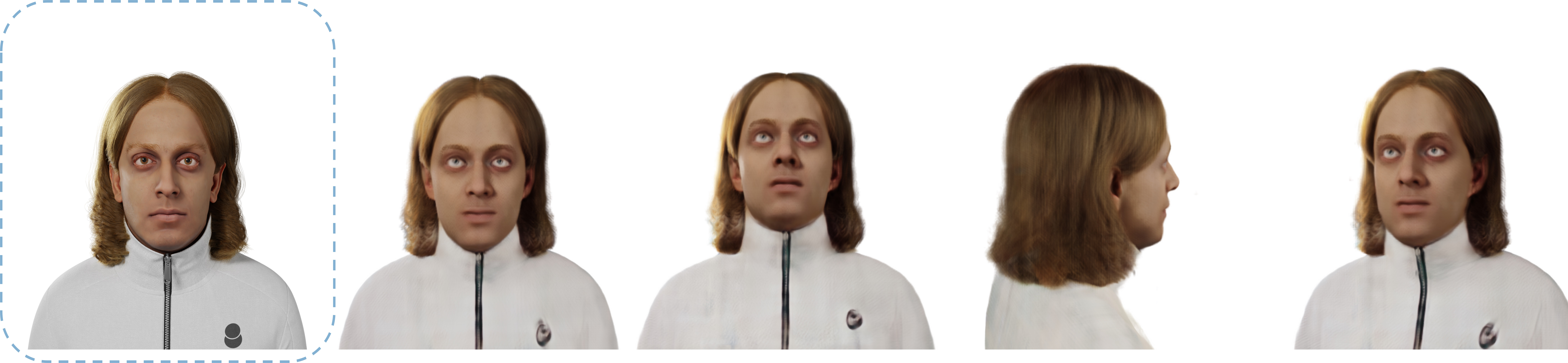} \\
    \includegraphics[width=1.0\linewidth]{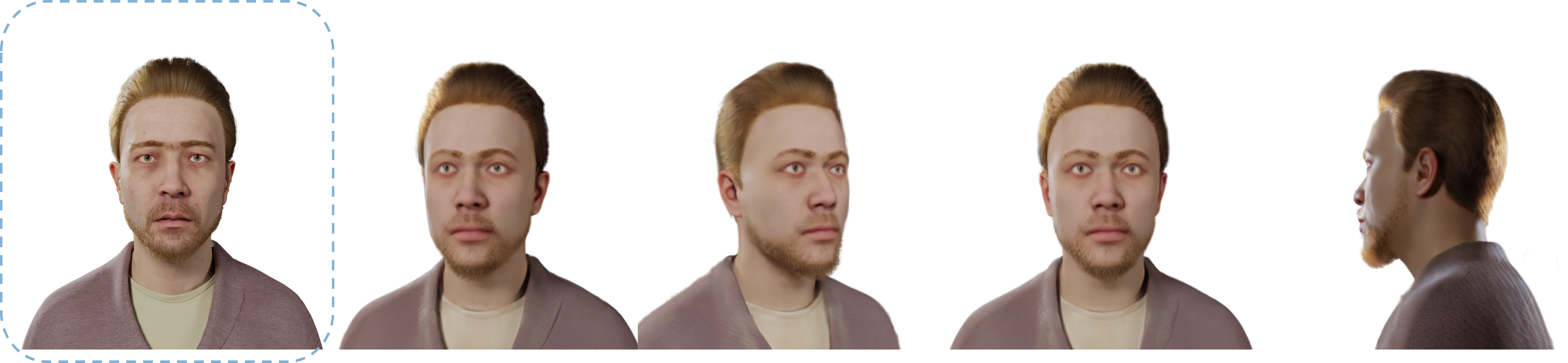} \\
    \includegraphics[width=1.0\linewidth]{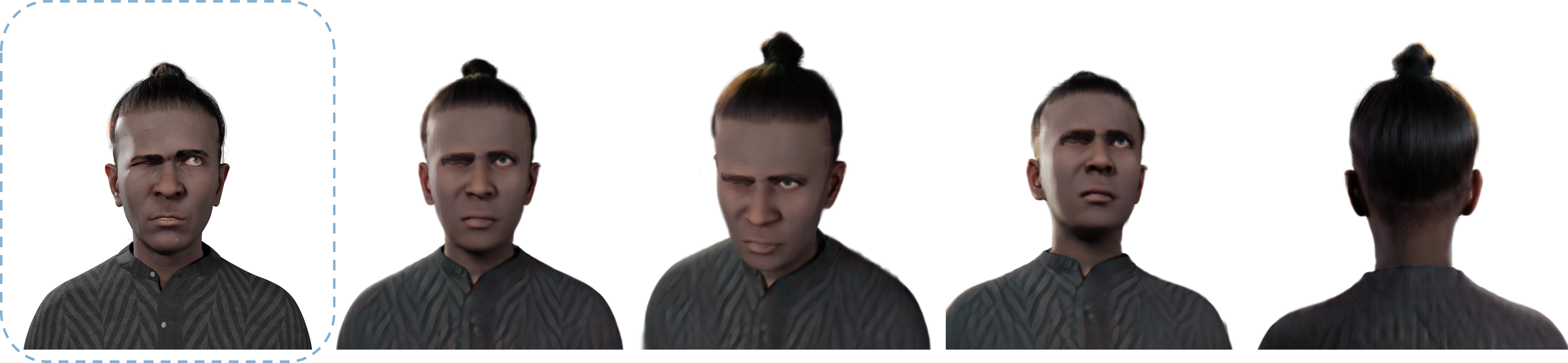} \\
    \includegraphics[width=1.0\linewidth]{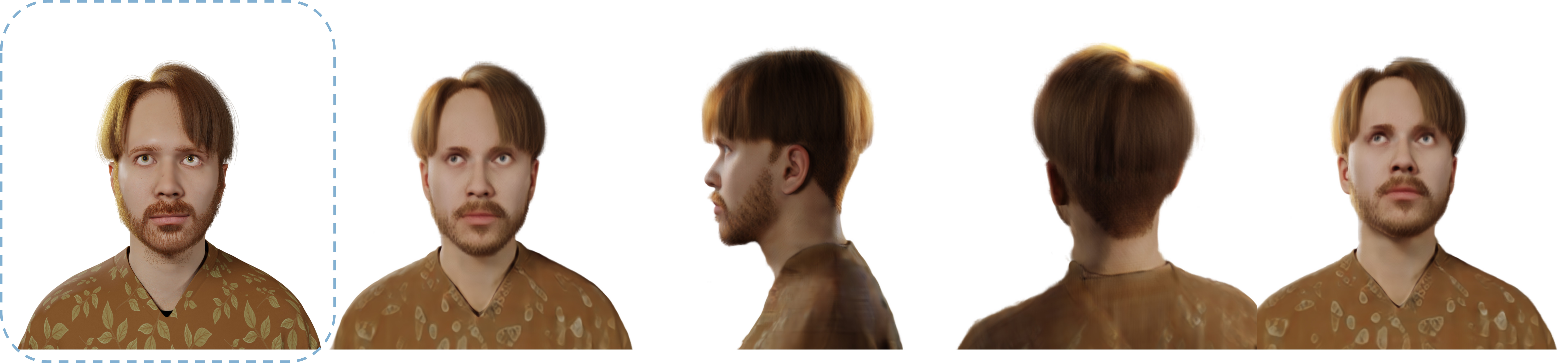} \\
  \end{tabular}
  \caption{Conditional generation samples by our model. Reference portraits are shown in dashed boxes.}
  \label{fig/supp:cond_vis}
\end{figure*}

\noindent\textbf{Conditional avatar generation.} We provide more renderings of generated avatars conditioned on the single portraits from our test set in~\cref{fig/supp:cond_vis}. Our model is capable of creating high-fidelity avatars with compelling details and vivid expressions, demonstrating the strong capability of the proposed model.

\begin{figure*}[th]
  \centering
  \small
  \setlength\tabcolsep{1pt}
  \begin{tabular}{cc} 
    \includegraphics[width=0.5\linewidth]{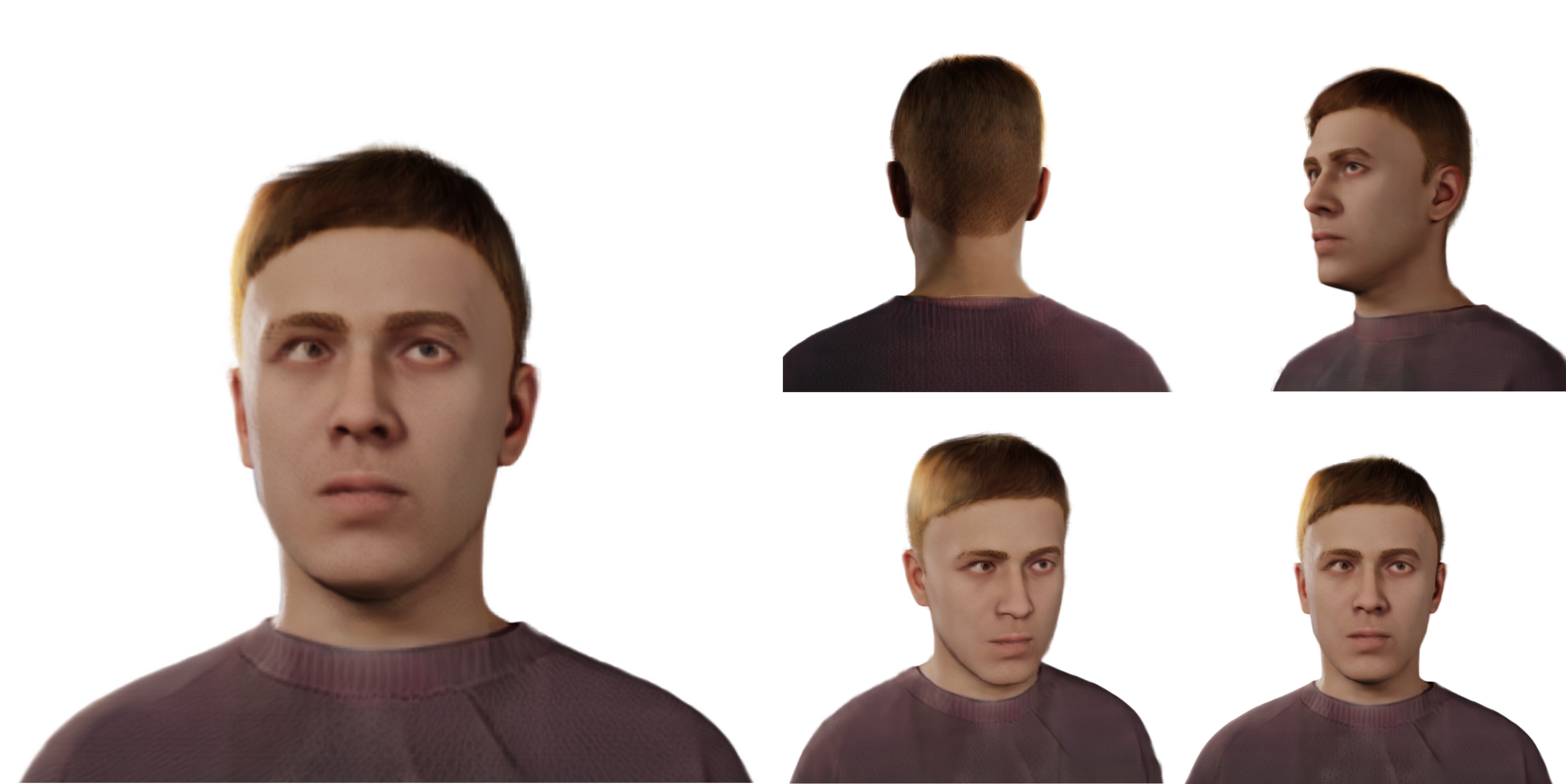}  &
    \includegraphics[width=0.5\linewidth]{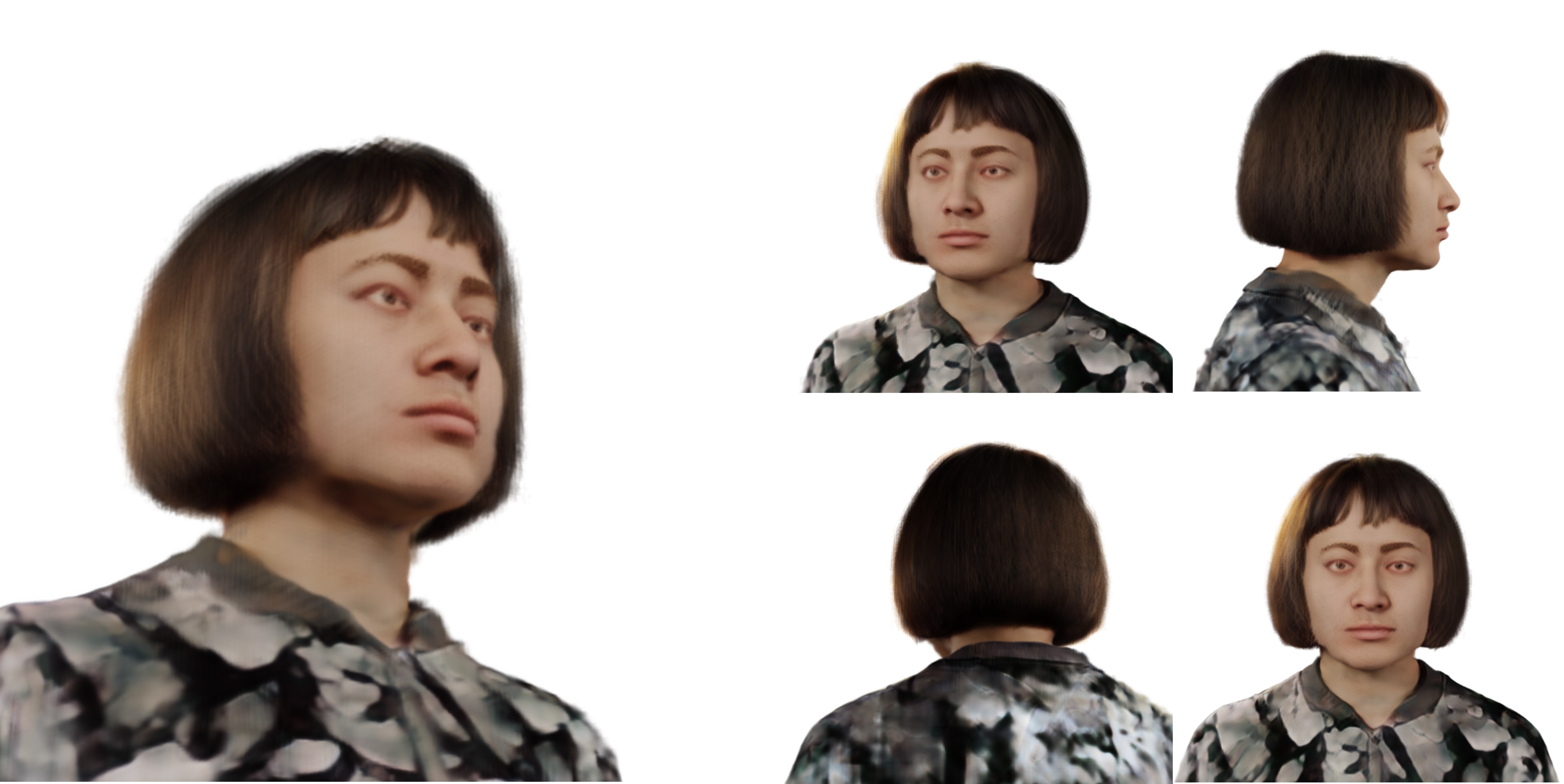} \\
     \includegraphics[width=0.5\linewidth]{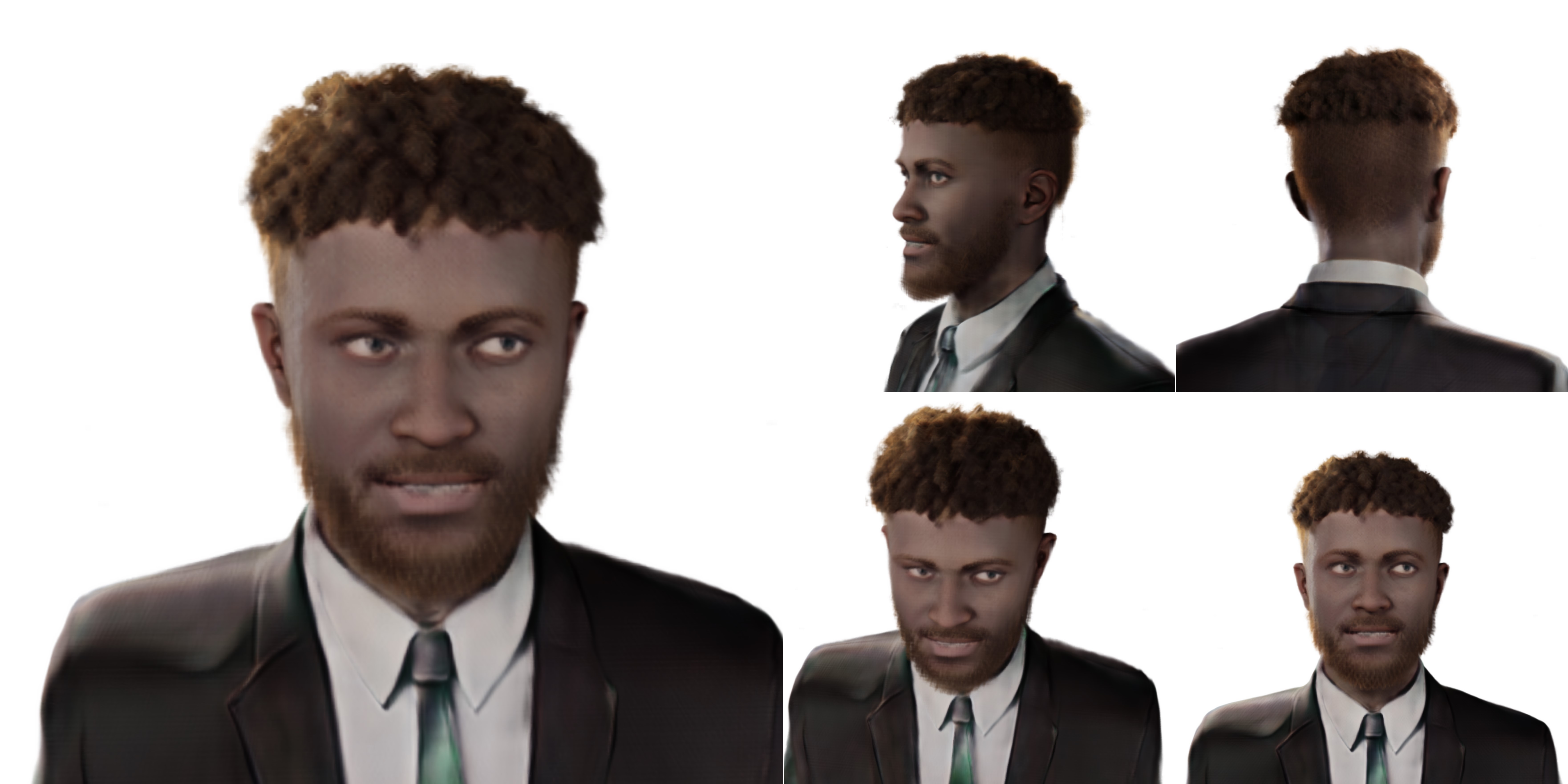}  &
    \includegraphics[width=0.5\linewidth]{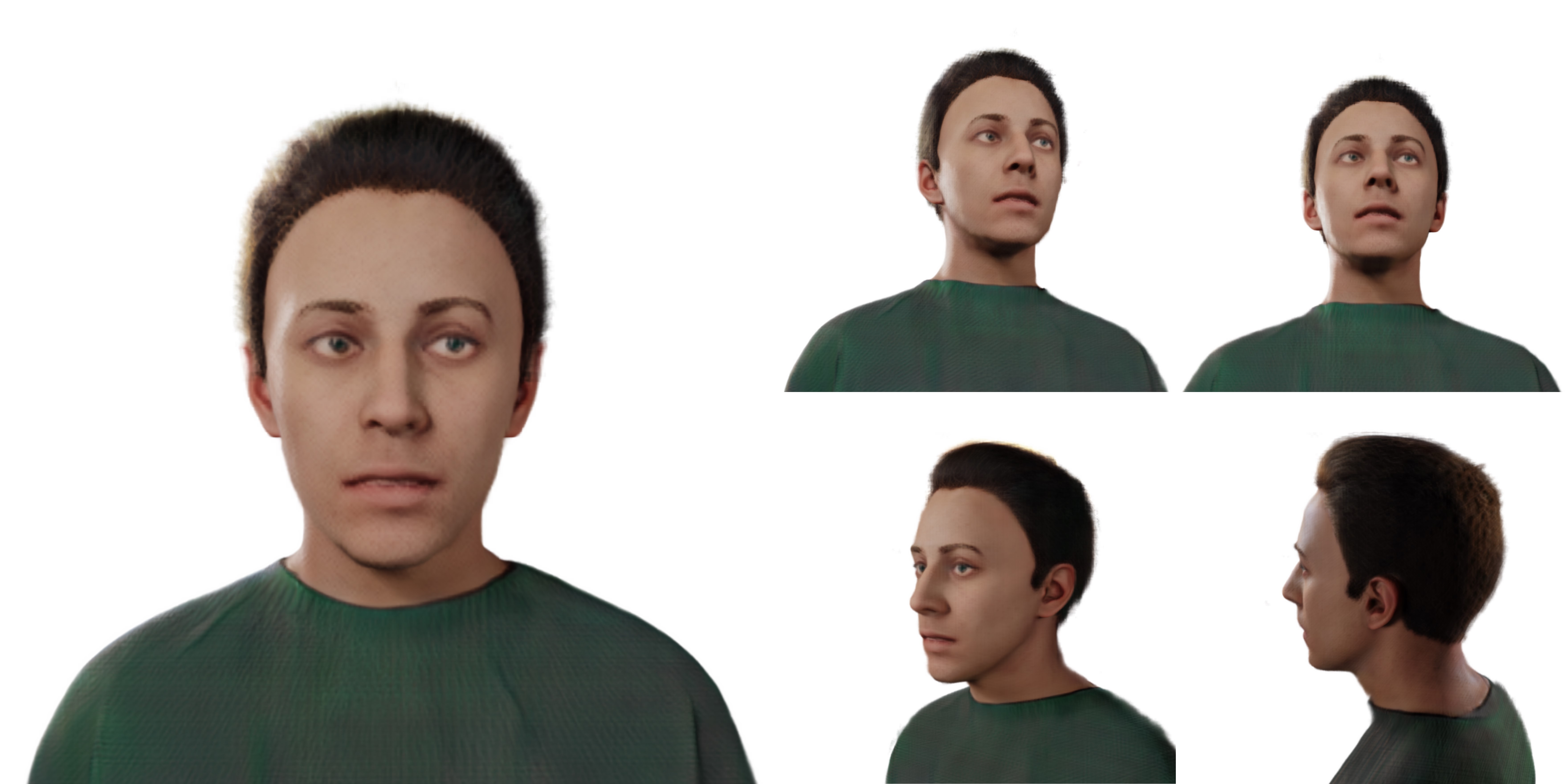} \\
    \includegraphics[width=0.5\linewidth]{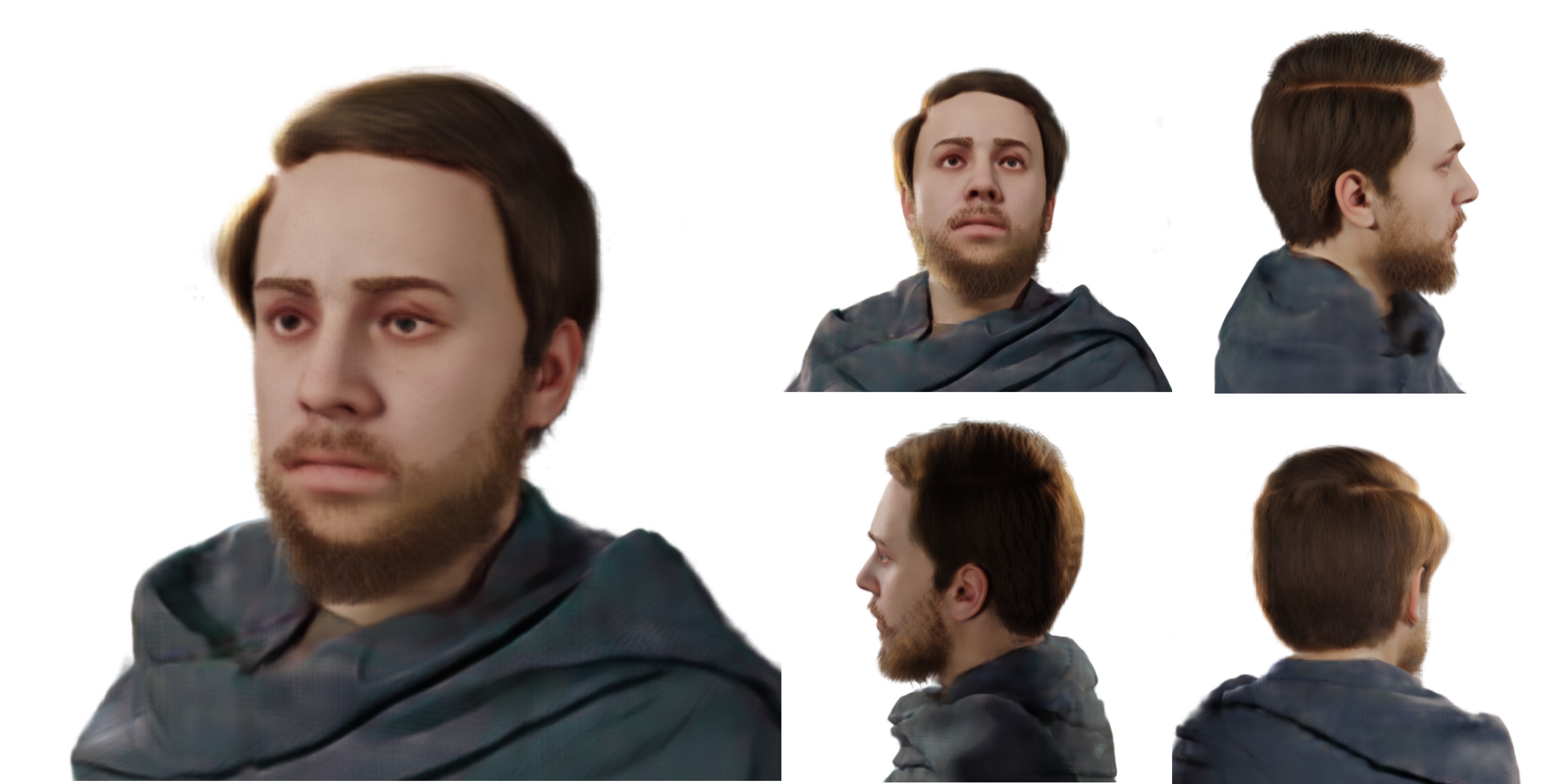}  &
    \includegraphics[width=0.5\linewidth]{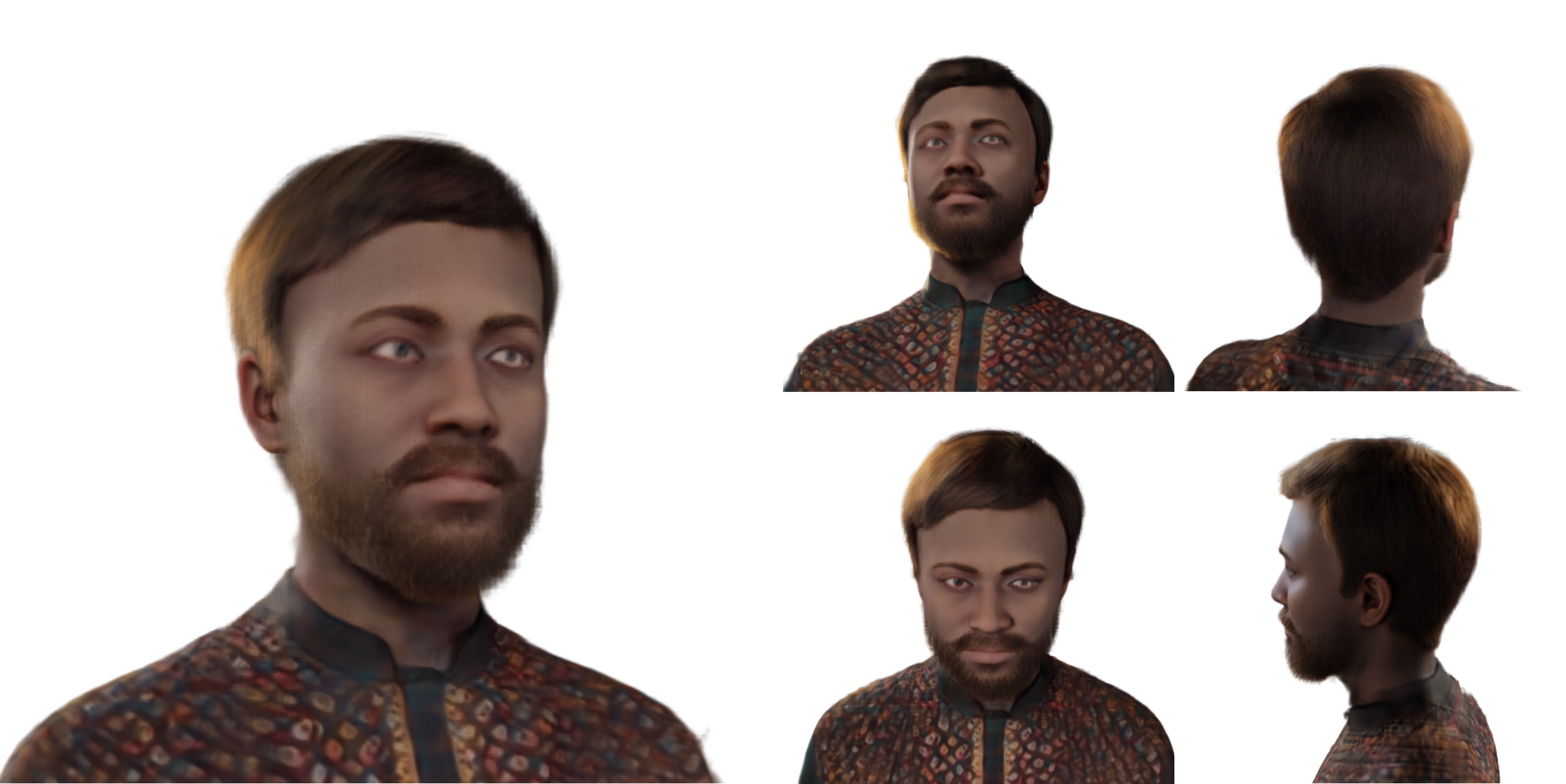} \\
    \includegraphics[width=0.5\linewidth]{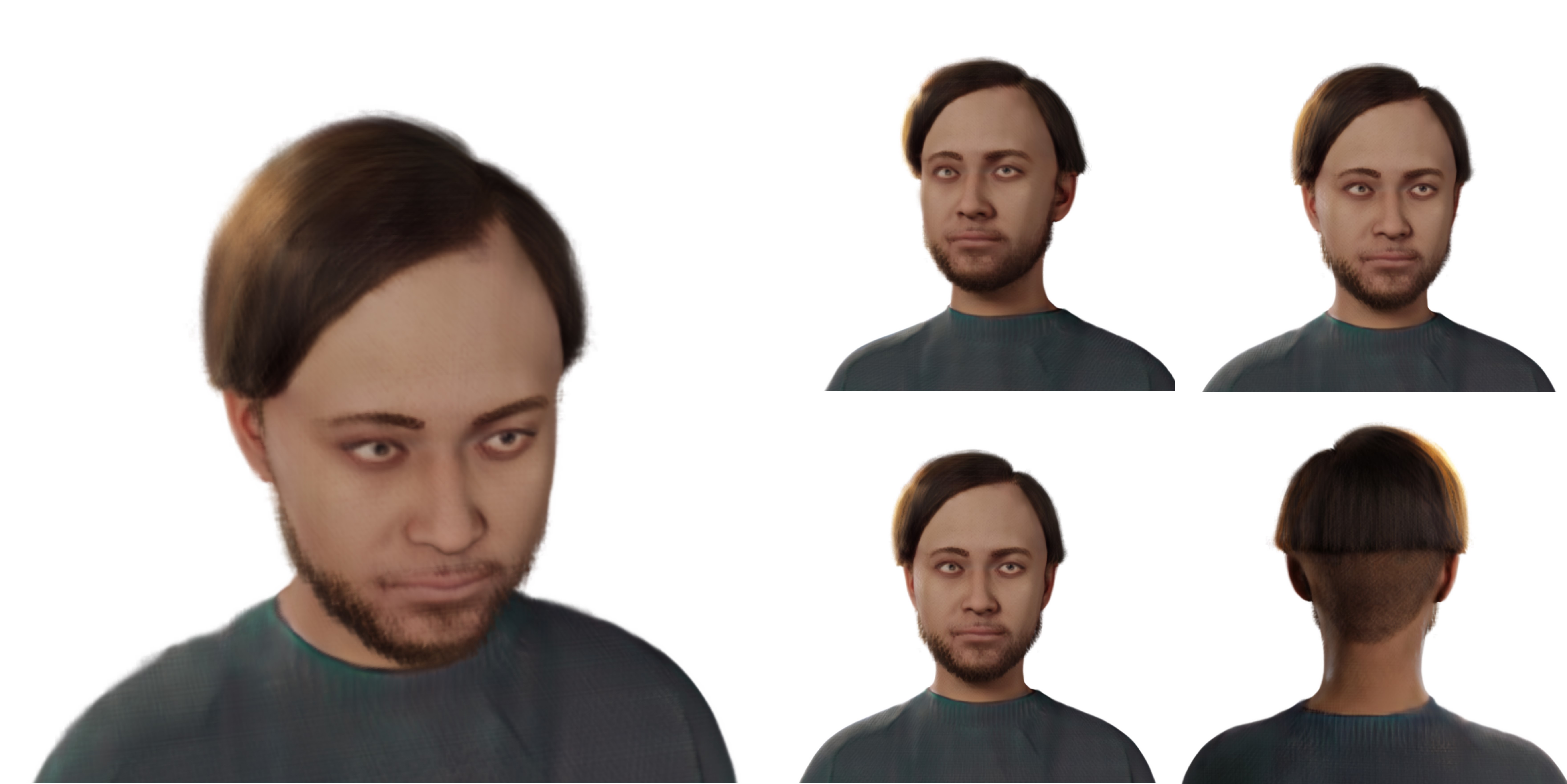}  &
    \includegraphics[width=0.5\linewidth]{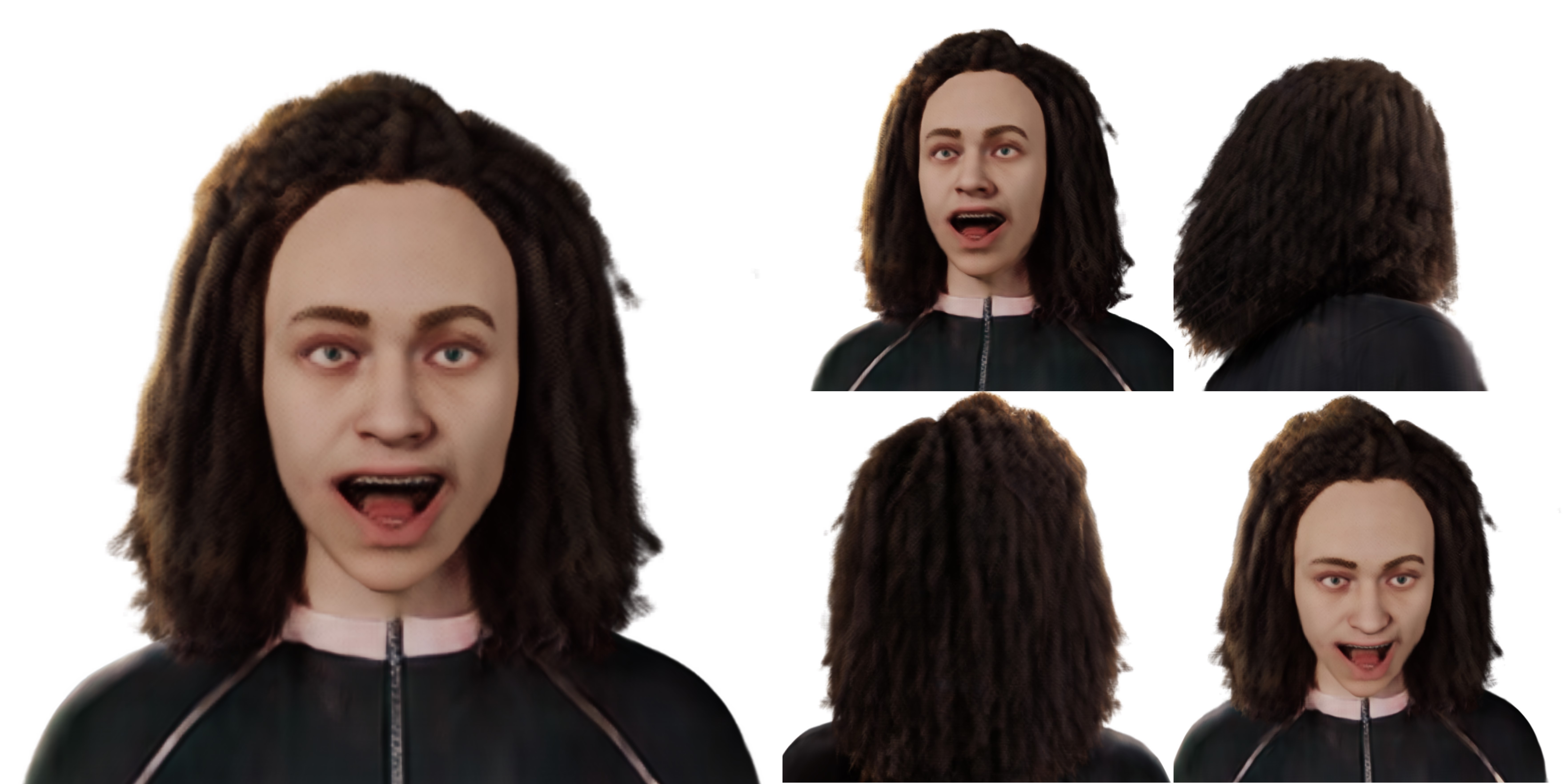} \\
  \end{tabular}
  \caption{Unconditional generation samples by our model.}
  \label{fig/supp:uncond_vis}
\end{figure*}

\noindent\textbf{Unconditional avatar generation.} ~\cref{fig/supp:uncond_vis} show more unconditional avatars created by our model. Our model is able to produce diverse high-quality avatars with rich details, including complex clothing and hairstyles.

\begin{figure*}[t]
    \small
    \centering
    \begin{tabular}{c@{\hspace{3mm}}c@{\hspace{1mm}}c@{\hspace{3mm}}c@{\hspace{1mm}}c@{\hspace{3mm}}} 
    \includegraphics[width=0.15\linewidth]{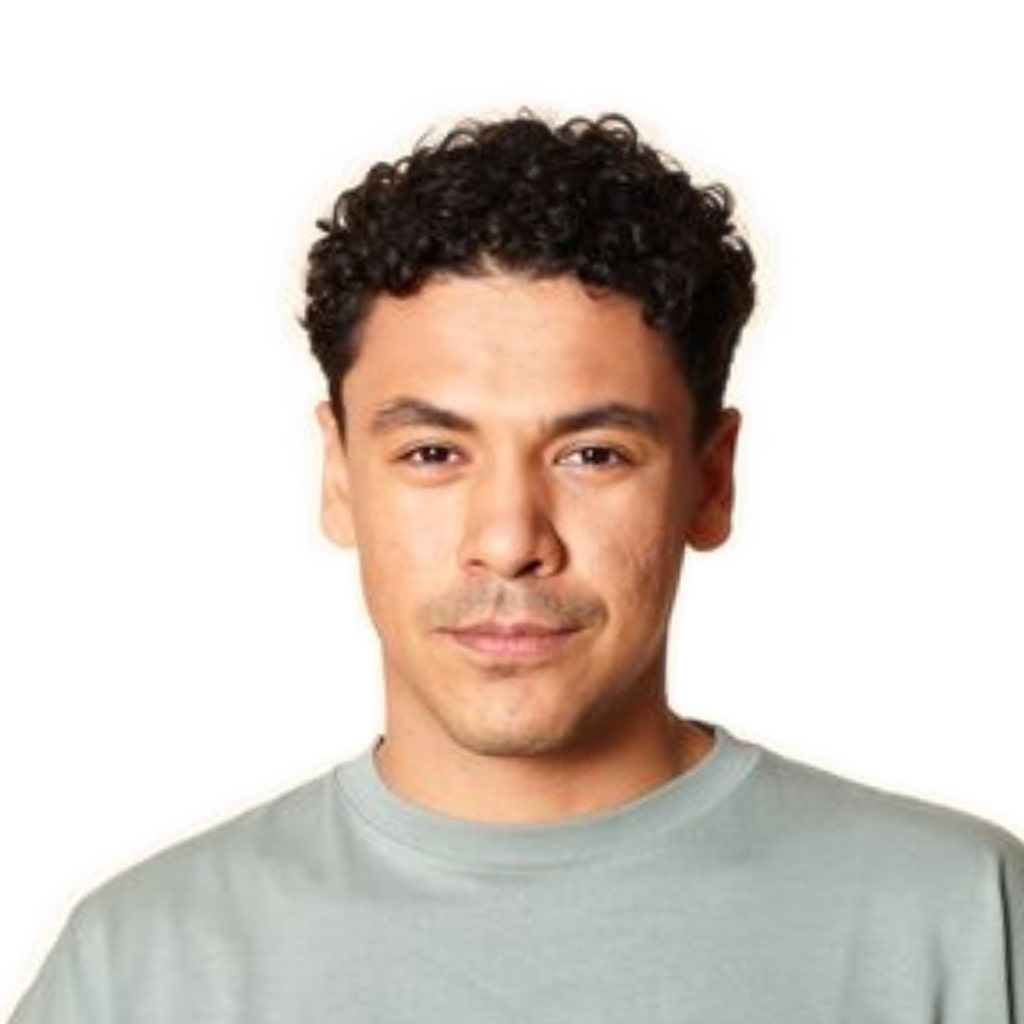}  &
    \includegraphics[width=0.15\linewidth]{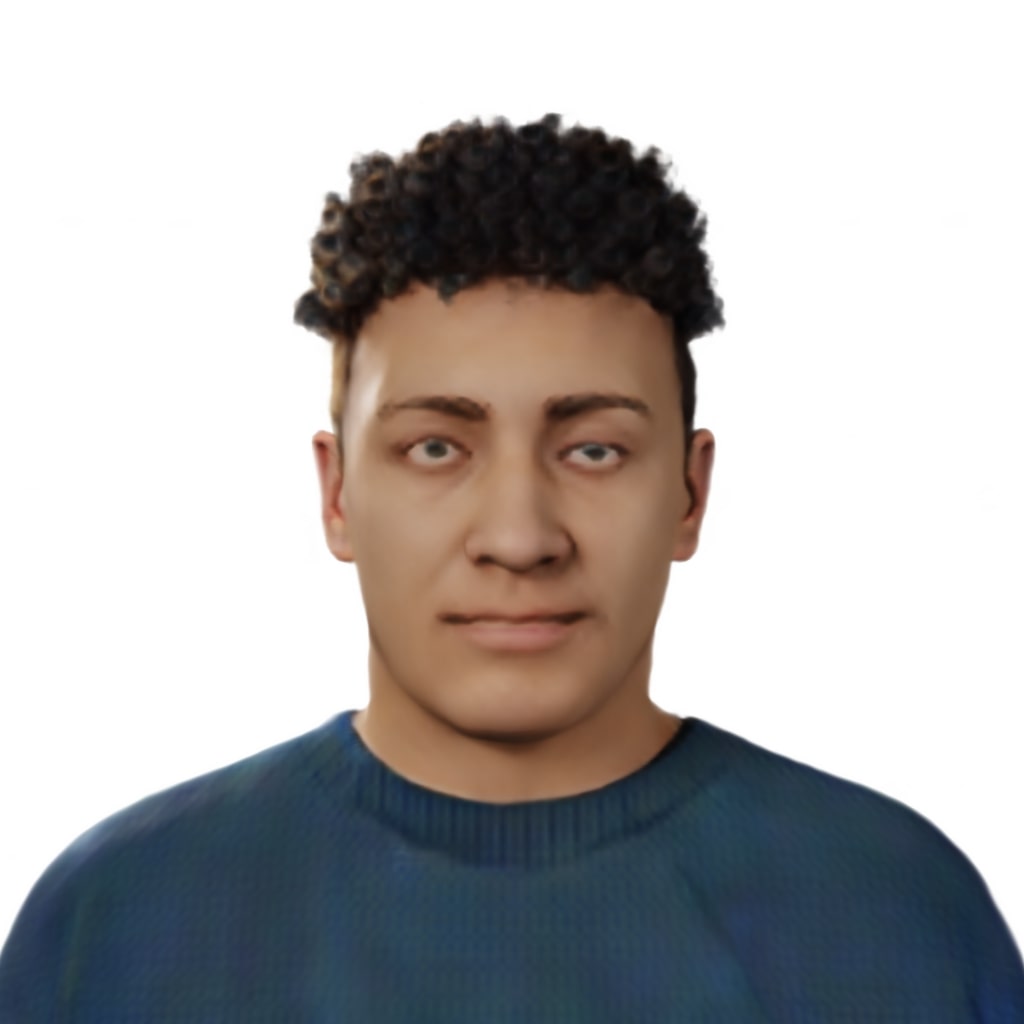} &
    \includegraphics[width=0.15\linewidth]{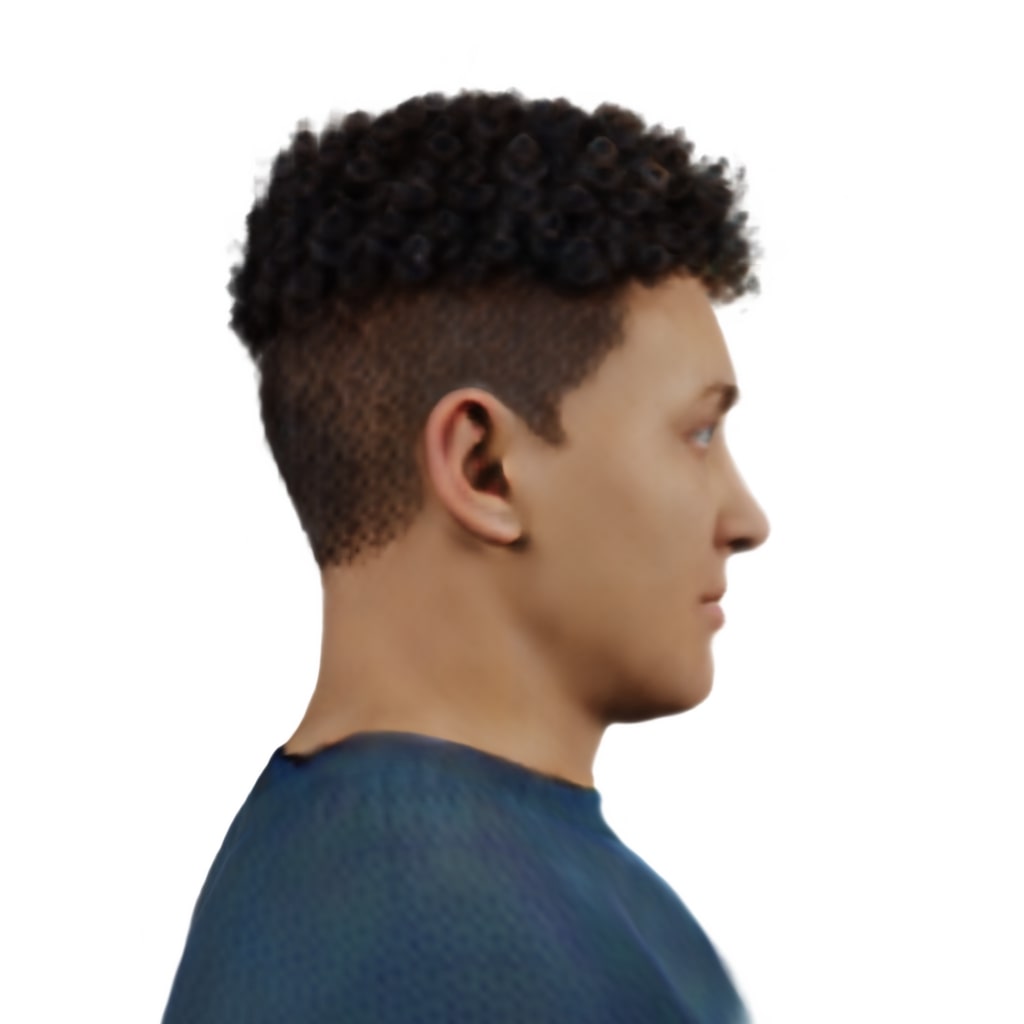} &
    \includegraphics[width=0.15\linewidth]{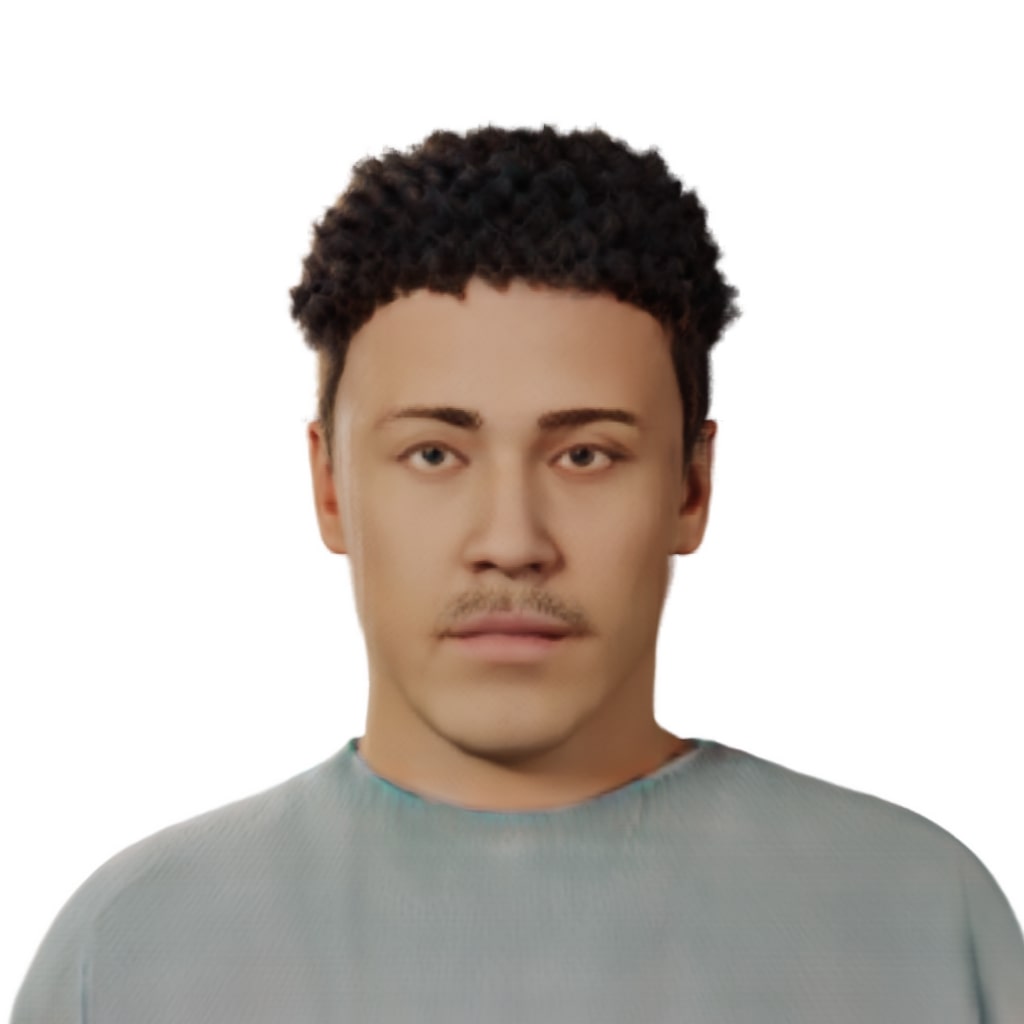} &
    \includegraphics[width=0.15\linewidth]{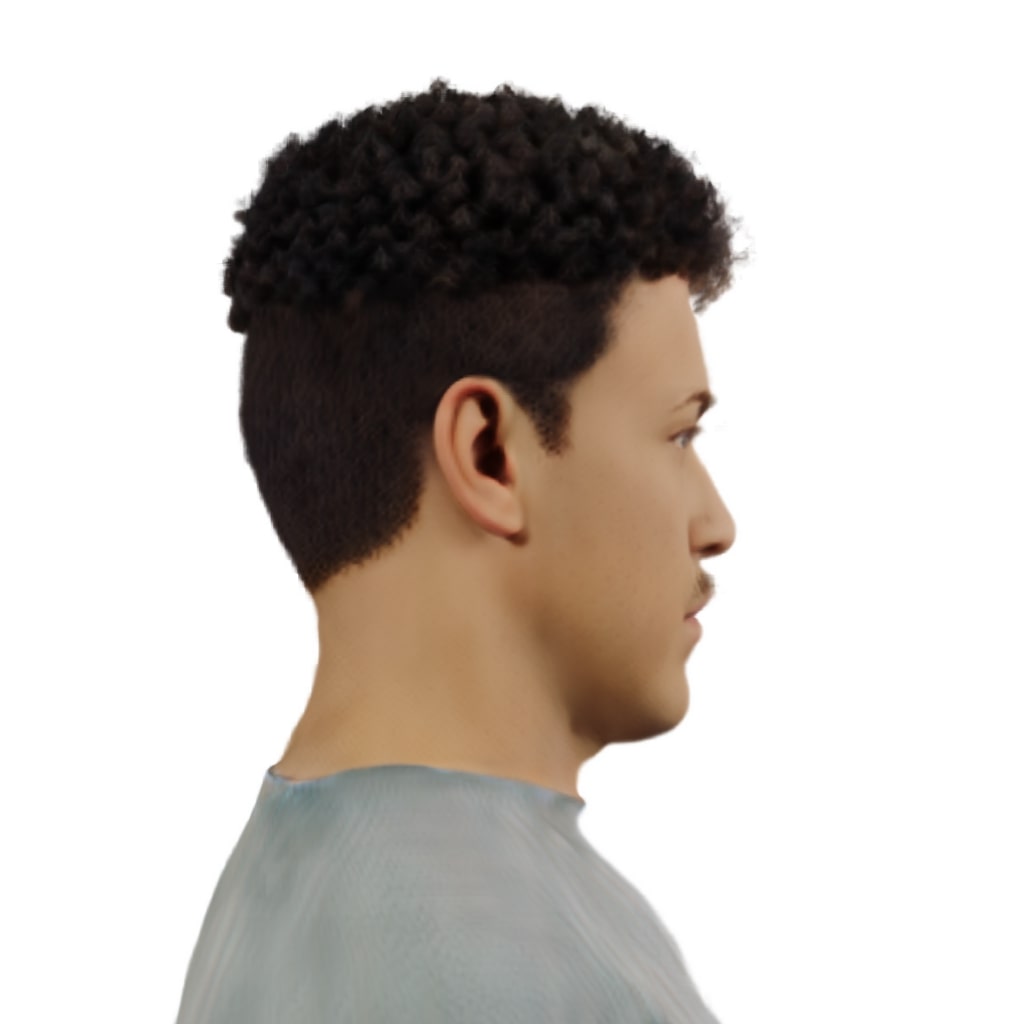} \\
    \includegraphics[width=0.15\linewidth]{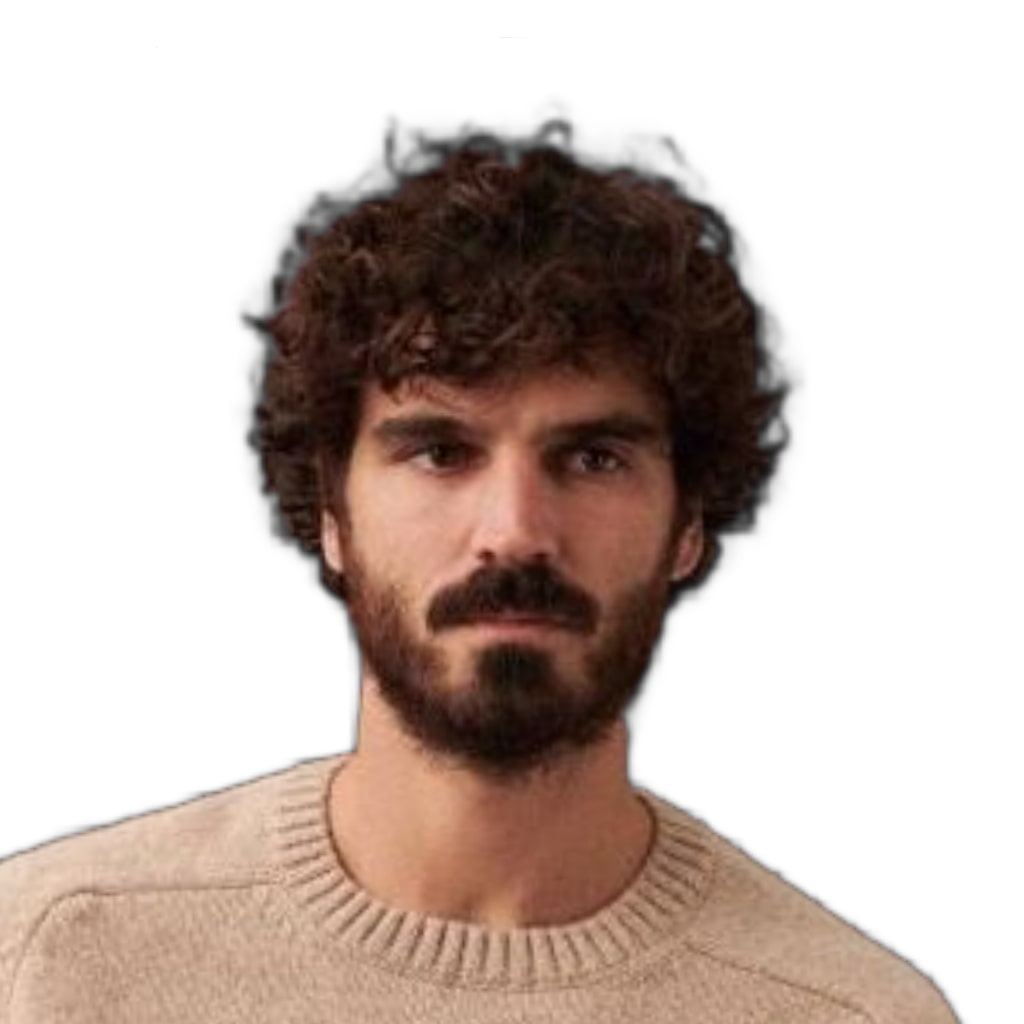}  &
    \includegraphics[width=0.15\linewidth]{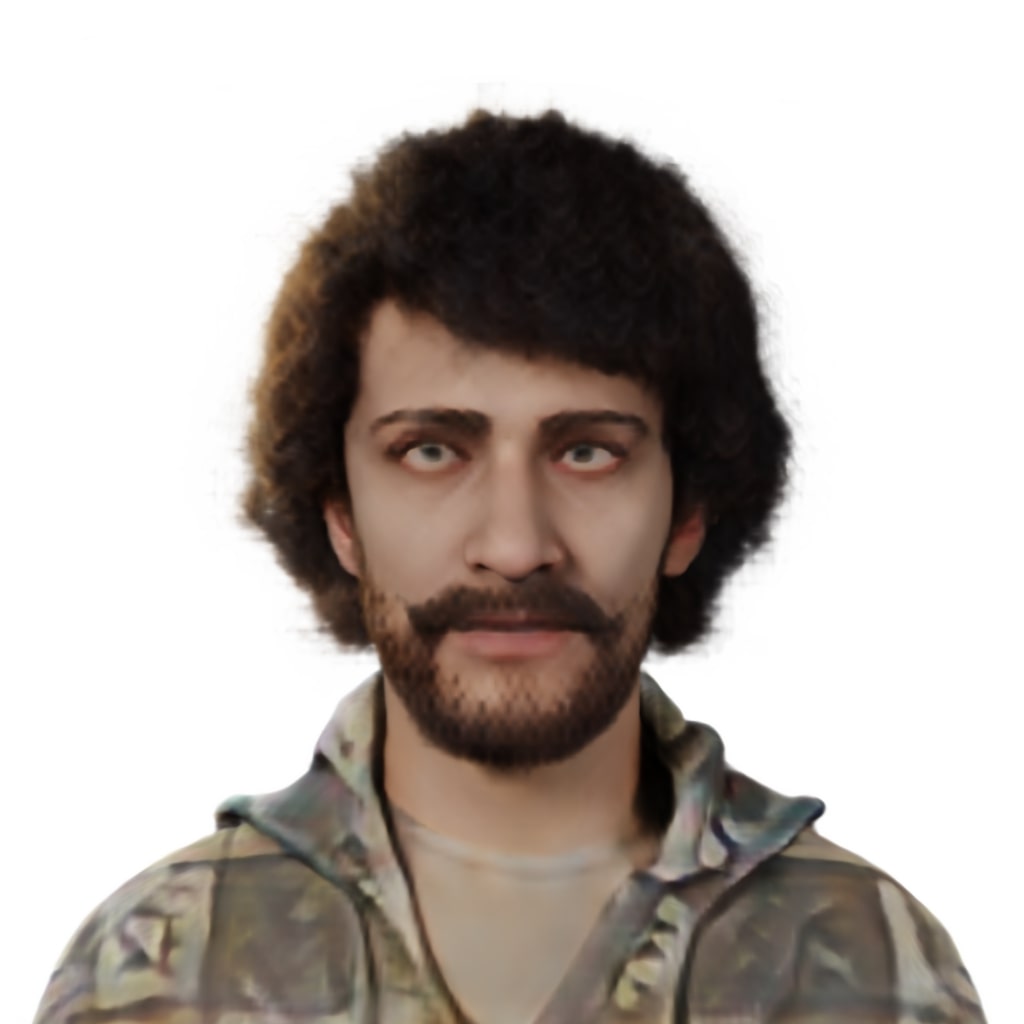} &
    \includegraphics[width=0.15\linewidth]{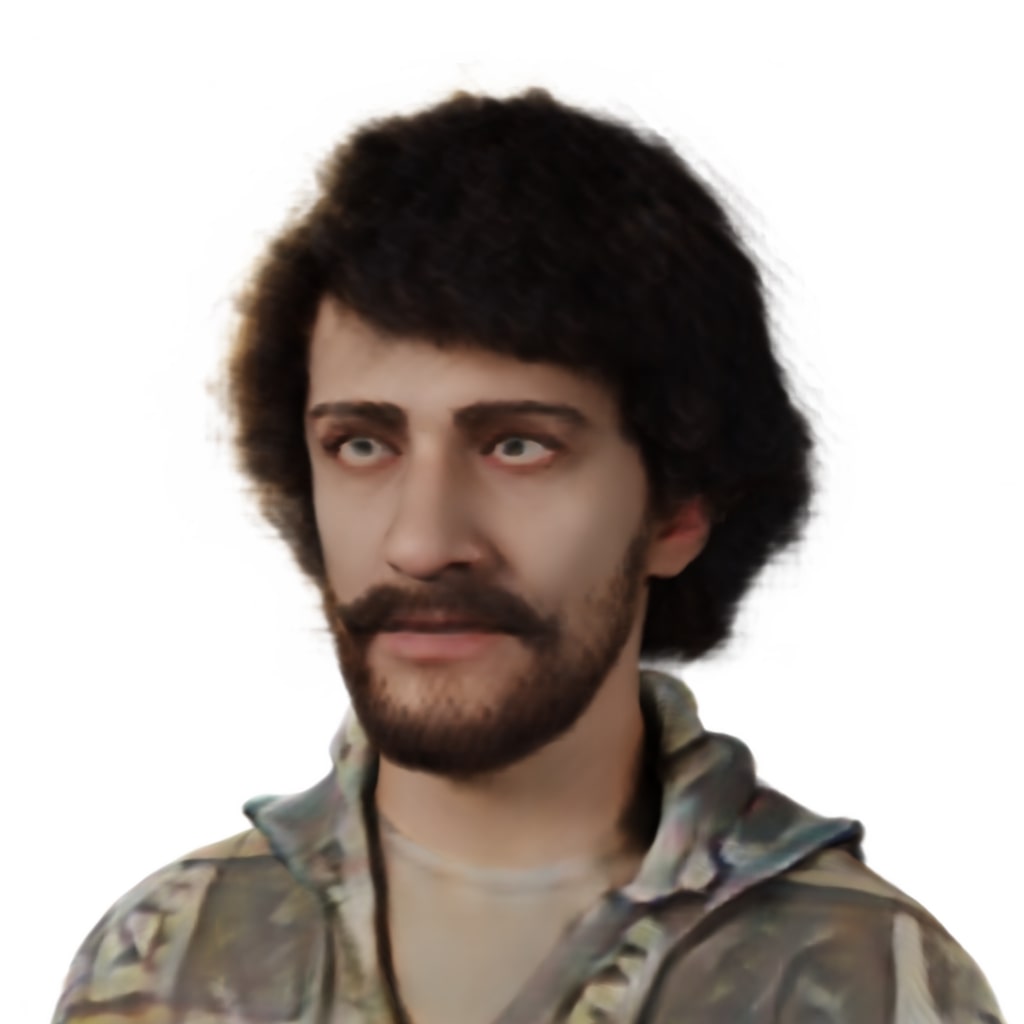} &
    \includegraphics[width=0.15\linewidth]{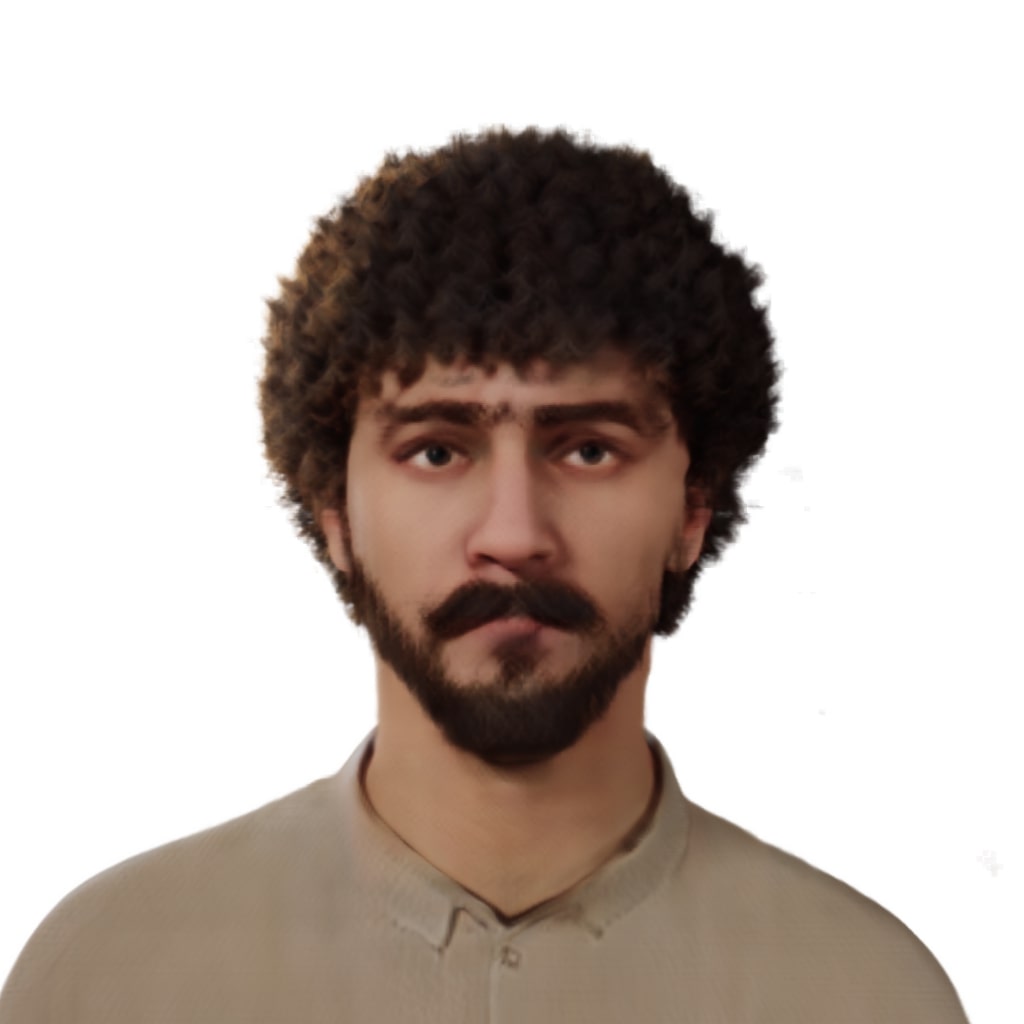} &
    \includegraphics[width=0.15\linewidth]{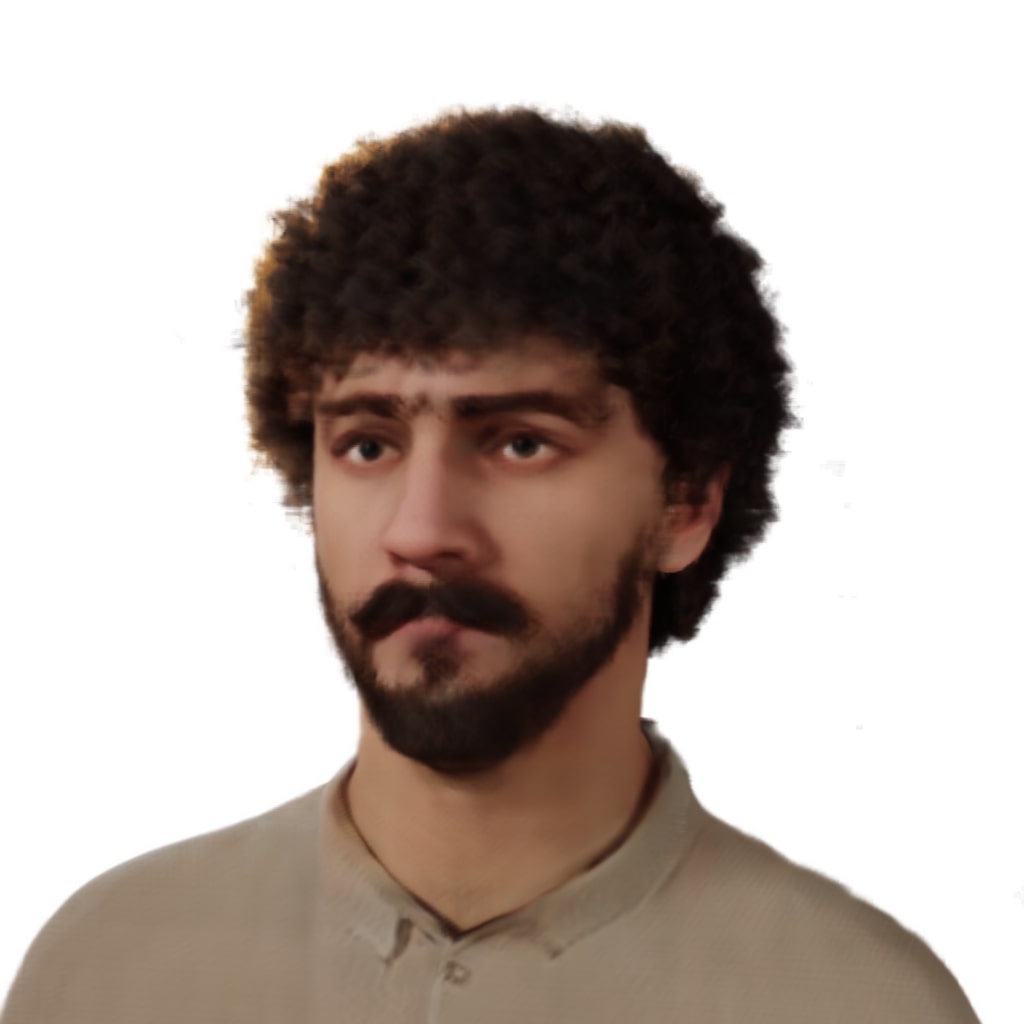} \\
    Reference & \multicolumn{2}{c}{Rodin} & \multicolumn{2}{c}{\textbf{Our RodinHD}}\\
  \end{tabular}
  \captionof{figure}{Samples of generated avatars conditioned on single in-the-wild portraits. Compared with Rodin, our method preserves more details of identity and clothing.}
  \label{fig/supp:real_world_dataset}
\end{figure*}

\noindent\textbf{Avatar creation from in-the-wild portrait.} In~\cref{fig/supp:real_world_dataset}, we present additional generated avatars conditioned on real-world images. Our methodology demonstrates a higher fidelity in preserving the identity of the subjects when compared with~\cite{wang2023rodin}. Furthermore, our results exhibit a remarkable ability to retain intricate details such as hairstyle and clothing attributes.

\begin{figure*}[th]
  \centering
  \small
  \setlength\tabcolsep{1pt}
  \begin{tabular}{c} 
    \includegraphics[width=0.8\linewidth]{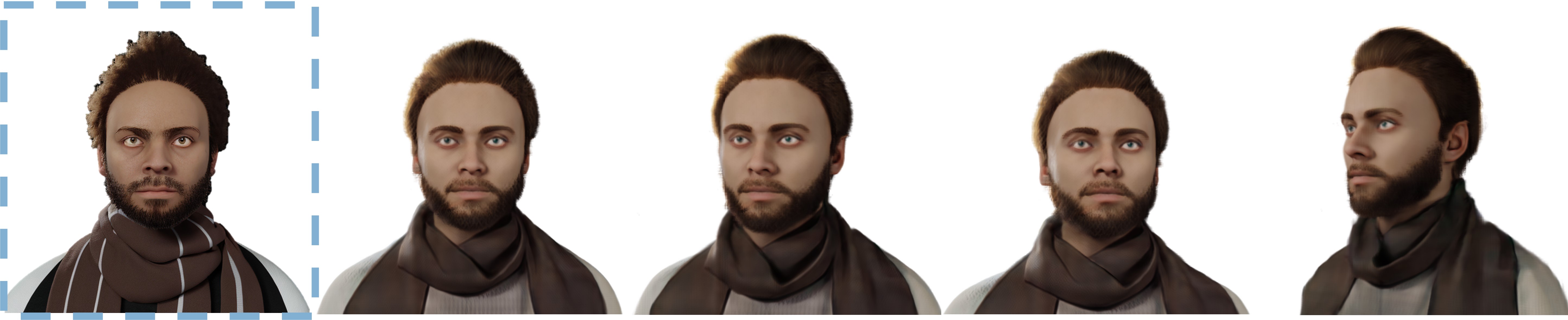} \\
    ``\textit{Blender Synthetic Avatar, brown hair, boy, brown eyes, scarf, beard, stubble}'' \\
    \includegraphics[width=0.8\linewidth]{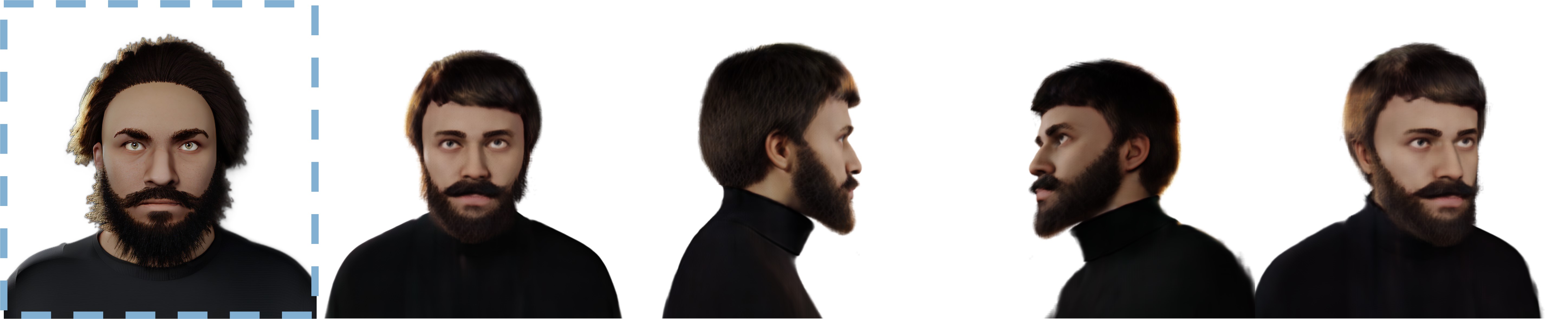} \\
    ``\textit{Blender Synthetic Avatar, brown hair, boy, brown eyes, beard, black sweater}'' \\
  \end{tabular}
  \caption{Samples of text-to-avatar creation using our model. The leftmost reference portraits are first created by finetuned 2D diffusion model given the text prompts. Then our 3D diffusion model creates 3D avatars conditioned on the generated portraits.}
  \label{fig/supp:text_cond_vis}
\end{figure*}

\noindent\textbf{Text-to-avatar creation.} We provide more samples of high-quality text-to-avatar creation in~\cref{fig/supp:text_cond_vis}. We first convert the text prompt to reference portrait by our finetuned 2D text-to-image diffusion models, thereafter generate a high-fidelity avatar conditioned on the reference portrait. It is worth noticing that the trigger word we used ``Blender Synthetic Avata'' is not necessarily needed to be added in the prompts since we can omit it and perform cropping and alignment to the generated images, similar to how we handle realistic image inputs.

\noindent\textbf{User study.} We further conduct user study to measure the identity similarity (ID), 3D consistency and visual fidelity. We ask 15 subjects to rank different methods with 20 sets of comparisons in each study. The average ranking in~\cref{supp/table:user_study} shows that our method earns user preferences the best in identity preservation and 3D consistency, only slightly worse than Rodin adding 2D refinement (Rodin$^*$) in fidelity. We think more follow-up research can be conducted to further improve the visual quality while ensuring the 3d consistency.

\section{Responsible AI Considerations}

Our model is trained on the synthetic dataset~\cite{wood2021fake} of 3D digital avatars akin to those crafted by artists, as opposed to photo-realistic humans. This approach to training data selection alleviates privacy and copyright concerns associated with the use of real human face collections. Despite these precautions, it is important to acknowledge that 3D avatar created by our model from real-world images could potentially be exploited for the dissemination of disinformation, similar to other generative models. We must therefore emphasize the importance of responsible use of our technology. As a safeguard against misuse, we recommend the implementation of measures such as embedding visible tags or watermarks into the distributed renderings produced by our model.

\section{Limitation}
\begin{figure*}[th!]
  \centering
  \small
  \setlength\tabcolsep{1pt}
  \begin{tabular}{cc@{\hspace{3mm}}cc} 
    \includegraphics[width=0.18\linewidth]{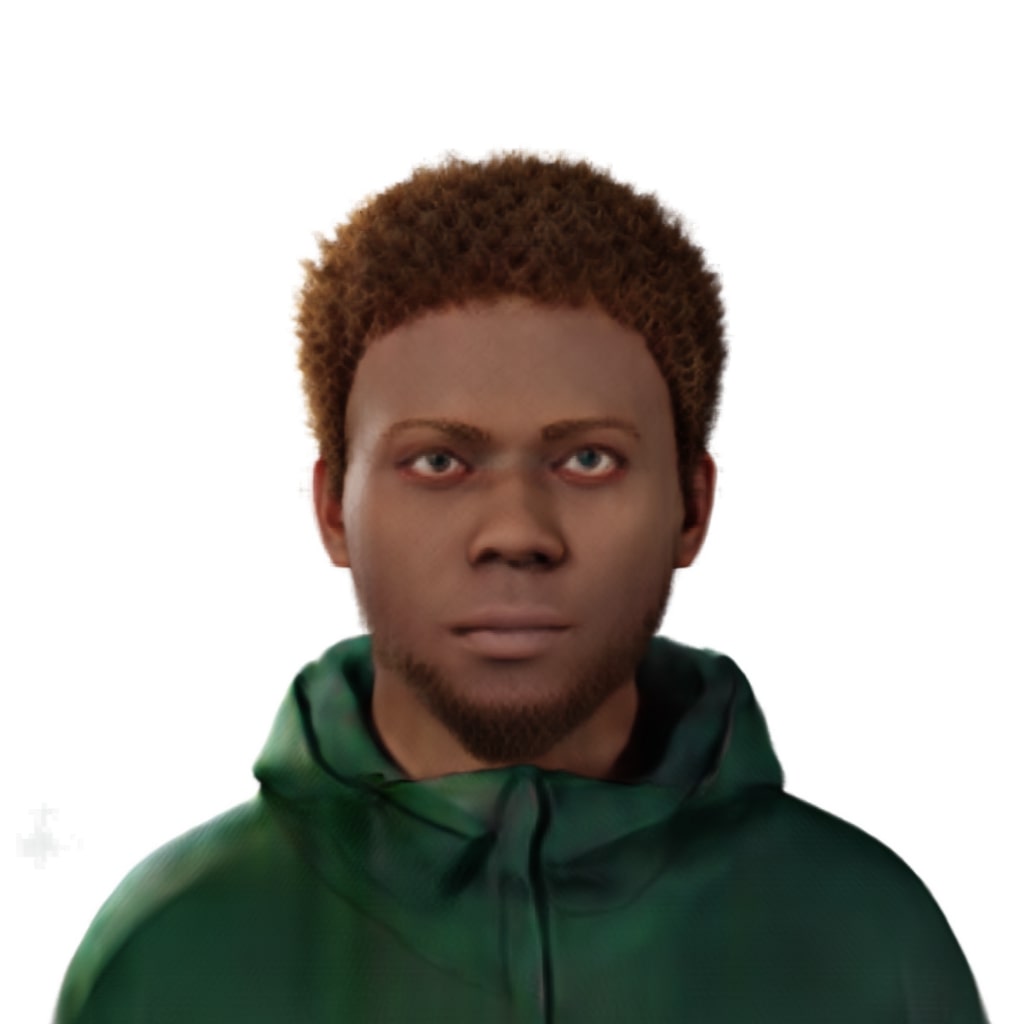} &
    \includegraphics[width=0.18\linewidth]{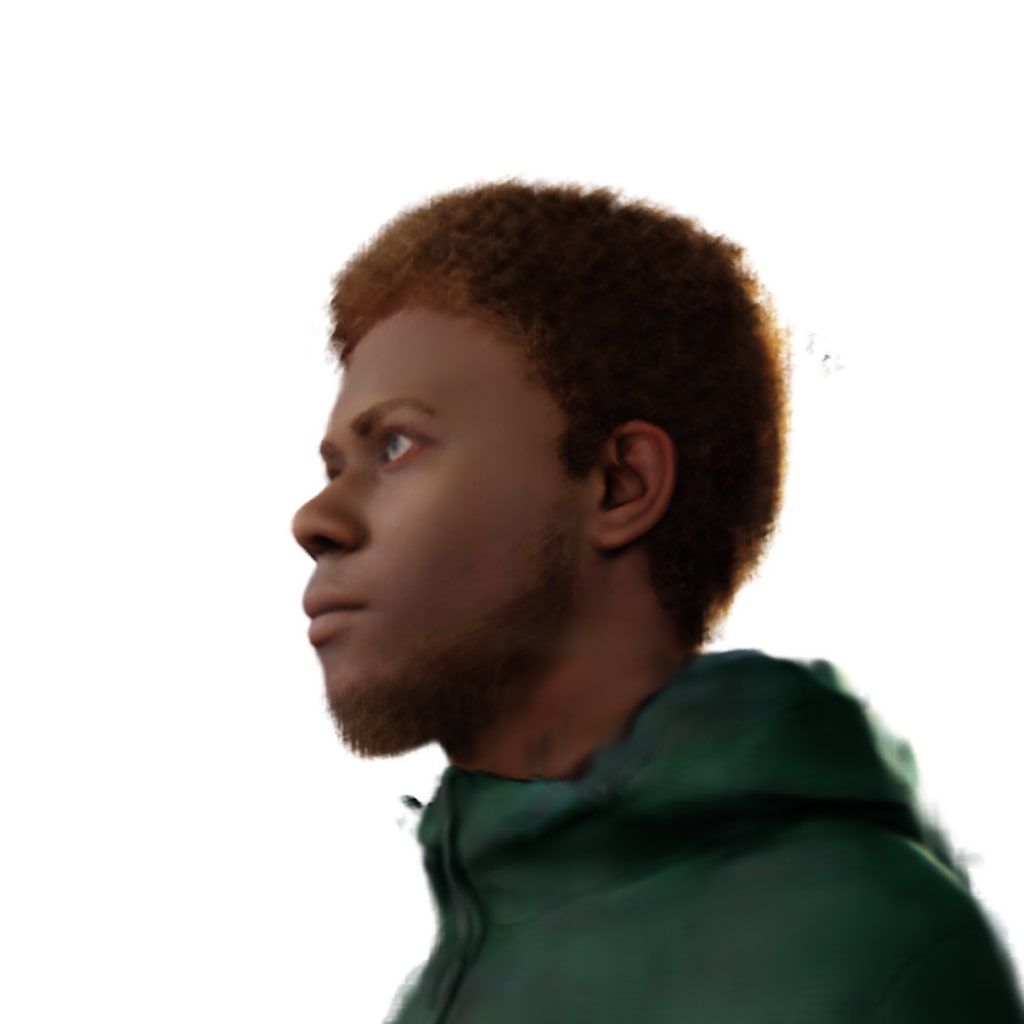} &
    \includegraphics[width=0.18\linewidth]{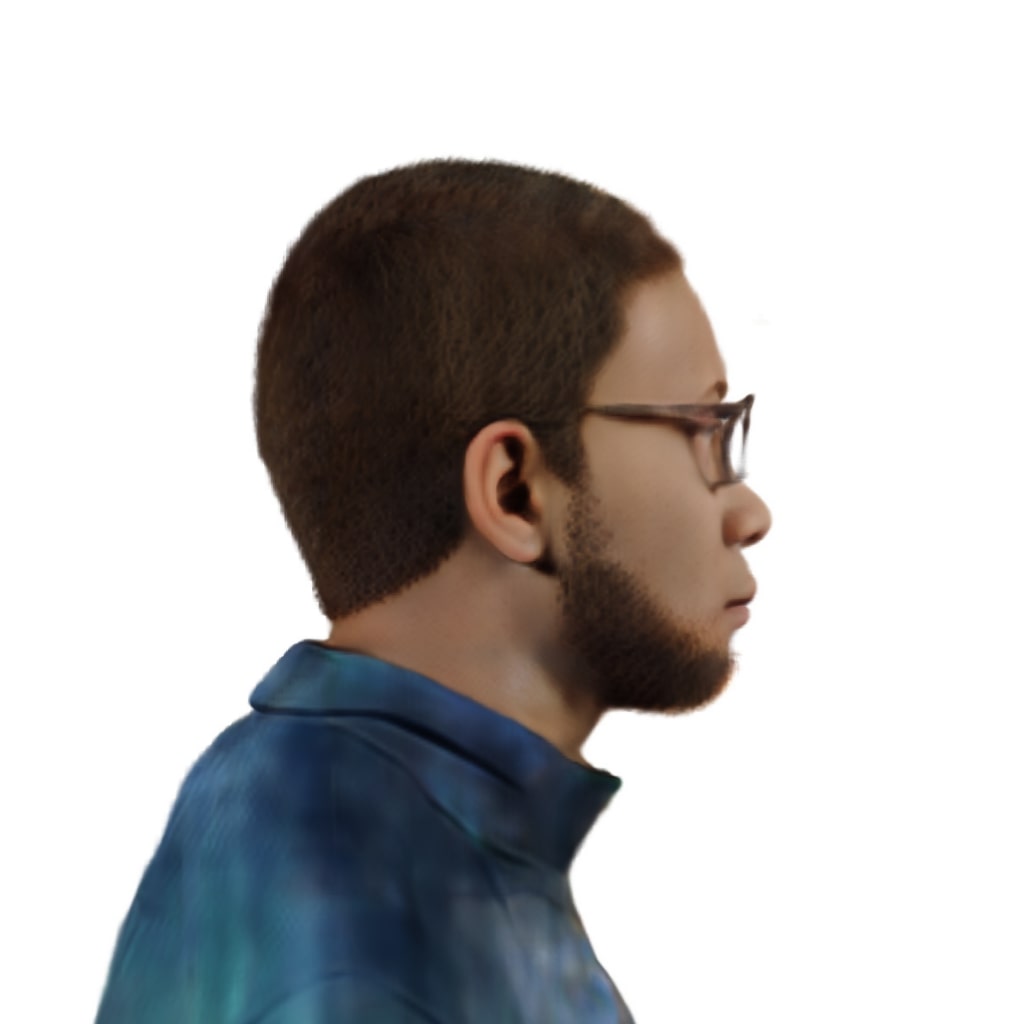} &
    \includegraphics[width=0.18\linewidth]{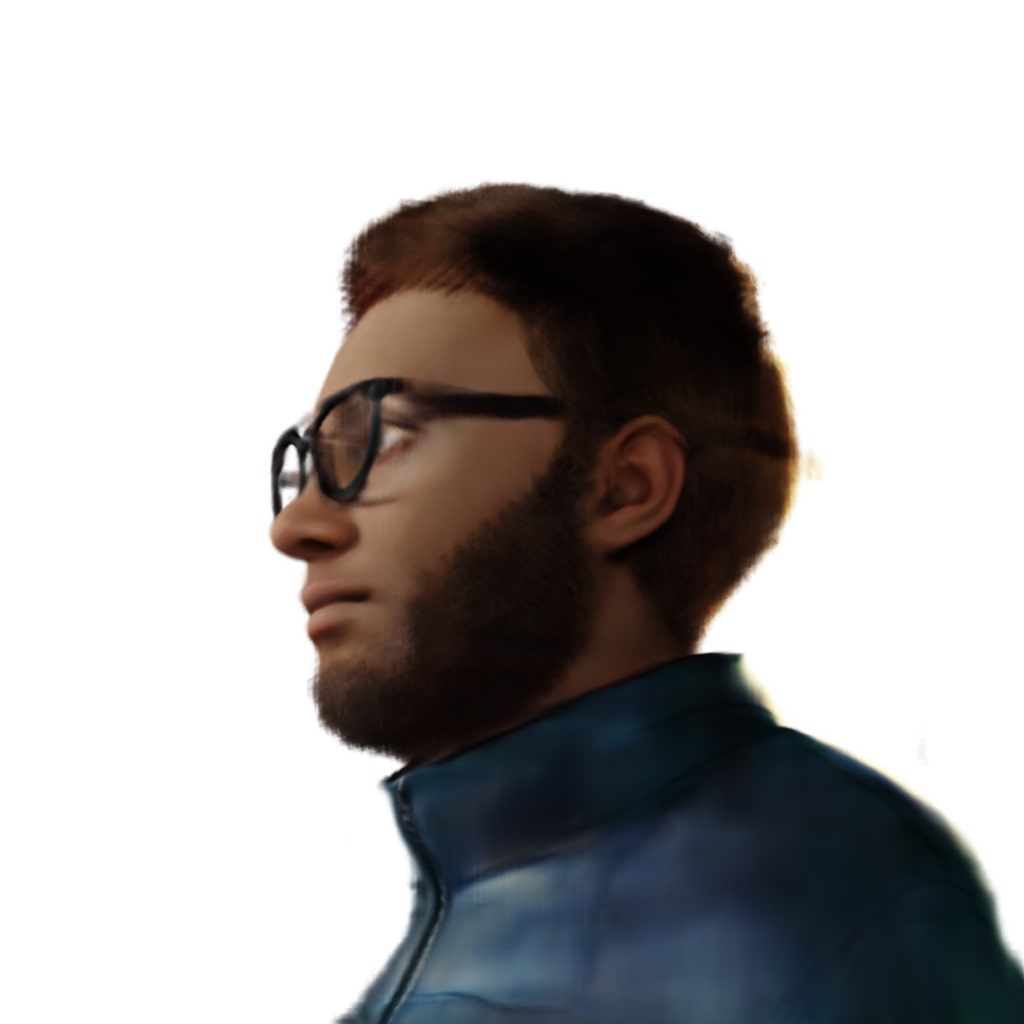} \\
    \multicolumn{2}{c}{(a)} &
    \multicolumn{2}{c}{(b)} \\
  \end{tabular}
  \caption{Failure cases.}
  \label{fig/supp:failure_cases}
\end{figure*}

As illustrated in~\cref{fig/supp:failure_cases}, our model still has some limitations. Floating points occasionally appear in the generated avatars as shown in~\cref{fig/supp:failure_cases} (a), which are typical NeRF artifacts~\cite{barron2022mip}. Handling glasses remains challenging due to limited training data in~\cref{fig/supp:failure_cases} (b).

\clearpage

\end{document}